%% file: main.tex
\documentclass{article}
\usepackage{Arxiv25}

\usepackage{times}
\usepackage{soul}
\usepackage{url}
\usepackage{hyperref}
\hypersetup{
  colorlinks,
  citecolor=blue,
  linkcolor=red,
  urlcolor=purple
}

\usepackage[utf8]{inputenc}
\usepackage{graphicx}
\usepackage{subcaption}
\usepackage[font=small]{caption}

\usepackage{amsmath}
\usepackage{amssymb}
\usepackage{amsthm}
\usepackage{booktabs}
\usepackage{algorithm}
\usepackage{algorithmic}
\usepackage[switch]{lineno}
\usepackage{makecell}
\usepackage{enumitem}
\usepackage{transparent}
\usepackage{colortbl}

\usepackage[table]{xcolor}          %
\usepackage{multirow}  
\usepackage{xcolor}    

\usepackage{lineno}

\title{Random Walk Guided Hyperbolic Graph Distillation}

\author{
Yunbo Long$^1$
\and
Liming Xu$^1$\and
Stefan Schoepf$^1$\and
Alexandra Brintrup$^{1,2}$\\
\affiliations
$^1$ Department of Engineering, University of Cambridge, Cambridge, UK \\
$^2$ The Alan Turing Institute, London, UK
\emails
\{yl892, lx249, ss2823, ab702\}@cam.ac.uk
}

\begin{document}

\maketitle

\begin{abstract}
Graph distillation(GD) is an effective approach to extract useful information from large scale network structures. 
However, existing methods, which operate in {\it Euclidean} space to generate condensed graphs, struggle to capture the inherent tree-like geometry of real-world networks, resulting in distilled graphs with limited task-specific information for downstream tasks.
Furthermore, these methods often fail to extract dynamic properties from graphs, which are crucial for understanding information flow and facilitating graph continual learning.
This paper presents the \underline{Hy}perbolic Graph \underline{D}istillation with \underline{R}andom Walks \underline{O}ptimization (HyDRO), a novel graph distillation approach that leverages hyperbolic embeddings to capture complex geometric patterns and optimize the spectral gap in {\it Hyperbolic} space. 
Experiments show that HyDRO demonstrates strong task generalization, consistently outperforming state-of-the-art methods in both node classification and link prediction tasks.
HyDRO also effectively preserves graph random walk properties, producing condensed graphs that achieve enhanced performance in continual graph learning. 
Additionally, HyDRO achieves competitive results on mainstream graph distillation benchmarks, while maintaining a strong balance between privacy and utility, and exhibiting robust resistance to noises.
\end{abstract}

\section{Introduction}

\begin{figure*}[t!]
    \centering
    \includegraphics[width=0.95\textwidth]{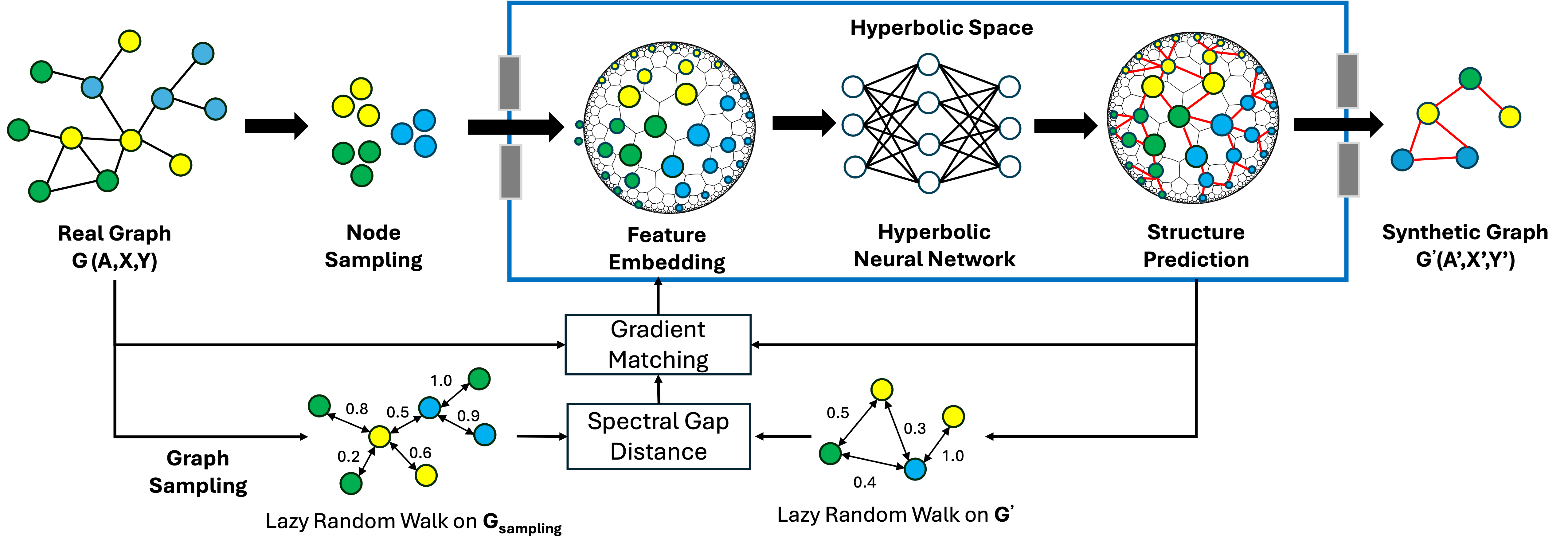} 
    \caption{The pipeline of the proposed approach---HyDRO---for graph distillation.}
    \label{fig:flow} 
\end{figure*}

Graph distillation, also referred to as graph condensation, has been gaining significant attention recently \cite{gao2024graph,hashemi2024comprehensive}.
It focuses on distilling useful information from original graphs into smaller synthetic graphs that can serve as efficient substitutes.
These condensed graphs can offer significant benefits for training graph neural networks (GNNs) in machine learning tasks, effectively reducing redundancy in large graphs while minimizing computational and storage costs.
However, traditional methods, such as graph sparsification~\cite{fung2011general} and graph coarsening~\cite{loukas2018spectrally}, often struggle to preserve the critical information necessary to achieve desirable performance of GNNs downstream tasks.
To tackle this challenge, GCond \cite{jin2021graph}, the first graph distillation method inspired by dataset distillation \cite{wang2018dataset}, introduced a gradient matching approach that simultaneously optimizes both graph structure and node features. 
Building on this, SFGC \cite{zheng2024structure} innovatively distils graphs into a set of nodes without an explicit graph structure, embedding topological information within node attributes. 
Furthermore, other state-of-the-art methods, such as GDEM \cite{liu2023graph} and GCSNTK \cite{liu2022graph}, respectively demonstrate that eigenbasis alignment \( U \) and kernel matrices \( K \) can better preserve structural information and enhance downstream task performance.

All these current methods assume an Euclidean structure for graph data, optimizing graph features and structures by performing operations (e.g., distance calculation) in Euclidean space.
However, real-world graphs often exhibit non-Euclidean topologies and hidden hierarchies \cite{fan2019graph,liang2022survey}, characterized by power-law degree distributions. 
As a result, these methods {\it cannot} yield accurate measurements of intrinsic distances between nodes, resulting in distortions in graph data, particularly in heavy-tailed regions \cite{linial1995geometry}. 
Hyperbolic geometric space, with its exponentially growing spatial capacity, is more well-suited for capturing tree-like and hierarchical structures than Euclidean space \cite{balazevic2019multi}, making it more resilient to distortions and noises in real-world graphs.
Hyperbolic neural network have also been shown to perform well in learning GNNs-based tasks, such as node classification and link prediction~\cite{wen2024hyperbolic}.
Moreover, compared to Euclidean space, where high-dimensional data boundaries are more distinctly defined, the complex and non-linear geometry of hyperbolic space makes it more difficult for attackers to model decision surfaces, thereby potentially enhancing privacy preservation.

Although condensed graphs are designed to achieve good performance across multiple tasks, preserving structural properties--—such as graph density and degree distribution—--remains challenging~\cite{gong2024gc}. 
This challenge is because graph distillation methods, which aim to condense large, sparse graphs into smaller, denser ones, often modify the structure in a manner that inadvertently disrupts these features.
Current state-of-the-art methods primarily focus on preserving the graph's global structure and static spectral properties, such as the Laplacian energy distribution in SGDD~\cite{yang2024does} and the eigenvectors in GDEM~\cite{liu2023graph}.
However, dynamic properties in condensed graphs, such as random walk behavior \cite{broder1997static}, is less explored.
For example, citation or social networks often involve diffusion processes, and it is critical to understand how quickly information or influence can spread across the network \cite{lovasz1993random,burioni2005random}. 
By analyzing dynamic properties like random walks in condensed graphs, researchers can better simulate the efficiency of phenomena propagation~\cite{leonard2016lazy} and gain a deeper understanding of both the local and global structures of the graph~\cite{perozzi2014deepwalk}. 
Besides, condensed graphs can be applied to Continual Graph Learning (CGL)~\cite{liu2023cat} for managing evolving graph data in resource-constrained environments. 
However, current methods focusing on static graph features fail to capture dynamic changes, hindering their ability to mitigate catastrophic forgetting and maintain performance in real-world, evolving graph settings.

To address these challenges, 
we introduce HyDRO (Hyperbolic Graph Distillation with Random Walk Optimization), a model that generates tree-like structures in condensed graphs by embedding original graph features in hyperbolic space. 
By preserving random walk behavior, HyDRO guides the distillation process through spectral gap minimization. 
We evaluate HyDRO on downstream task performance, privacy, robustness, and dynamic structure preservation, achieving state-of-the-art results in most evaluations.

The main contributions are as follows.
\begin{itemize}[nosep]
    \item We propose HyDRO, which captures more detailed graph structures in condensed graphs, and outperforms state-of-the-art methods by achieving the highest accuracy in link prediction tasks, while also maintaining competitive performance in node classification.
    
    \item We demonstrate that HyDRO is the first graph distillation method to preserve dynamic information in graphs, achieving state-of-the-art performance in continual graph learning tasks.
    
    \item Comprehensive experiments demonstrate that HyDRO offers a more well-rounded solution to graph distillation, effectively balancing performance across mainstream tasks, dynamic information preservation, membership privacy protection, and noise robustness, outperforming state-of-the-art graph distillation methods in these key areas.
\end{itemize}


\section{Preliminaries}\label{sec:preliminary}

\subsection{Poincar\'e Ball Manifold}

The Poincar\'e ball manifold is a model of hyperbolic geometry represented as the open unit ball in $\mathbb{R}^d$, which can be formally defined as $B^d = \{ x \in \mathbb{R}^d \mid \|x\|^2 < \frac{1}{c} \}$, where $\|x\|$ denotes the Euclidean norm and $c > 0$ is a constant determining the curvature of the manifold. 
The radius of the Poincar\'e ball is given by $r = \frac{1}{\sqrt{c}}$. 
The Poincar\'e ball supports various operations crucial for differential geometry, such as distance computation, exponential and logarithmic mappings, and vector transport. 
For the details of operations in Poincar\'e ball manifold, see \autoref{app:preliminary}.

\subsection{Lazy Random Walk Matrix}

In graph theory, a random walk \cite{lawler2010random} in a graph is a stochastic process where a walker moves from one node to another node based on given transition probabilities. 
This matrix is called the {\bf random walk matrix}, denoted as \(\mathbf{W}\). 
For a standard random walk, the transition probability from node \(i\) to node \(j\) is given by \(\mathbf{W} = D^{-1}A\), where \(D\) represents the diagonal degree matrix and \(D_{ii}\) is the sum of the weights of edges connected to node \(i\). 
This matrix indicates how the random walk transitions between nodes according to the graph's structure. 
In contrast, a \textbf{lazy random walk matrix} \( W_{\text{lazy}} \)~\cite{xia2019random} is a variant that includes a probability of staying at the current node, which ensures that the random walk is ergodic, meaning it will eventually visit every node if the graph is connected. 
The lazy random walk matrix is given by \(W_{\text{lazy}} = \epsilon I + (1 - \epsilon) D^{-1}A\), where \(D\) is the degree matrix, \(A\) is the adjacency matrix, \(I\) is the identity matrix, and \(\epsilon \in [0, 1]\) is a coefficient that adjusts the probability of staying at the current node or transitioning to a neighboring node. 
By setting \(\epsilon = \frac{1}{2}\), the lazy random walk ensures a balanced trade-off between exploration and staying behavior, facilitating controlled exploration of the graph.


\subsection{Spectral Gap}


The spectral gap~\cite{hoffman2021spectral} is the difference between the moduli of the largest eigenvalue (\(\omega_1\)) and the second-largest eigenvalue (\(\omega_2\)) of \(W\), measuring the rate at which the random walk converges to its stationary distribution. 
The eigenvalues of \(W\), denoted as \(\omega_i\), are related to the eigenvalues of the normalized Laplacian \(L\) by \(\omega_i = 1 - \frac{\lambda_i}{2}\), where \(L = I - D^{-1/2}AD^{-1/2}\) and \(\lambda_i\) are the \(i\)-th eigenvalues of \(L\). 
Therefore, the spectral gap (\(\omega_1 - \omega_2\)) can be simplified to \(\frac{\lambda_2}{2}.
\) 
A larger spectral gap leads to faster convergence to the stationary distribution. 
Further information about the impact of the spectral gap on the random walks and graph diffusion process can be found in \autoref{app:preliminary}.

\section{HyDRO}\label{sec:Methodology}

This section shows the pipeline of the proposed HyDRO approach and describes the workflow to distill the random walk properties into the synthetic data in hyperbolic space.

\subsection{Hyperbolic Graph Distillation}

Graph distillation involves distilling a smaller graph \( \mathcal{G'} = \{\mathbf{A'}, \mathbf{X'}, \mathbf{Y'}\} \) such that a model training on \( \mathcal{G'} \) can achieve comparable performance to the original graph \( \mathcal{G} = \{\mathbf{A}, \mathbf{X}, \mathbf{Y}\} \) \cite{jin2021graph}. 
To capture hierarchical structures, node features \( \mathbf{X'} \) are embedded into hyperbolic space using hyperbolic neural networks \cite{ganea2018hyperbolic}, with optimization via Riemannian stochastic gradient descent, which adapts gradient descent to hyperbolic geometry.
For the Riemannian gradient, we incorporate momentum \( \mu \) to accelerate convergence and weight decay \( \lambda \) for regularization, adapting them to the Riemannian geometry of hyperbolic space.
And the input matrix \( \mathbf{X'} = [\mathbf{x}'_1, \mathbf{x}'_2, \dots, \mathbf{x}'_n] \) consists of feature vectors \( \mathbf{x}'_i \) for each node \( i \). 
Given \( \mathbf{X'} \), the edge features are computed by concatenating the feature vectors of the pairs of nodes \( (\mathbf{x}'_i, \mathbf{x}'_j) \), which produces an edge embedding:
\begin{equation}
    \mathbf{e}_{ij} = \mathbf{x}'_i \oplus \mathbf{x}'_j \in \mathbb{R}^{2d}
    \label{eq:edge_embedding}
\end{equation}
where \( d \) is the dimension of the node features. 
The operation \( \oplus \) represents the merging of two vectors, \( \mathbf{x}'_i \) and \( \mathbf{x}'_j \), into a single vector by appending the elements of the second vector to the first.
We then apply the exponential map from Euclidean space to hyperbolic space, denoted as \( \mathcal{M}_{h} \), as follows:
\begin{equation}
    \mathbf{e}_{ij}^{\text{h}} = \mathcal{M}_{h}(\mathbf{e}_{ij}) \in \mathbb{H}^d
    \label{eq:expmap}
\end{equation}

At each layer of the hyperbolic neural network, we transform \( \mathbf{e}_{ij}^h \) using a Möbius linear transformation \( \mathcal{T}_{\text{Mobius}} \) \cite{mccullagh1996mobius}, which is specifically adapted for the Poincaré ball. The transformation is represented as:
\begin{equation}
    \begin{aligned}
        \mathbf{z}^{\text{h}} &= \mathcal{T}_{\text{Mobius}}(\mathbf{e}_{ij}^{\text{h}}, \mathbf{W}, \mathbf{b}, \kappa, \mathcal{B}) \\
        &= \mathcal{M}_{\exp, \mathbf{b}} \left( \mathbf{W} \cdot \mathcal{M}_{\log, 0}(\mathbf{e}_{ij}^{\text{h}}, \kappa, \mathcal{B}) \right) \in \mathbb{H}^d
    \end{aligned}
\end{equation}
\label{eq:mobius_linear}
where \( \mathbf{W} \) is the weight matrix, \( \mathbf{b} \) is the bias vector, \( \mathcal{M}_{\log, 0} \) is the logarithmic map that takes points from the hyperbolic manifold to the tangent space at the origin, and \( \mathcal{M}_{\exp, \mathbf{b}} \) maps back from the tangent space to the hyperbolic manifold at point \( \mathbf{b} \). And \( \kappa \) represents the curvature of the Poincaré ball, and \( \mathcal{B} \) refers to the Poincaré ball manifold. The curvature \( \kappa \) is incorporated into the logarithmic and exponential maps to adjust the transformation according to the geometry of the Poincaré ball, ensuring that the transformations are consistent with the hyperbolic space's curvature.
The output \( \mathbf{z}^{\text{h}} \) is then passed through multiple layers, denoted by \( \mathcal{L}_i \), where each layer consists of an instance of \( \text{MobiusLinear} \). This sequential processing enables the model to capture complex relationships in the data through multiple transformations. 
After each layer, except for the last one, a batch normalization layer is applied to stabilize the learning process by normalizing the output. Following this, a hyperbolic ReLU activation function \cite{ganea2018hyperbolic} is applied.
This function initially converts the hyperbolic input into Euclidean space, applies the conventional ReLU activation, and subsequently maps the output back to hyperbolic space, thereby preserving the manifold structure. 
The output from the layers is reshaped into a square adjacency matrix \( \text{adj} \in \mathbb{R}^{n \times n} \), where each entry represents the strength of potential connections between nodes.
To ensure that the adjacency matrix remains undirected, it is symmetrized by averaging it with its transpose \( \text{adj} = \frac{1}{2}(\text{adj} + \text{adj}^{T}) \). Following this, a sigmoid activation function is employed to restrict the matrix element values between 0 and 1, allowing them to be interpreted as connection probabilities. Additionally, to eliminate self-loops (connections from a node to itself), the diagonal elements of the adjacency matrix are set to zero, yielding \( \mathbf{A'} = \text{adj} - \text{diag}(\text{adj}) \).

\subsection{Spectral Gap Alignment}
In this section, we detail our approach for optimizing the random walks properties in graph-based models using a spectral gap closure strategy. 
To compute the spectral gap, we first define the sampling adjacency matrix \( A_{\text{sub}} \) as the adjacency matrix sampled from the real graph \( A \) at each epoch. 
The lazy random walk matrix for the condensed graphs can be defined as $W_{\text{syn}} = \frac{1}{2} \left(I + D^{-1}A'\right). $
For the real graph, since \( A_{\text{sub}} \) is typically sparse, it leads to inefficiencies and inaccuracies in random walk calculations. 
By normalizing \( A_{\text{sub}} \), we ensure that each node's influence on the random walk is properly accounted for. 
The normalized adjacency matrix \( \tilde{A}_{\text{sub}} \) can then be calculated as follows:
\begin{equation}
    \tilde{A}_{\text{sub}} = A_{\text{sub}} + I, \quad \tilde{W}_{\text{sub}} = D^{-1/2} \tilde{A}_{\text{sub}} D^{-1/2} 
\end{equation}
where \( I \) is the identity matrix and \( D \) is the degree matrix. 
The lazy random walk matrix for the sampled real adjacency is then given by $
    W_{\text{sub}} = \frac{1}{2} \left(I + \tilde{W}_{\text{sub}}\right). $
Next, we calculate the eigenvalues \( \lambda_{\text{sub}} \) and \( \lambda' \) of the matrices \( W_{\text{sub}} \) and \( W' \), respectively. 
The spectral gaps for both the sampled and the synthetic adjacency matrices are defined as follows $g_{\text{sub}} = 1 - \lambda_{2,\text{sub}}, \quad g_{\text{syn}} = 1 - \lambda_{2,\text{syn}}$, where \( \lambda_{2,\text{sub}} \) and \( \lambda_{2}' \)  are the second largest eigenvalues of \( W_{\text{sub}} \) and \( W' \), respectively. 
The spectral gaps provide critical insights into the traversal characteristics of random walks, with larger gaps implying more efficient exploration of the graph structure \cite{coja2009spectral}.
To enhance random walks, we define a loss function based on the spectral gaps $\mathcal{L}_{\text{rw}} = |g_{\text{syn}} - g_{\text{sub}}|.$


\subsection{Pipeline}
As illustrated in \autoref{fig:flow}, HyDRO first samples nodes in Euclidean space. 
The features of these sampled nodes, denoted as \( \mathbf{X^\prime} \), are then embedded into the Poincar\'e model manifold \( \mathcal{M}_\mathcal{P} \)~\cite{nickel2017poincare} within hyperbolic space \( \mathbb{H}^n \).
A hyperbolic neural network is employed to generate the synthetic graph structure \( \mathbf{A^\prime} \) in this non-Euclidean space. 
Simultaneously, we compute the spectral gap distance between the original graph \( \mathbf{A_{\text{sampling}}} \) and its synthetic graph \( \mathbf{A'} \). 
The primary optimization objective is to minimize the total loss for hyperbolic graph distillation, which includes a gradient matching loss \cite{jin2021graph} that measures the discrepancy between the learned parameters \( \theta_{\mathcal{G'}}^* \) on the synthetic dataset \( \mathcal{G'} \) and the parameters \( \theta_{\mathcal{G}}^* \) obtained from the original dataset \( \mathcal{G} \). 
This optimization process can be formulated as follows:
\begin{equation}
    \begin{aligned}
        \mathcal{G'}^* &= \arg \min_{\mathcal{G'}} \mathcal{M}(\theta_{\mathcal{G'}}^*, \theta_{\mathcal{G}}^*) \quad \\ 
        s.t. \quad \theta_t^* &= \arg \min_{\theta} \mathcal{L}(\text{GNN}_{\theta}(t)), \quad t \in \{\mathcal{G'}, \mathcal{G}\}
    \end{aligned}
\end{equation}

Additionally, we incorporate a loss term for the spectral gap distance, along with a regularization loss. 
The total optimization loss is thus denoted as:
\begin{equation}
    L_{\text{total}} = L_{\text{gm}} + L_{\text{rw, norm}} + \beta L_{\text{reg}}
\end{equation}
where \( \beta \) is the regularization coefficient, helping prevent overfitting by constraining the model parameters.

\section{Experimental Settings}
\label{sec:experiment_setting}

\input{tables/NC}

\input{tables/NC_LP}


\paragraph{Datasets.} 
We evaluate the performance of HyDRO across multiple tasks using a set of benchmark graph datasets that are conventionally adopted for evaluating graph distillation, including 
Cora~\cite{kipf2016semi}, 
Citeseer~\cite{kipf2016semi}, 
PubMed~\cite{kipf2016semi},
DBLP~\cite{tang2008arnetminer},
ogbn-arxiv~\cite{hu2020open}, 
Flickr~\cite{zeng2019graphsaint}, 
Reddit~\cite{zeng2019graphsaint}, and DBLP~\cite{tang2008arnetminer}. 
Wiki-CS~\cite{mernyei2022} and Coauthor-Physics~\cite{shchur2018pitfalls} are also used.
These two datasets have larger number of node classes and thus are more suitable for evaluating CGL.
More details about these datasets are presented in \autoref{app:dataset}.

\paragraph{Baselines.}\label{par:baselines}
We compare it against both traditional graph reduction methods and state-of-the-art graph distillation approaches in various tasks.
For traditional methods, we include random selection, Herding~\cite{welling2009herding}, KCenter~\cite{sener2017active}, Coarsening~\cite{huang2021scaling}, and Averaging ~\cite{msgc}.
For graph distillation methods, we consider both structure-free (e.g., GCSNTK~\cite{wang2024fast}, GEOM~\cite{zhang2024navigating}, and SFGC~\cite{zheng2024structure}) and structure-based methods (e.g., GCOND~\cite{jin2021graph}, DosCond~\cite{doscond}, MSGC~\cite{msgc}, GDEM~\cite{liu2023graph}, and SGDD~\cite{yang2024does}).
These methods represents the latest developments in graph distillation.

\paragraph{Evaluation Settings.}
Following common practices in previous studies ~\cite{jin2021graph,liu2023graph}, benchmark tasks such as node classification, neural architecture search (NAS), and cross-architectural transferability are used for evaluation. 
Additionally, further tasks are incorporated into experiments to provide a more comprehensive assessment.
These tasks include task transferability (e.g., from node classification to link prediction), preservation of random walk dynamics, impact on continual graph learning, privacy preservation, and denosing.
Together, these tasks would offer a thorough evaluation of HyDRO's performance.

Adhering to the setup in~\cite{jin2021graph}, both transductive and inductive settings are adopted: transductive settings are applied to the Cora, PubMed, Citeseer, DBLP and Ogbn-arxiv datasets, while inductive settings are used for the Flickr and Reddit datasets.
In node classification, previous studies have not adopted a common configuration, each study defines its own settings and reports the best results \cite{sun2024gc}.
We thus follow the same setting as the seminar papers \cite{jin2021graph,yang2024does} in this experiment.
For the remaining experiments, we standardize the evaluation process by selecting distilled graphs based on validation results and subsequently evaluating them on the test sets of the original graphs. 
To evaluate the distilled graphs, a two-layer graph convolution network (GCN) with 256 hidden units is trained for 500 epochs.

\paragraph{Setup and Hyperparameters.} 
We adopted the same graph preprocessing procedures and dataset split as described in the prior work~\cite{jin2021graph} (See \autoref{app:dataset}). 
We utilized hyperbolic neural network \cite{ganea2018hyperbolic} for distilling graphs. 
For the node classification task, we performed a limited hyperparameter search for HyDRO to reduce fine-tuning cost~\cite{gong2024gc}. 
The number of hidden layers was chosen from the set \{2, 3, 4\}, while the number of hidden units was selected from \{128, 256\}. 
Additionally, the regularization coefficient ($\beta$) was fixed at 0.1, with momentum values selected from \{0, 0.01\} and curvature values from \{0.01, 0.1\}.
For the remaining tasks, we fine-tuned key hyperparameters, such as the learning rate and number of hidden layers, while keeping other hyperparameters consistent with their original values as defined in the respective papers.
See more about hyperparameter fine-tuning in \autoref{app:hyperparameters}.

\section{Results and Discussion}\label{sec:results}
This section details the results of the experiments as described in \autoref{sec:experiment_setting}.
The evaluation is repeated {\it ten} times, and the mean and standard deviation of the results are computed and reported in the format $\text{mean} \pm \text{std dev}$.
The best results are highlighted in {\color{blue}\bf blue boldface}, while the second-best results are shown in {\bf bold}. 

\subsection{Downstream Tasks}
\label{sec:NC}

\begin{figure*}[ht]
    \centering
    \begin{subfigure}[t]{\textwidth}
        \centering
        \includegraphics[width=\textwidth]{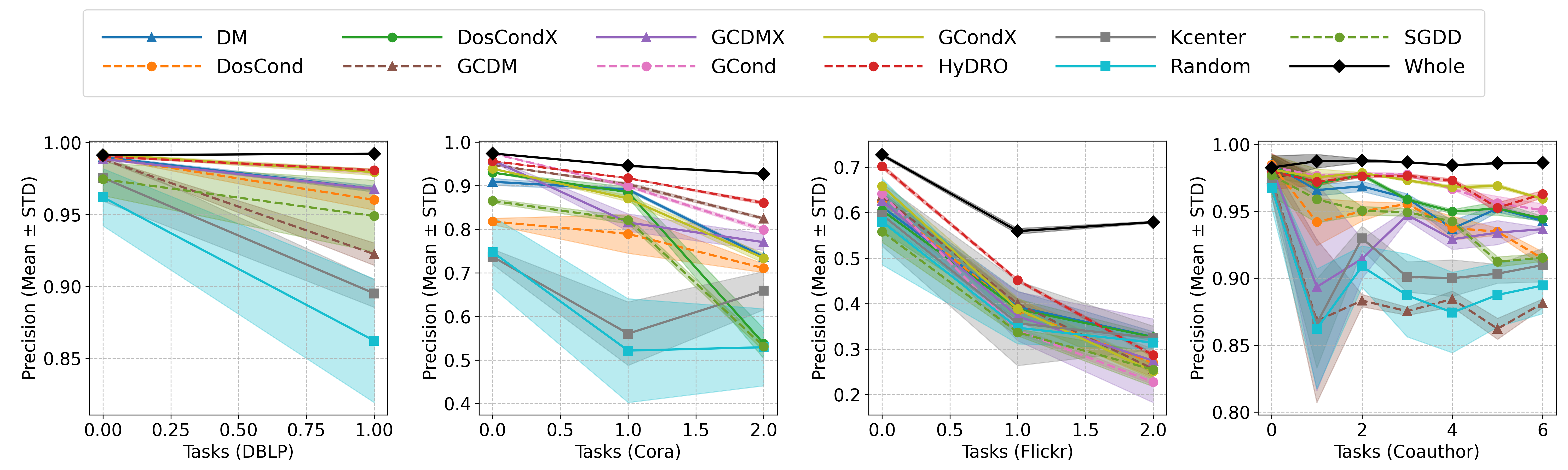}
    \end{subfigure}%
    \caption{Performance of graph distillation methods on mainstream datasets under the graph continual learning setting.}
    \label{fig:cgl-all}
\end{figure*}

\paragraph{Node Classification.} 
\label{sec:results}
The test results for node classification are presented in \autoref{tab:node_classification}. 
As shown in \autoref{tab:node_classification}, graph distillation methods significantly outperform traditional methods.
Our proposed method, HyDRO, achieves the highest accuracy on the PubMed dataset among all methods and ranks second on the Cora, Citeseer, and Ogbn-arxiv datasets, just slightly behind GEOM. 
However, in contrast to GEOM which only focuses on node feature optimization, HyDRO can learn graph structural elements and outperforms all structure-based methods on datasets in transductive settings. 
In Flickr and Reddit datasets, which include larger graphs, HyDRO surpasses GEOM and achieves the second highest accuracy after GDEM. 
However, we encountered the same problem in reproducing the results of GDEM in large graphs for node classification, as reported by other studies \cite{gong2024gc4nc,liu2024gcondenser}.
In the Ogbn-arxiv dataset (reduction rate of 0.05\%) and the Reddit dataset (reduction rate of 0.01\%), HyDRO achieves lower accuracy than GEOM and GDEM---the two state-of-the-arts in their category---but it still ranks third among all 10 methods under study.

\paragraph{Link Prediction.}
We further investigate whether HyDRO can extract condensed graphs that are richer in information, thereby enhancing tasks generalization.
To this end, we introduce the link prediction task, which may require more graph information than node classification.
For this experiment, we adopted the same setting for all the methods to ensure a fair comparison.
Each model was fine-tuned on the node classification task, and the fine-tuned models were then evaluated on the link prediction task.
The evaluation results are shown in \autoref{tab:coss-tasks}, which includes outcomes for both node classification and link prediction.
As shown in \autoref{tab:coss-tasks}, graph distillation methods significantly outperform traditional methods in link prediction; and structure-based graph distillation methods achieve higher prediction accuracy.
Among all methods, HyDRO achieves the best overall performance, with the highest link prediction accuracy on almost all datasets and fairly good performance in node classification, highlighting HyDRO's effectiveness in task generalization.
Except HyDRO, the performance of all the other methods show a downgrade from node classification to link prediction, showing less effectiveness in task generalization.

\subsection{Neural Architecture Search}\label{sec:NAS}
This section presents the evaluation results of HyDRO in the task of neural architecture search (NAS)---a task aimed at automatically discovering optimal neural network architectures.
APPNP (Approximate Personalized Propagation of Neural Predictions) \cite{klicpera2018predict-appnp}, which involves more hyperparamters than GCN, is selected for testing. 
We use same evaluation metrics as the prior studies \cite{jin2021graph,yang2024does,sun2024gc} for this experiment: 
Top-1 test accuracy, 
the Pearson correlation between validation accuracies, 
and the rank correlation (using Pearson coefficient) between the validation accuracies of the condensed and original graphs.
We perform architecture searches on the Citeseer, Pubmed, DBLP, and Flickr datasets, each with reduction rates of 0.9\%, 0.02\%, 0.001\%, and 0.005\%, respectively.
The experimental results are presented in \autoref{tab:NAS} of \autoref{app:nas}.
As can be seen, HyDRO achieves the overall best performance, with highest Top-1 accuracy on the Citeseer and Flickr datasets and highest rank correlation on the Flickr and DBLP datasets. 
Moreover, HyDRO has also comparable performance in other settings. 
More information on NAS evaluation results can be found in \autoref{app:nas}.

\subsection{Cross-Architecture Transferability} 
\label{sec:Transferability}

In this section, we present the evaluation results of the cross-architecture transferability task.
The experiment considers a wide range of architectures, including SGC~\cite{wu2019simplifyingsgc}, 
GCN~\cite{kipf2016semi}, 
Cheby~\cite{cheby}, 
GraphSage~\cite{hamilton2017inductive}, 
APPNP~\cite{klicpera2018predict-appnp}, and GAT~\cite{gat}. 
These models represent both spectral and spatial types of GNNs. 
Additionally, SGFormer~\cite{wu2024simplifying}---a graph transformer-based architecture---was included to evaluate the effectiveness of the condensed graphs in attention mechanism models.
Experimental results, as shown in \autoref{tab:cross_table_1} of \autoref{app:cross}, show that HyDRO achieves the best performance on Citeseer dataset and ranks second, just after GEOM, on the DBLP and PubMed datasets. 
This shows that HyDRO excel across most of GNN architectures and maintains stable performance with relatively low variance across all three datasets.
Additionally, many methods exhibit a significant performance drop on SGFormer, a transfomer-based architecture. 
This include GEOM, one of the best structure-free graph distillation methods, which shows a notable decline on the Citeseer dataset. 
In contrast, HyDRO demonstrates superior transferability on SGFormer across all three datasets.
Further details can be found in \autoref{app:cross}.

\begin{figure*}[ht]
    \centering
   
    \begin{subfigure}[t]{0.255\textwidth}
        \centering
        \includegraphics[width=\textwidth]{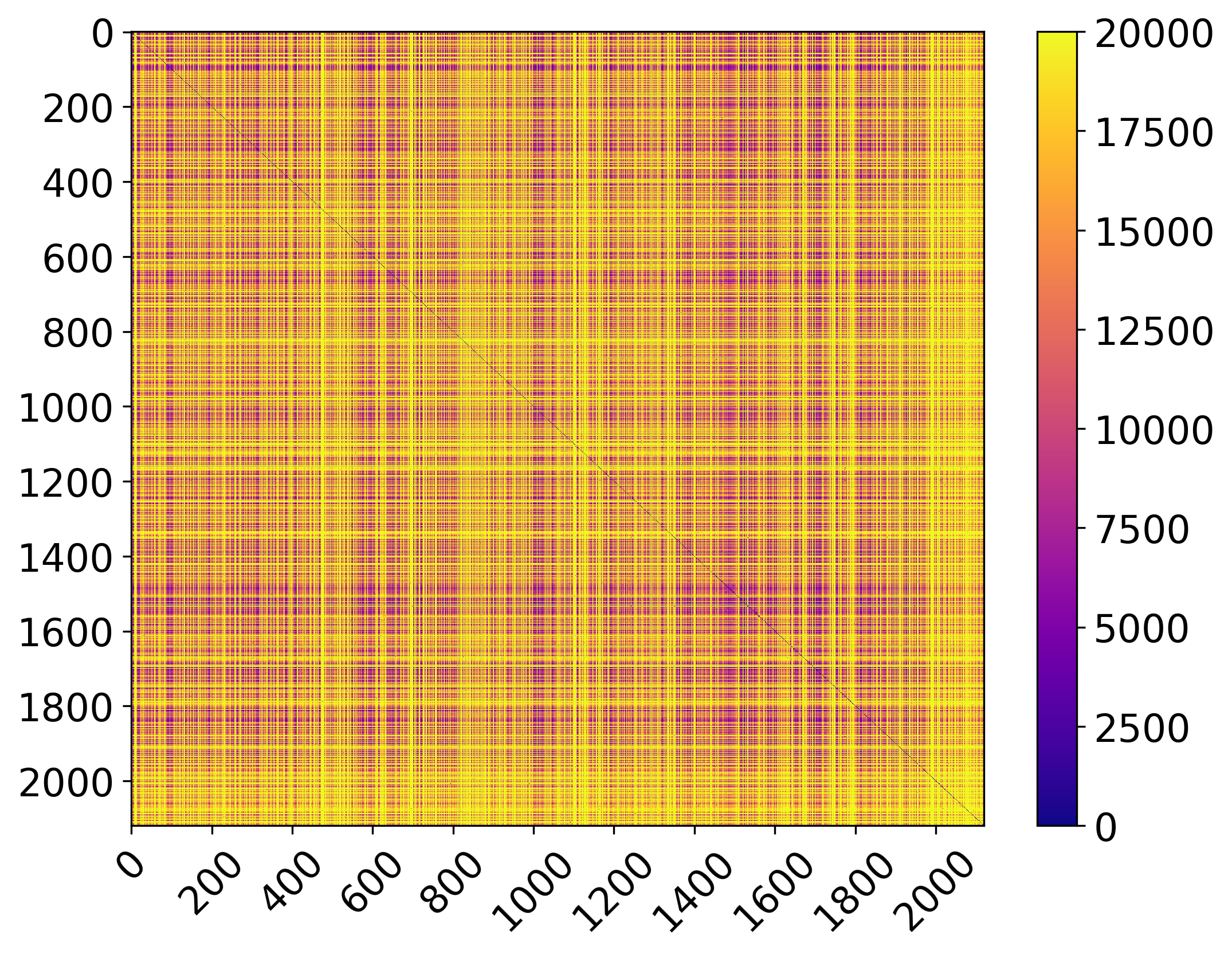}
        \caption{Original}  
        \label{fig:ct_original}
    \end{subfigure}
    \begin{subfigure}[t]{0.255\textwidth}
        \centering
        \includegraphics[width=\textwidth]{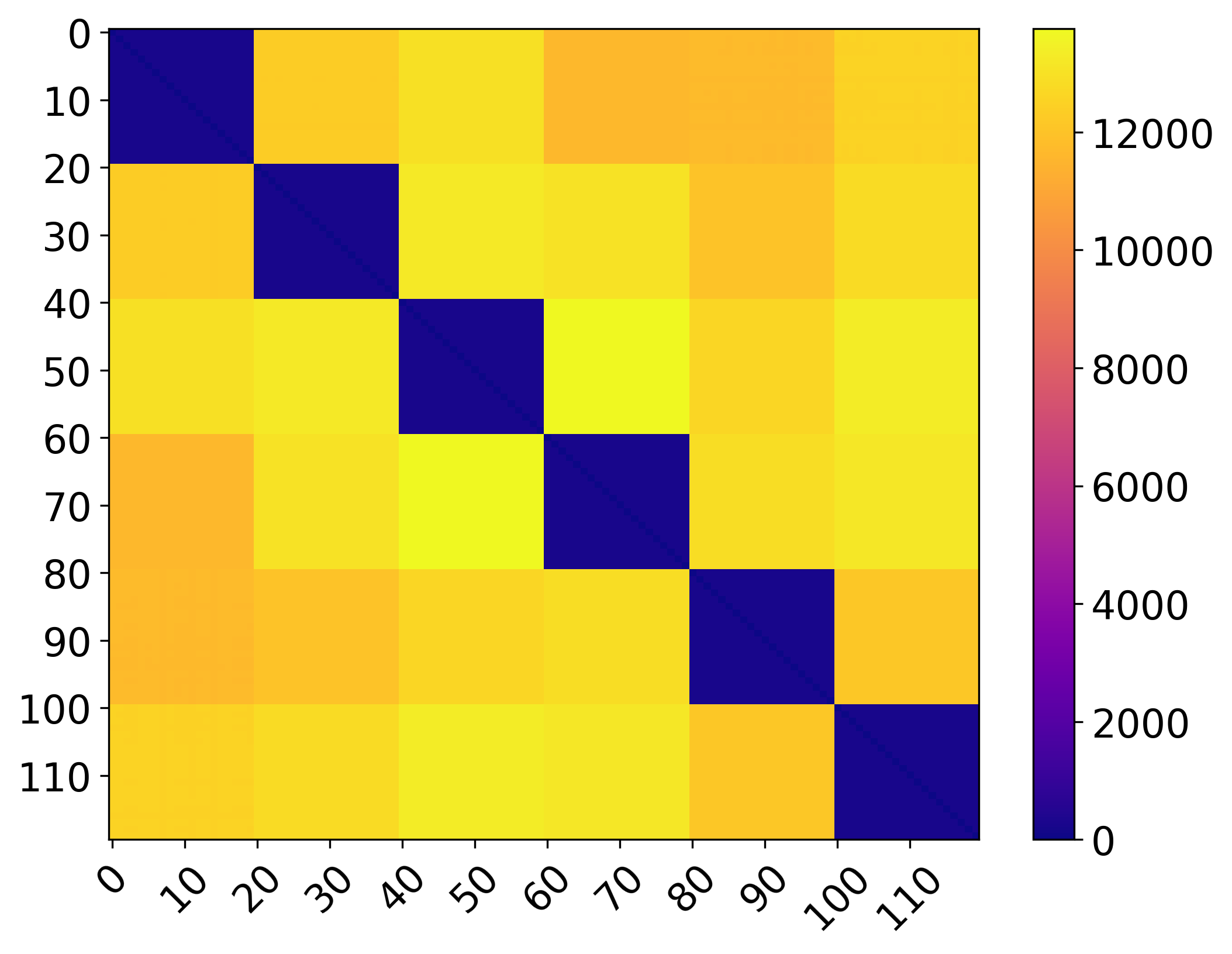}
        \caption{HyDRO}  
        \label{fig:ct_hydro}
    \end{subfigure}%
    \begin{subfigure}[t]{0.2475\textwidth}
        \centering
        \includegraphics[width=\textwidth]{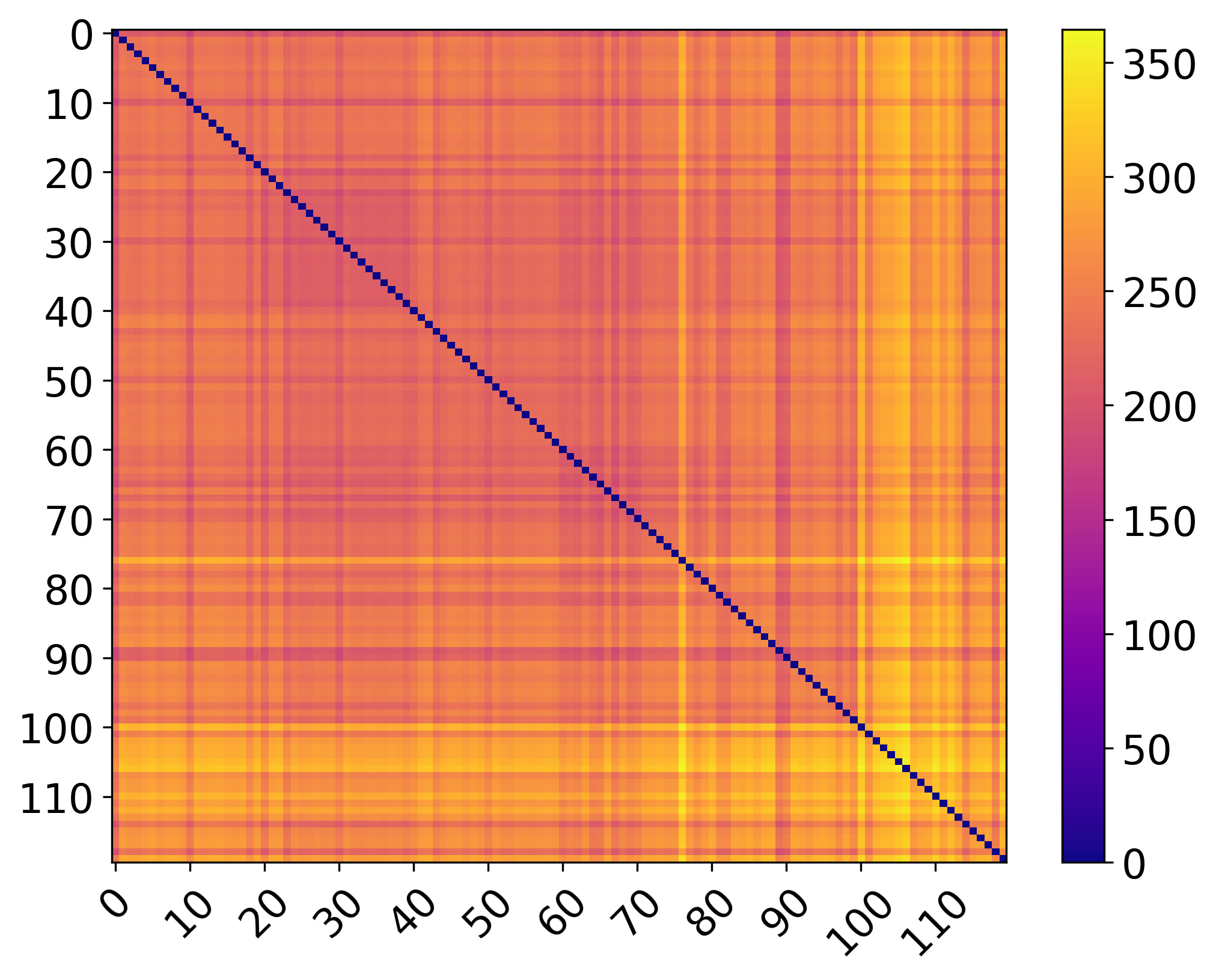}
        \caption{SGDD}  
        \label{fig:ct_sgdd}
    \end{subfigure}%
    \begin{subfigure}[t]{0.2475\textwidth}
        \centering
        \includegraphics[width=\textwidth]{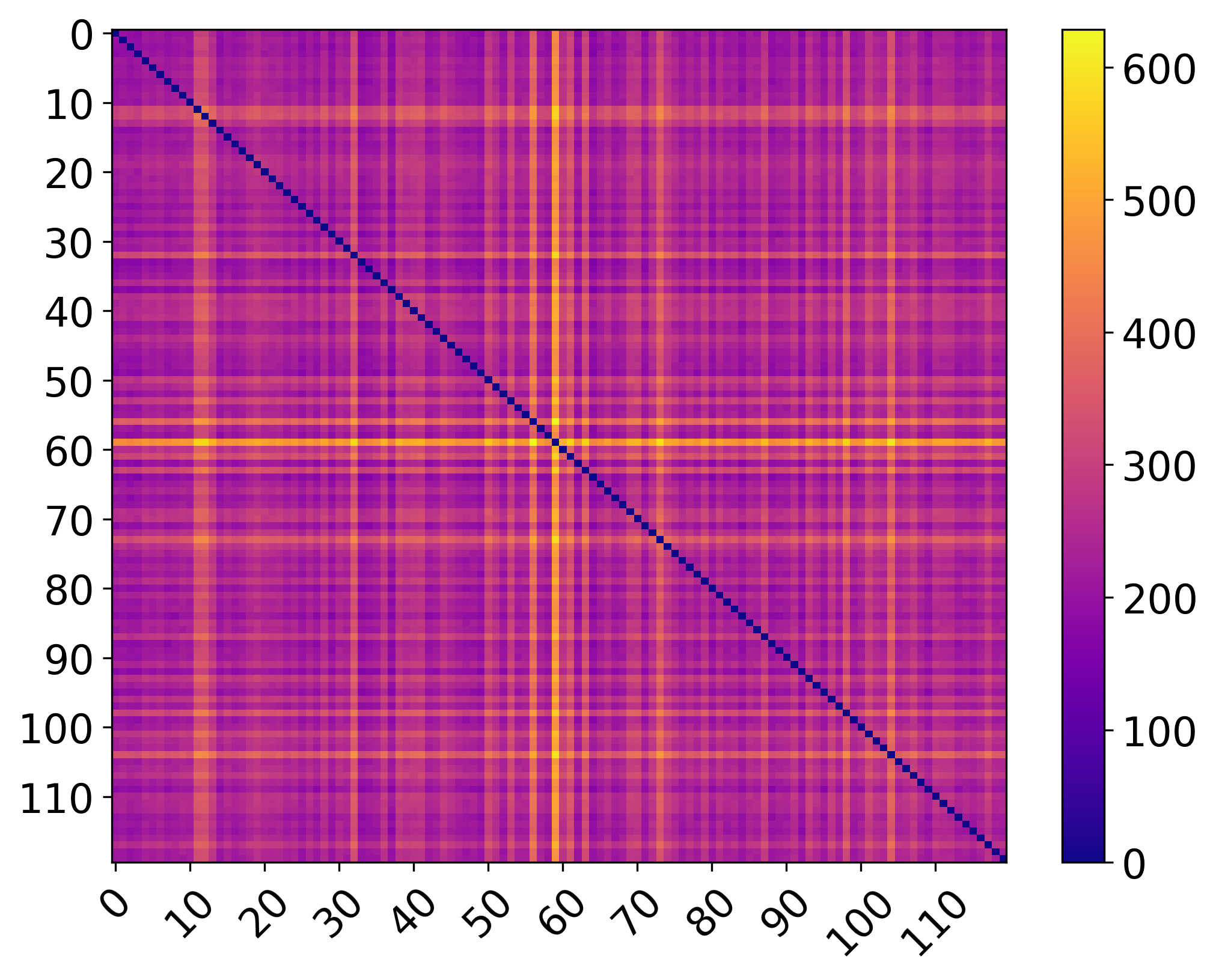}
        \caption{GDEM}  
        \label{fig:ct_gdem}
    \end{subfigure}%
    \caption{Commute time between pairs of nodes on Citeseer dataset (reduction rate 3.6\%). ({\bf Note}: {\it x} and {\it y}-axis denote the node ID. 
    The values closer to the original graph (\ref{fig:ct_original}) indicate better preservation of random walk properties.
    The color bars denote the value of commute time.)}
    \label{fig:commute}
    
\end{figure*}

\subsection{Preservation of Random Walk Properties}
\label{sec:random}

We introduce {\bf Commute Time} \cite{von2014hitting} to evaluate random walk properties of graphs. 
Commute time in a graph represents the mean number of steps it takes for a random walk to travel from one node to another and return \cite{qiu2007graph}.
As it is derived from the Laplacian matrix, commute time is relatively insensitive to changes in graph size, making it well-suited for evaluating a graph's random walk properties. 
It captures the graph's degree and adjacency structure, reflecting the ``global resistance'' between nodes. 
This metric directly influences the efficiency of information diffusion and quantifies the difficulty of transitions in graphs. 
Commute time is critical for determining the convergence time and stability of random walks, which are essential for modeling diffusion processes in graphs. 
We compare HyDRO against SGDD and GDEM, which are the only two graph distillation methods that can capture a graph's spectral properties. 
\autoref{fig:commute} shows the evaluation results. 
As illustrated in \autoref{fig:commute}, HyDRO achieves significantly better performance in preserving random walk properties, with most closely-matched commute time values as the original graph.
Further details on commute time calculations, more experiments and results can be found in \autoref{app:Random}.

\subsection{Impact on Continual Graph Learning}\label{sec:CGL}
Continual graph learning (CGL) is used to benchmark the performance of condensed graphs in dynamic environments. 
We utilise the Condense and Train (CaT) framework, as defined in ~\cite{liu2023cat}, for the continual learning evaluation.  
The graph datasets are divided into subgraphs, each introducing two new classes at each step, while single-class subgraphs are excluded to ensure consistency.
Offline approaches such as SFGC, GDEM, and GEOM, which require retraining on the entire graph, are inefficient for continual learning due to high computational costs and the risk of catastrophic forgetting. 
Therefore, this study only focuses on evaluating structure-based methods (HyDRO, GCond, DosCond, SGDD, GCDM, GCDM \cite{liu2022graph}) and structure-free methods (GCondX, DosCondX, GCDMX\cite{liu2022graph}, DM \cite{liu2023cat}), comparing them against baseline methods (random subgraph, KCenter, and whole graph).
As shown in \autoref{fig:cgl-all}, HyDRO outperforms both structure-based and structure-free methods on datasets such as Cora, Flickr, and Coauthor-Physics, and competes with methods like DM and GCDMX on Citeseer and Arxiv datasets, while surpassing all structure-based methods. 
These results highlight HyDRO's effectiveness in continual learning scenarios. 
Additional results and hyperparameter settings are provided in \autoref{app:cgl}.

\subsection{Privacy Preservation} 
\label{sec:Membership}
Condensed graphs may unintentionally reveal node membership through overfitting, making it essential to evaluate the resilience of graph distillation methods against such attacks. 
We use Membership Inference Attack (MIA) accuracy~\cite{duddu2020quantifying} (the lower the better) to assess an adversary's ability to determine whether a node belongs to the training set.
As shown in  \autoref{tab:pp}, although GDEM performs poorly on the original datasets, it achieves the lowest MIA accuracy on PubMed and Citeseer, but struggles on larger datasets like Flickr.
In contrast, HyDRO achieves the second-lowest MIA accuracy and the second-highest original accuracy on both datasets, with a 1.8\% reduction. 
It also obtains the highest original accuracy on Flickr with a 0.9\% reduction and comparable MIA accuracy. 
These results highlight HyDRO's robust privacy protection while effectively balancing utility and privacy.

\subsection{Denoising Ability}
\label{sec:noise}
Real-world graphs are prone to noise and adversarial attacks, it is thus necessary to evaluate the HyDRO's denosing ability.
Our experiment considers three types of noises: feature noise, structural noise, and adversarial structural noise.
Feature noise evaluates how well condensed graphs preserve meaningful information despite data corruption, while structural noise tests graph distillation methods' resilience to random edge additions. 
Adversarial structural noise, which is introduced by PR-BCD \cite{hussain2022adversarial}, employs PGD to iteratively perturb graph edges, maximizing damage to model performance. 
We evaluate the robustness of those distilled graphs by setting perturbation rates to 50\% for all the three noises.
The results in \autoref{tab:robustness} show that HyDRO achieves competitive accuracy across these three types of noise on the Flickr and PubMed datasets.
See \autoref{app:denoising} for more details.


\section{Conclusion}
We present \textbf{HyDRO}, a novel hyperbolic graph distillation method that leverages the graph spectral gap alignment to guide the optimization of graph features and structure embedding in Poincar\'e ball. 
Our experimental results show that the condensed graphs generated by our proposed approach lead to overall best performance on node classification tasks while excelling in generalization to link prediction tasks, outperforming all benchmark methods. 
HyDRO consistently maintains a top-2 position for in NAS and cross-architecture transferability evaluation, surpassing the state-of-the-arts in many cases. 
Moreover, by distilling graph random walk properties, the condensed graphs of HyDRO result in superior performance in continual graph learning compared to state-of-the-art methods.
Additionally, HyDRO proves to be effective in membership privacy preservation and resistance to noises during the distillation process.




\bibliographystyle{named}
\bibliography{Arxiv25}
\newpage
\appendix

\onecolumn

\section{Additional Preliminaries}
\label{app:preliminary}

\subsection{Distance and Parallel Transport}

The squared distance \( d^2(p_1, p_2) \) between two points \( p_1, p_2 \in \mathbb{B}^n_\kappa \) (the Poincaré ball model with curvature \( \kappa \)) is given by:  
\[
d^2(p_1, p_2) = \left( \tanh^{-1} \left( \sqrt{\kappa} \cdot \| \oplus_\kappa(-p_1, p_2) \| \right) \cdot \frac{2}{\sqrt{\kappa}} \right)^2,
\]
where \( \oplus_\kappa \) denotes the Möbius addition operation, defined as:  
\[
x \oplus_\kappa y = \frac{(1 + 2\kappa \langle x, y \rangle + \kappa \| y \|^2)x + (1 - \kappa \| x \|^2)y}{1 + 2\kappa \langle x, y \rangle + \kappa^2 \| x \|^2 \| y \|^2},
\]
and \( \tanh^{-1} \) is the inverse hyperbolic tangent function. Here, \( \|\cdot\| \) denotes the Euclidean norm, and \( \langle \cdot, \cdot \rangle \) denotes the Euclidean inner product.
Parallel transport is defined for moving a tangent vector \( \mathbf{u} \) from a point \( x \) to another point \( y \) on the manifold. It is given by:
\[
\operatorname{PT}_{x \to y}(\mathbf{u}) = F_{\operatorname{gyr}}[y, -x](\mathbf{u}) \cdot \frac{\lambda_x}{\lambda_y},
\]
where \( F_{\operatorname{gyr}}[y, -x](\mathbf{u}) \) is the gyration function, which accounts for the non-Euclidean geometry of the manifold. And \( \lambda_x = \frac{2}{1 - \kappa \| x \|^2} \) and \( \lambda_y = \frac{2}{1 - \kappa \| y \|^2} \) are conformal factors derived from the curvature \( \kappa \) and the Euclidean norms of \( x \) and \( y \).
These operations ensure consistency with the intrinsic curvature of the Poincaré ball model, enabling the preservation of geometric properties during vector transport.

\subsection{Exponential and Logarithmic Mapping}
In addition, the exponential and logarithmic maps are fundamental operations for the Poincaré ball model. The exponential map at point \( p \) in the direction of a tangent vector \( u \) is defined as 
\[
\exp_p(u) = F_{\text{mobius}}\left(p, \tanh\left( \frac{\sqrt{c}}{2} \lambda_p \|u\| \right) \cdot \frac{u}{\|u\|} \right),
\]
where \( p \) is the point on the Poincaré ball from which the exponential map originates and \( u \) is the tangent vector at \( p \), pointing in the direction along which the exponential map is applied. Additionally, \( c \) is the constant that determines the curvature of the space (usually related to the geometry of the hyperbolic space), and \( \|u\| \) is the norm (magnitude) of the tangent vector \( u \). \( \lambda_p \) is the scaling function at point \( p \), given by 
\[
\lambda_p = \frac{2}{1 - c \|p\|^2},
\]
where \( \|p\| \) is the norm of the point \( p \) in the Poincaré ball. This function ensures that the map remains within the unit ball.

The logarithmic map retrieves a tangent vector at point \( p_1 \) pointing towards \( p_2 \). It is defined as:
\[
\log_{p_1}(p_2) = \frac{2}{\sqrt{c} \lambda_{p_1}} \, \tan^{-1}\left( \sqrt{c} \cdot \|F_{\text{mobius}}(-p_1, p_2, c)\| \right) \cdot F_{\text{mobius}}(-p_1, p_2, c),
\]
where \( p_1 \) and \( p_2 \) are the two points in the Poincaré ball from which the logarithmic map is calculated. The term \( F_{\text{mobius}}(-p_1, p_2, c) \) denotes the Möbius addition of \( p_1 \) and \( p_2 \) in the Poincaré ball model, and is given by the formula:
\[
F_{\text{mobius}}(p_1, p_2, c) = \frac{(1 - c \|p_1\|^2) p_1 + (1 - c \|p_2\|^2) p_2}{1 - c \|p_1 - p_2\|^2},
\]
where \( \| \cdot \| \) represents the Euclidean norm (distance) between points, and the term \( c \) governs the curvature of the hyperbolic space.

Additionally, \( \tan^{-1} \) is the inverse hyperbolic tangent function, written as:
\[
\tan^{-1}(x) = \frac{1}{2} \ln\left( \frac{1+x}{1-x} \right),
\]
which is used to map the result from the Möbius addition into the tangent space. Furthermore, \( \lambda_{p_1} \) is the scaling function at point \( p_1 \), given by 
\[
\lambda_{p_1} = \frac{2}{1 - c \|p_1\|^2},
\]
where \( \|p_1\| \) is the norm of the point \( p_1 \) in the Poincaré ball, ensuring the map respects the curvature of the space. These calculations allow for the transformation between tangent spaces and the manifold itself, which is essential for computations in hyperbolic spaces like the Poincaré ball model.

\subsection{Möbius Addition}
Möbius addition is a fundamental operation in the Poincaré ball model, facilitating the combination of points within the hyperbolic space. The Möbius addition of two points \( x \) and \( y \) in the Poincaré ball is given by
\[
F_{\text{mobius}}(x, y, c) = \frac{(1 + 2c \langle x, y \rangle + c\|y\|^2)x + (1 - c\|x\|^2)y}{1 + 2c \langle x, y \rangle + c^2\|x\|^2\|y\|^2},
\]
where \( \langle x, y \rangle \) denotes the Euclidean inner product between \( x \) and \( y \), \( \| \cdot \| \) is the Euclidean norm, and \( c \) is the curvature constant.
This operation is critical for navigating the manifold and preserving its geometric structure. 
Additionally, a point in the Poincaré ball can be transformed into a hyperboloid representation as
\[
\text{to\_hyperboloid}(x, c) = \frac{1}{K - \|x\|^2} \begin{pmatrix} K + \|x\|^2 \\ 2\sqrt{K}x \end{pmatrix},
\]
where \( K = \frac{1}{c} \). This transformation is essential for visualizing points in hyperbolic space and understanding its geometry. The Poincaré ball manifold provides a rich structure for exploring geometric properties and optimization in hyperbolic spaces.

\subsection{Random Walks Properties}

The \textbf{spectral gap} of the random walk matrix provides insights into various properties of the graph related to random walks. Specifically, the \textbf{mixing time} \( t_{\text{mix}} \) represents the number of steps needed for the random walk to approach its stationary distribution within a specified accuracy. The mixing time is inversely related to the spectral gap \( \Delta \), with a larger spectral gap corresponding to a shorter mixing time, indicating quicker convergence to the stationary distribution $ t_{\text{mix}} \sim \frac{1}{\Delta}.$
The \textbf{total variation distance} \( d_{\text{TV}}(t) \) measures how close the distribution of the random walk is to the stationary distribution \( \pi \) at time \( t \). It is given by:
\[
d_{\text{TV}}(t) = \frac{1}{2} \sum_{i} \left| p_i(t) - \pi_i \right|,
\]
where \( p_i(t) \) is the probability of being at node \( i \) at time \( t \), and \( \pi_i \) is the stationary probability of node \( i \). A larger spectral gap leads to a smaller total variation distance, signifying faster convergence:
\[
d_{\text{TV}}(t) \sim \exp(-\Delta t).
\]
Additionally, \textbf{Cheeger's inequality} links the spectral gap to the graph's connectivity through conductance \( \Phi \), which quantifies the graph's bottleneck. Cheeger's inequality is given by:
\[
\frac{\Phi^2}{2} \leq \nu_2 \leq 2\Phi,
\]
where \( \Phi \) is the conductance and \( \nu_2 \) is the second-smallest eigenvalue of the normalized Laplacian \( L_{\text{norm}} \). The conductance \( \Phi \) is defined as:
\[
\Phi = \min_{S \subset V} \frac{\sum_{i \in S, j \notin S} W_{ij}}{\min(\text{vol}(S), \text{vol}(V \setminus S))},
\]
where \( S \) is a subset of nodes, \( W_{ij} \) represents the weight of the edge between nodes \( i \) and \( j \), and \( \text{vol}(S) \) denotes the volume of the set \( S \), which is the sum of the degrees of nodes in \( S \). This implies that higher conductance, indicating better connectivity, results in a larger spectral gap, leading to shorter mixing times and faster convergence of the random walk.
The spectral gap also has a direct relationship with the \textbf{commute time} \( \tau_{ij} \), which measures the expected number of steps for a random walk to travel from node \( i \) to node \( j \) and then back to \( i \). The commute time is given by:
\[
\tau_{ij} = \frac{2}{\pi_i \pi_j} \left( \frac{1}{\Delta} \right),
\]
where \( \pi_i \) and \( \pi_j \) are the stationary probabilities of nodes \( i \) and \( j \), respectively. The commute time is inversely related to the spectral gap, as a larger gap leads to faster convergence and thus shorter expected travel times between nodes.
In diffusion modeling, such as information spread, the spectral gap influences the efficiency of propagation. This is modeled by the Master Equation (or Kolmogorov Forward Equation), which governs the diffusion process with exponential waiting times. The equation is:
\[
\frac{d p_i(t)}{dt} = \sum_{j \neq i} \left[ W_{ij} p_j(t) - W_{ji} p_i(t) \right],
\]
where \( p_i(t) \) is the probability of being at node \( i \) at time \( t \), and \( W_{ij} \) is the transition rate between nodes \( i \) and \( j \). The spectral gap \( \Delta \) is related to the eigenvalues \( \lambda_1, \lambda_2, \dots \) of the transition matrix \( W \) governing the random walk. 
A larger spectral gap leads to faster diffusion, as it increases the difference between the largest eigenvalue \( \lambda_1 \) (which is zero for the stationary distribution) and the second-largest eigenvalue \( \lambda_2 \), thereby accelerating the random walk's convergence to its stationary distribution. The spectral gap also impacts the speed of information spread across the network, with a larger gap indicating quicker diffusion.
Thus, the spectral gap is a crucial measure for understanding the efficiency and convergence properties of random walks on a graph, influencing both mixing times and diffusion processes.

\section{Dataset Details.}
\label{app:dataset}
We conduct experiments using six transductive datasets: Cora, Citeseer, Pubmed, Arxiv, DBLP, Wiki-CS and Coauthor-Physics, along with two inductive datasets: Flickr and Reddit. And the Wiki-CS and Coauthor-Physics two datasets are only used for the continual graph learning tasks. Besides, the reduction rate, \( r \), is defined as the ratio of nodes in the condensed graph to the nodes in the training graph. For transductive datasets, the training graph encompasses the entire graph, while for inductive datasets, the training graph consists only of the training set. The statistics of datasets are detailed in \autoref{tab:dataset_statistics}.

\begin{table}[h]
  \centering
  \caption{Statistics of the datasets for evaluation.}
  \resizebox{0.65\linewidth}{!}{
    \begin{tabular}{lrrrrrrr}
    \toprule
    \textbf{Dataset} & \textbf{Nodes} & \textbf{Edges} & \textbf{Classes} & \textbf{Features} & \textbf{Training/Validation/Test} \\
    \midrule
    Cora & 2,708 & 5,429 & 7 & 1,433 & 140/500/1000 \\
    Citeseer & 3,327 & 4,732 & 6 & 3,703 & 120/500/1000 \\
    PubMed & 19,717 & 44,338 & 3 & 500 & 60/500/1,000 \\
    Ogbn-arxiv & 169,343 & 1,166,243 & 40 & 128 & 90,941/29,799/48,603 \\
    DBLP & 17,716 & 105,734 & 4 & 1639 & 14,172/1,772/1,772\\
    Wiki-CS & 11,701 &  216,123 & 10 & 300 & 7,021/2,340/2,340 \\
    Coauthor-Physics & 34,493 &  495,924 & 5 & 8,415 & 27,595/3,449/3,449 \\
    Flickr & 89,250 & 899,756 & 7 & 500 & 44,625/22,312/22,313 \\
    Reddit & 232,965 & 57, 307, 946 & 210 & 602 & 15,3932/23,699/55,334 \\

    \bottomrule
    \end{tabular}}
  \label{tab:dataset_statistics}
\end{table}

For the choice of reduction rate, we adjust it based on the labeling rates of the training graphs for each dataset. For \textit{Citeseer} and \textit{Cora}, which have relatively small labeling rates of 3.6\% and 5.2\%, respectively, we select reduction rates (\(r\)) as \{25\%, 50\%, 100\%\} of the labeling rate. For \textit{DBLP}, with labeling rates of 80\%, we choose the reduction rates (\(r\)) as \{0.1\%, 0.5\%, 1\%\} for DBLP. Since \textit{Arxiv} has a higher labeling rate of 53\%, we select reduction rates (\(r\)) as \{1\%, 5\%, 10\%\} of the labeling rate. Finally, for the inductive datasets (\textit{Flickr}, \textit{Reddit}), where all training graph nodes are labeled (100\% labeling rate), we choose reduction rates (\(r\)) as \{1\%, 5\%, 10\%\} of the labeling rate. The details of these choices are shown in \autoref{tab:reduction_rate_statistics}.

\section{Downstream tasks}
\label{app:hyperparameters}

\subsection{Node Classification}
\label{app:single}

For HyDRO, in addition to the hyperparameters mentioned in \autoref{sec:NC}, we also fine-tune several additional settings. Specifically, the number of epochs is tested in the set \{400, 600, 800, 1000\}, and the learning rates for the feature and topology structures in hyperbolic space are explored within \{0.1, 0.01, 0.001, 0.0001\}. Furthermore, the outer loop is tuned over the values \{25, 20, 15, 10\}, while the inner loop is tuned over \{15, 10, 5, 1\}. The learning rate for the SGC is fixed at 0.01. For other methods, we adhere to the hyperparameter settings provided in their respective papers or code repositories.

\subsection{Task Generalization}
\label{app:double}

To ensure a fair comparison, we fine-tuned the following hyperparameters to evaluate HyDRO against all benchmarks. 
The number of epochs was fixed at 600, and the condensed graphs were selected based on the validation performance. For all models, we fine-tuned the learning rate for both the feature and structure within the set \{0.1, 0.01, 0.001, 0.0001\}. 
Specifically, for datasets such as Cora, Citeseer, DBLP, Pubmed, and Wiki-CS, the outer loop was tuned over \{25, 20\} and the inner loop over \{15, 10\}. For larger datasets, such as ogbn-arxiv, Flickr, and Reddit, the outer loop was fixed at 10 and the inner loop at 1. 
For each dataset, we first identified the hyperparameters that resulted in the highest reduction rate. Once identified, these hyperparameters were applied consistently across all different reduction rates to ensure uniformity in the evaluation. 
For the special hyperparameters specific to GEOM, SFGC, and GCSNTK, we fine-tuned the common hyperparameters and adhered to the recommended settings for the special hyperparameters as outlined in the respective papers or source codes for each dataset.
For GDEM, we encountered an issue related to time constraints during the eigendecomposition phase. 
To address this, we optimized the code by improving the efficiency of the SVD process, making the GDEM available for a fair comparison. Moreover, by implementing these hyperparameter fine-tuning settings, we standardize the hyperparameter search and the graph condensation process across various tasks, including Neural Architecture Search (NAS), cross-architecture evaluations, Membership Inference Attack (MIA) tests, and denoising robustness assessments. This unified approach ensures consistency in model optimization and evaluation, making it easier to compare different tasks fairly under a common framework.

\section{Preservation of Random Walk Properties}
\label{app:Random}

\subsection{Commute Time Calculation}

The concept of \textit{Commute Time} in a graph represents the expected time it takes for a random walk to travel from one node to another and return \cite{qiu2007graph}. This is an important concept in spectral graph theory, which helps in analyzing various properties of graphs, including graph diffusion process and network connectivity.
We describe the mathematical process for computing the commute time \(CT(u, v)\) between any two nodes \(u\) and \(v\) in a graph. The calculation is based on the Green's function of the graph's Laplacian matrix. Let the graph be represented by a weighted adjacency matrix \(A\) of size \(n \times n\), where \(n\) is the number of nodes in the graph. The elements \(A_{ij}\) represent the weight (or connectivity) of the edge between node \(i\) and node \(j\), and \(A_{ij} = 0\) if no edge exists between them. 
The degree matrix \(D\) is a diagonal matrix where each diagonal element represents the degree of a corresponding node in the graph. The degree of a node is the sum of the weights of all edges connected to that node. 
The Laplacian matrix \(L\) is defined as the difference between the degree matrix \(D\) and the adjacency matrix \(A\).
And the Green's function \(L^+\) is calculated as the Moore-Penrose pseudoinverse of the Laplacian. 
This matrix plays a central role in calculating various random walk-related quantities such as hitting time and commute time.
The Green's function \(L^+\) is calculated as:
\[
L^+ = \text{pinv}(L),
\]
where \(\text{pinv}(L)\) is the pseudoinverse of \(L\). The Green's function contains information about the "resistance" or "conductance" between pairs of nodes in the graph and is used to calculate the time it takes for a random walker to move between nodes.

\paragraph{Hitting Time.}
The \textit{hitting time} \(Q(u, v)\) \cite{qiu2007graph} is the expected time it takes for a random walk starting at node \(u\) to reach node \(v\). It can be computed using the Green's function \(L^+\) as follows:
\[
Q(u, v) = \text{vol}(G) \cdot \left( G_{v,v} - G_{u,v} \right),
\]
where \(G_{v,v}\) and \(G_{u,v}\) are entries in the Green's function matrix \(L^+\), and \(\text{vol}(G)\) is the volume of the graph, which is the sum of the degrees of all nodes:
\[
\text{vol}(G) = \sum_{i=1}^n D_{ii}.
\]
The volume captures the total connectivity of the graph, and \(G_{v,v} - G_{u,v}\) represents the "distance" between nodes \(u\) and \(v\) in terms of the Green's function.

\paragraph{Commute Time.}
The \textit{commute time} \(CT(u, v)\) between two nodes \(u\) and \(v\) is the total expected time for a random walk to travel from node \(u\) to node \(v\) and then return to node \(u\). This is the sum of the hitting times in both directions:
\[
CT(u, v) = Q(u, v) + Q(v, u).
\]
Substituting the expressions for \(Q(u, v)\) and \(Q(v, u)\), we get:
\[
CT(u, v) = \text{vol}(G) \cdot \left( G_{v,v} - G_{u,v} \right) + \text{vol}(G) \cdot \left( G_{u,u} - G_{v,u} \right).
\]
The commute time is a key measure of the flow of information or "travel time" between nodes in a graph and is important for applications such as network optimization and graph diffusion processes.
And we visualized the commute times in heat maps, where each entry \( (i, j) \) represents the commute time \( CT(i, j) \) between nodes \( i \) and \( j \). This heatmap provides a clear representation of the flow of information between nodes and is useful for analyzing the structure of the graph.

\begin{figure*}[ht]
    \centering
   
    \begin{subfigure}[t]{0.2475\textwidth}
        \centering
        \includegraphics[width=\textwidth]{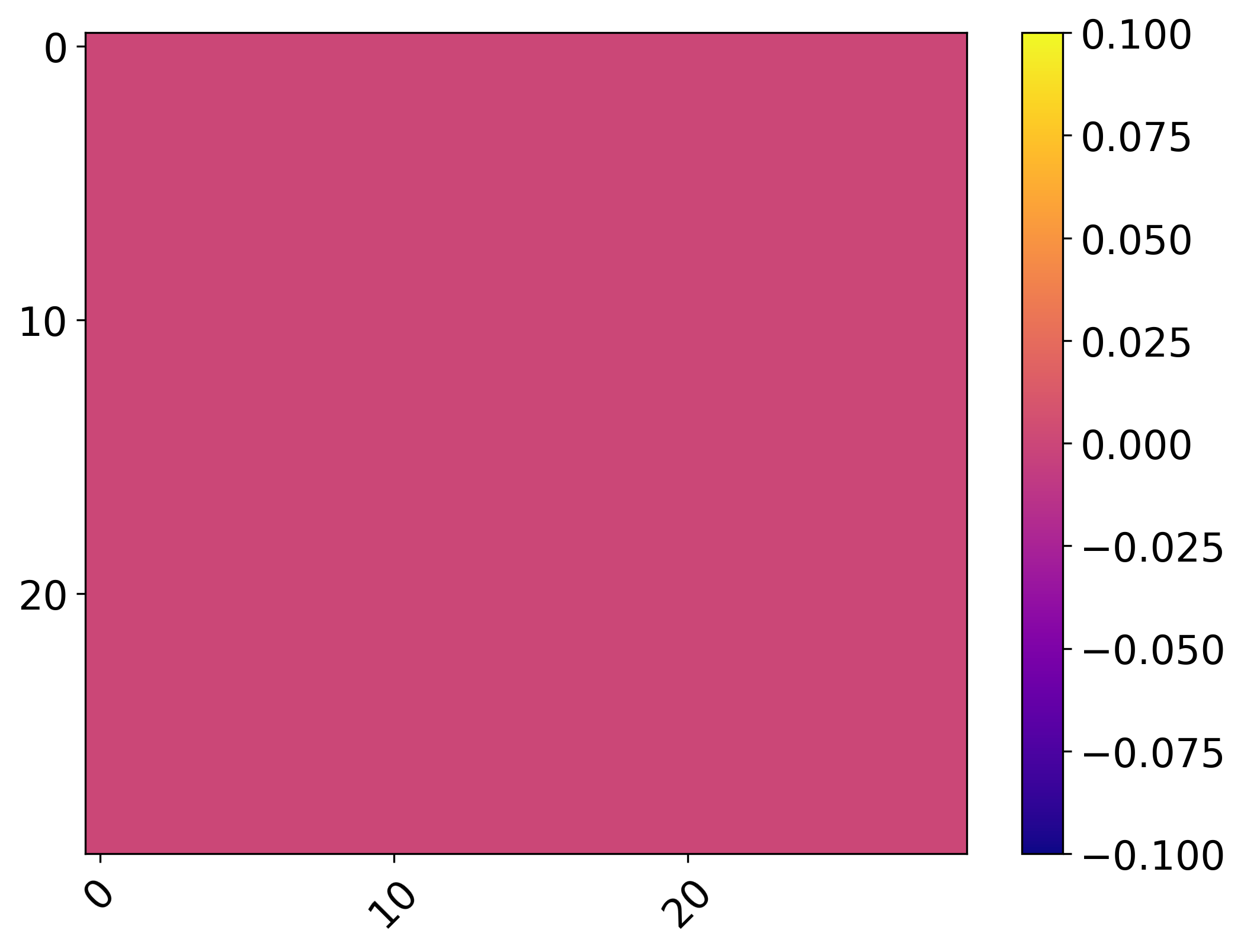}
        \caption{GEOM}  
        \label{fig:ct_original2}
    \end{subfigure}
    \begin{subfigure}[t]{0.2475\textwidth}
        \centering
        \includegraphics[width=\textwidth]{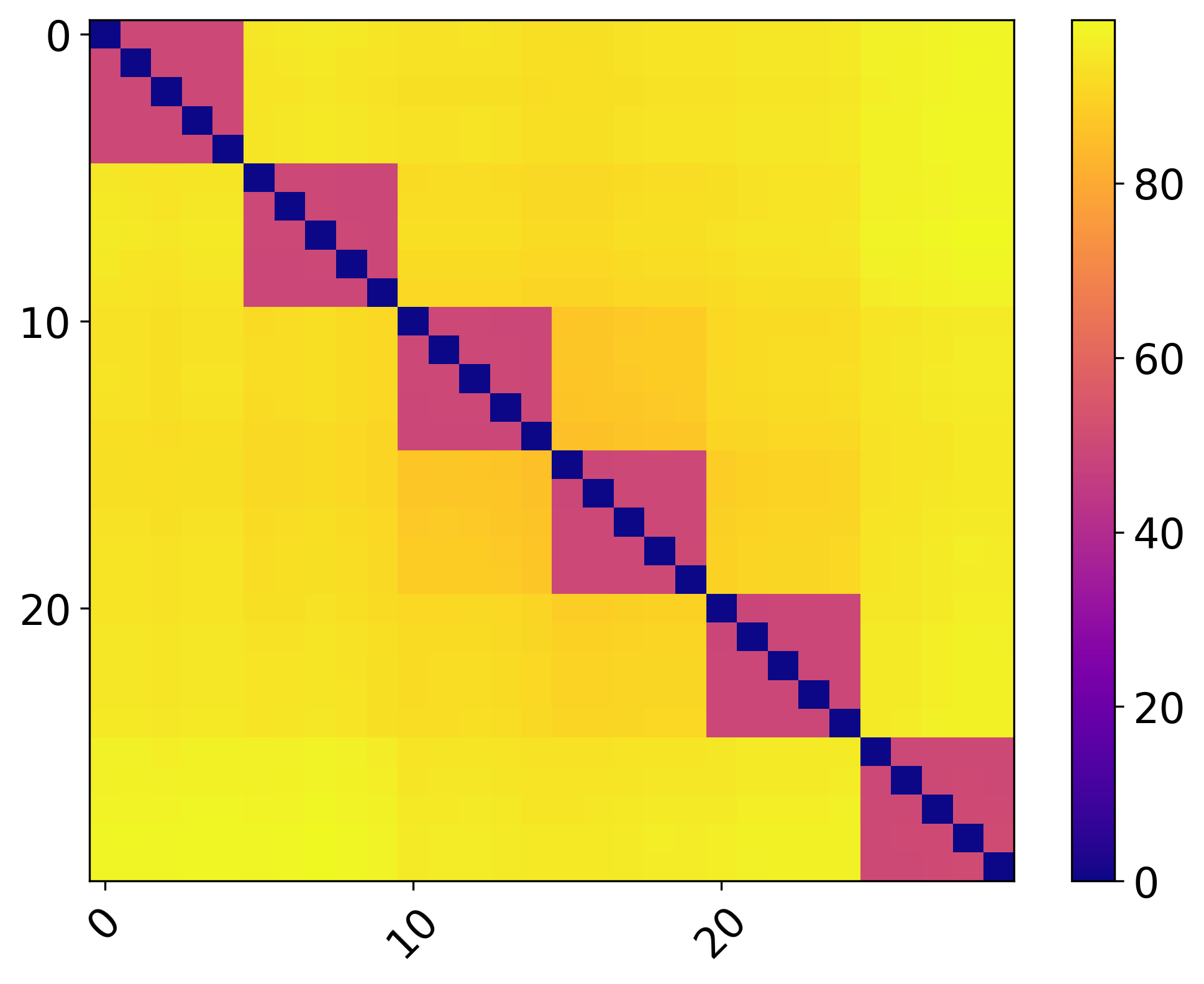}
        \caption{SGDD}  
        \label{fig:ct_hydro2}
    \end{subfigure}%
    \begin{subfigure}[t]{0.2475\textwidth}
        \centering
        \includegraphics[width=\textwidth]{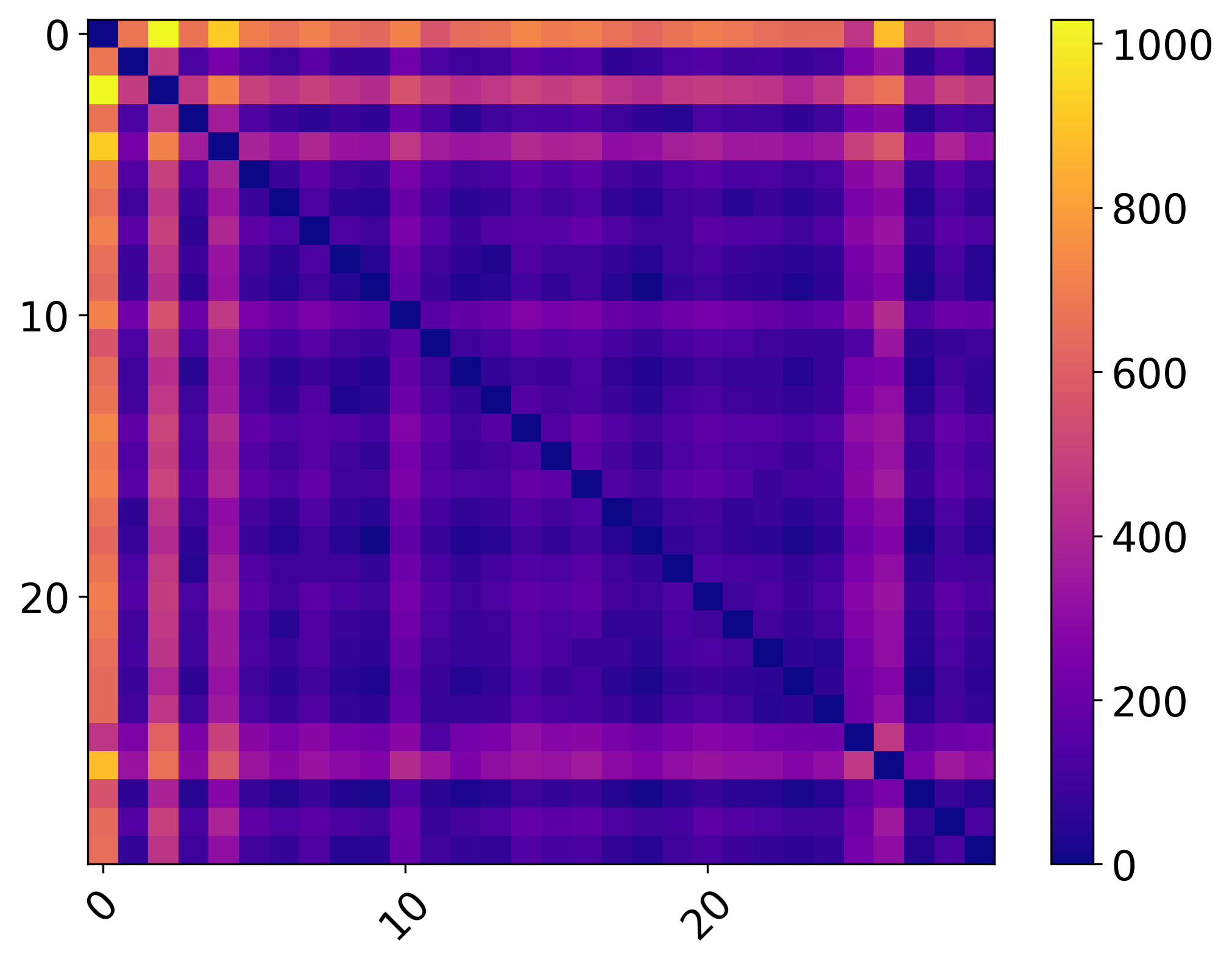}
        \caption{GDEM}  
        \label{fig:ct_sgdd2}
    \end{subfigure}%
    \begin{subfigure}[t]{0.2475\textwidth}
        \centering
        \includegraphics[width=\textwidth]{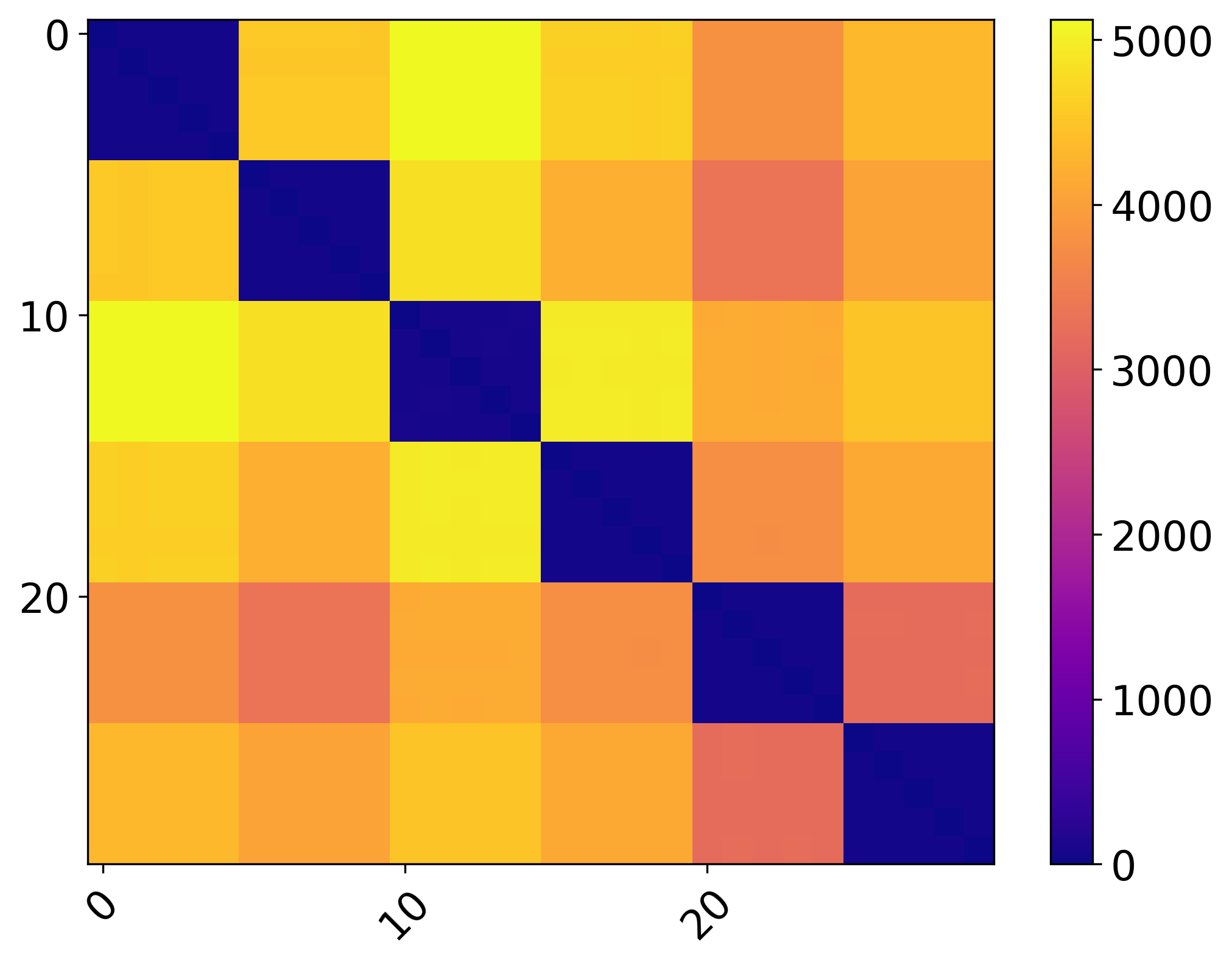}
        \caption{HyDRO}  
        \label{fig:ct_gdem2}
    \end{subfigure}%
    \caption{Commute time between pairs of nodes on Citeseer dataset (reduction rate 0.90\%). ({\bf Note}: {\it x} and {\it y}-axis denote the node ID. 
    The values closer to the original graph (\ref{fig:ct_original}) indicate better preservation of random walk properties.
    The color bars denote the value of commute time.)}
    \label{fig:commute2}
    
\end{figure*}

\begin{figure*}[ht]
    \centering
    
    \begin{subfigure}[t]{0.2475\textwidth}
        \centering
        \includegraphics[width=\textwidth]{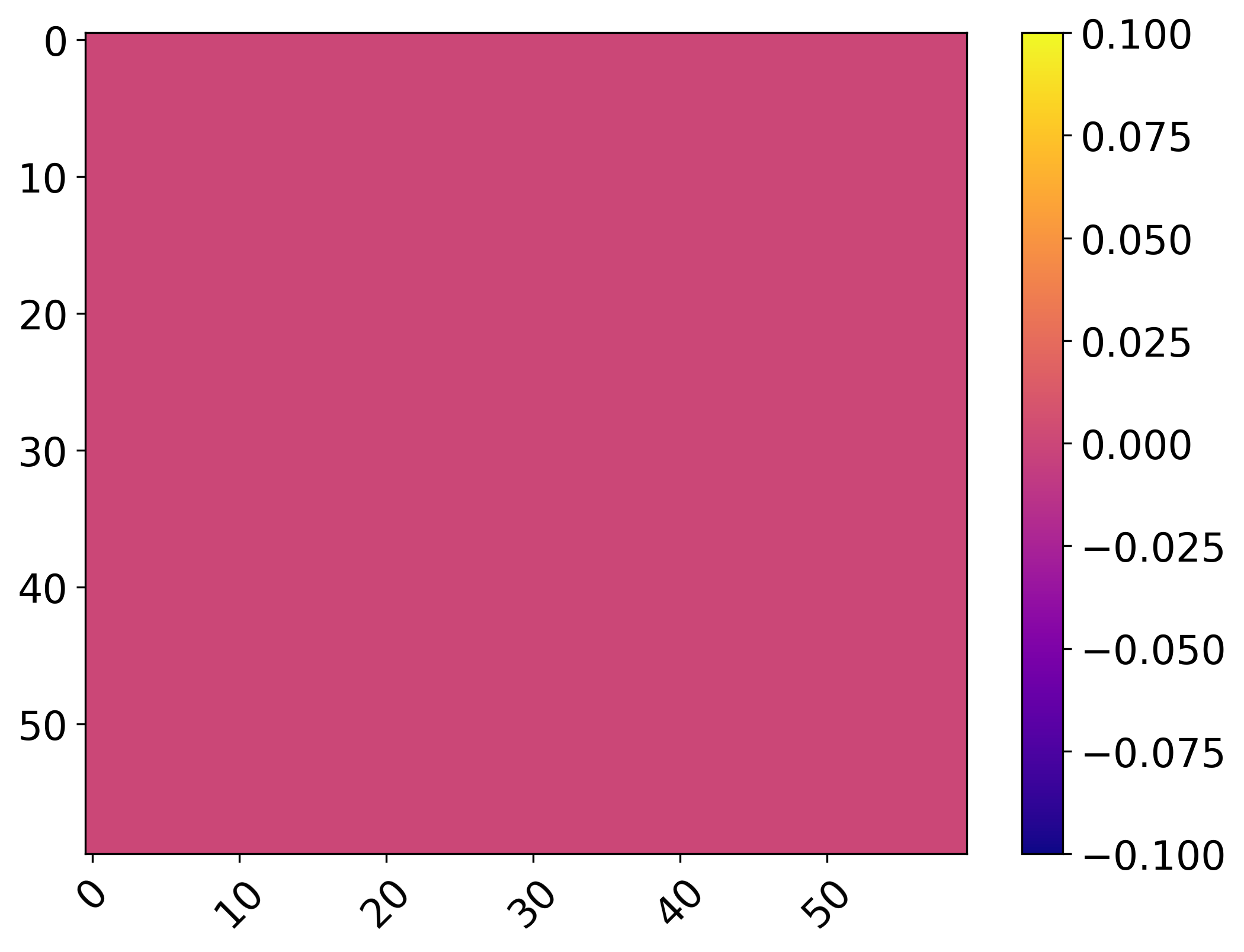}
        \caption{GEOM}  
        \label{fig:ct_original3}
    \end{subfigure}
    \begin{subfigure}[t]{0.2475\textwidth}
        \centering
        \includegraphics[width=\textwidth]{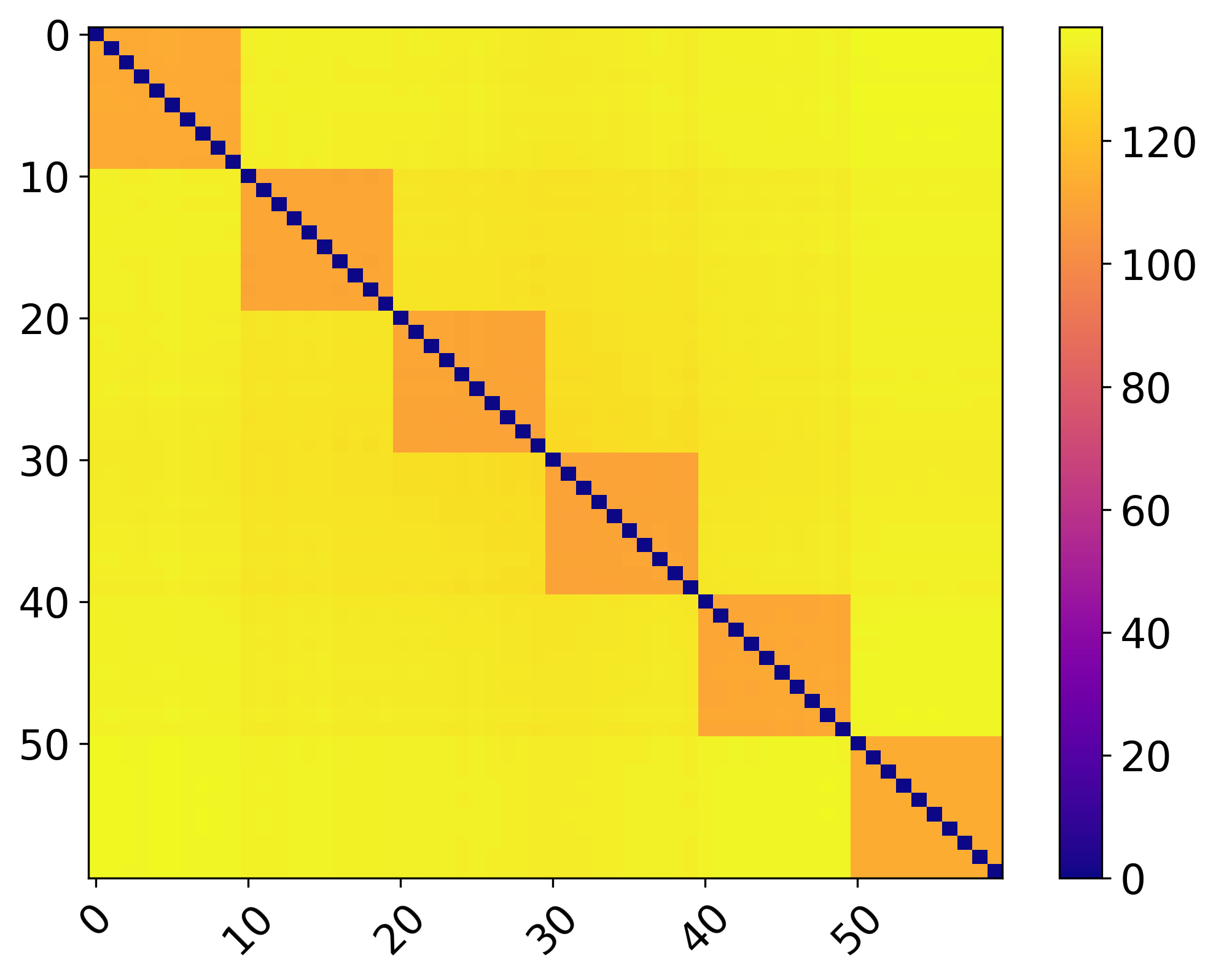}
        \caption{SGDD}  
        \label{fig:ct_hydro3}
    \end{subfigure}%
    \begin{subfigure}[t]{0.2475\textwidth}
        \centering
        \includegraphics[width=\textwidth]{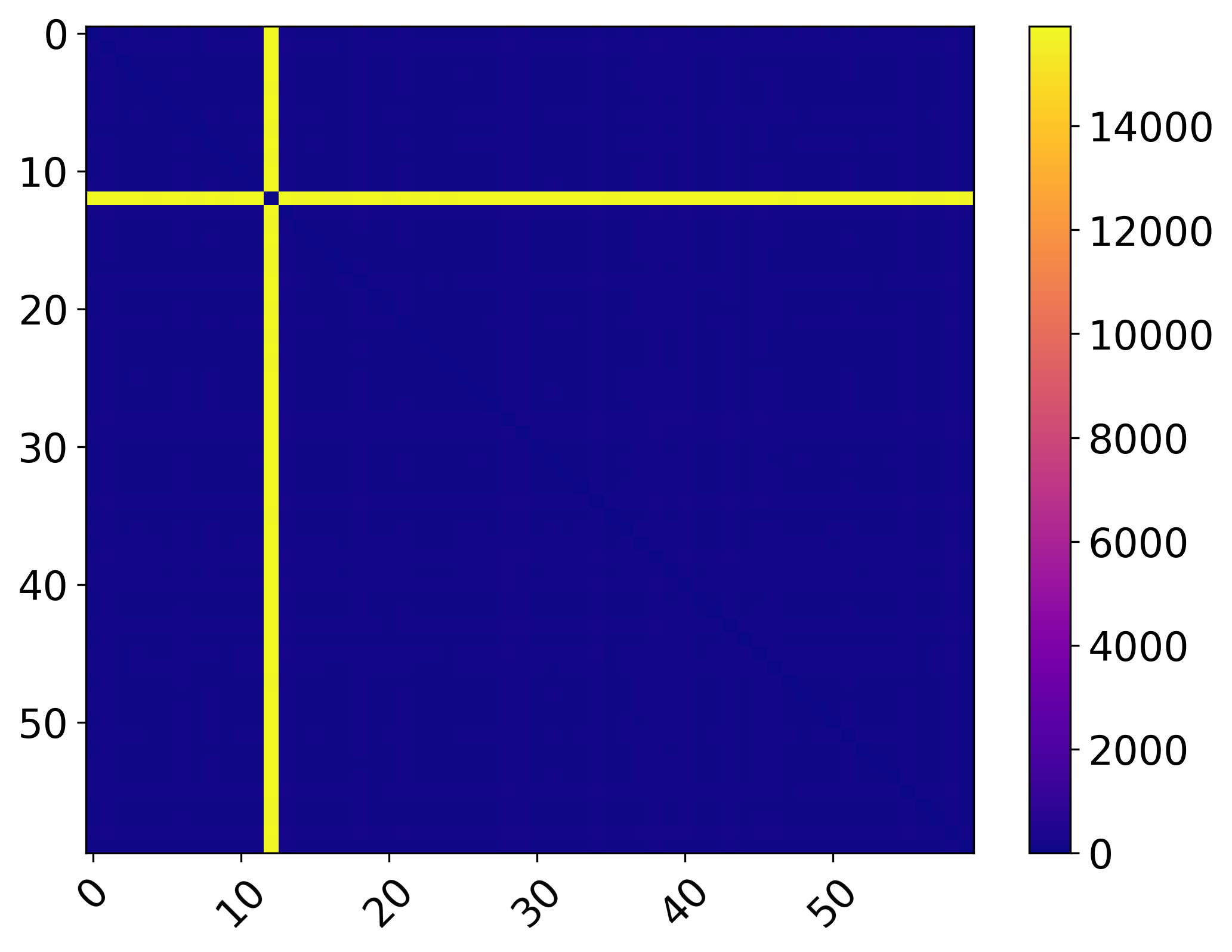}
        \caption{GDEM}  
        \label{fig:ct_sgdd3}
    \end{subfigure}%
    \begin{subfigure}[t]{0.2475\textwidth}
        \centering
        \includegraphics[width=\textwidth]{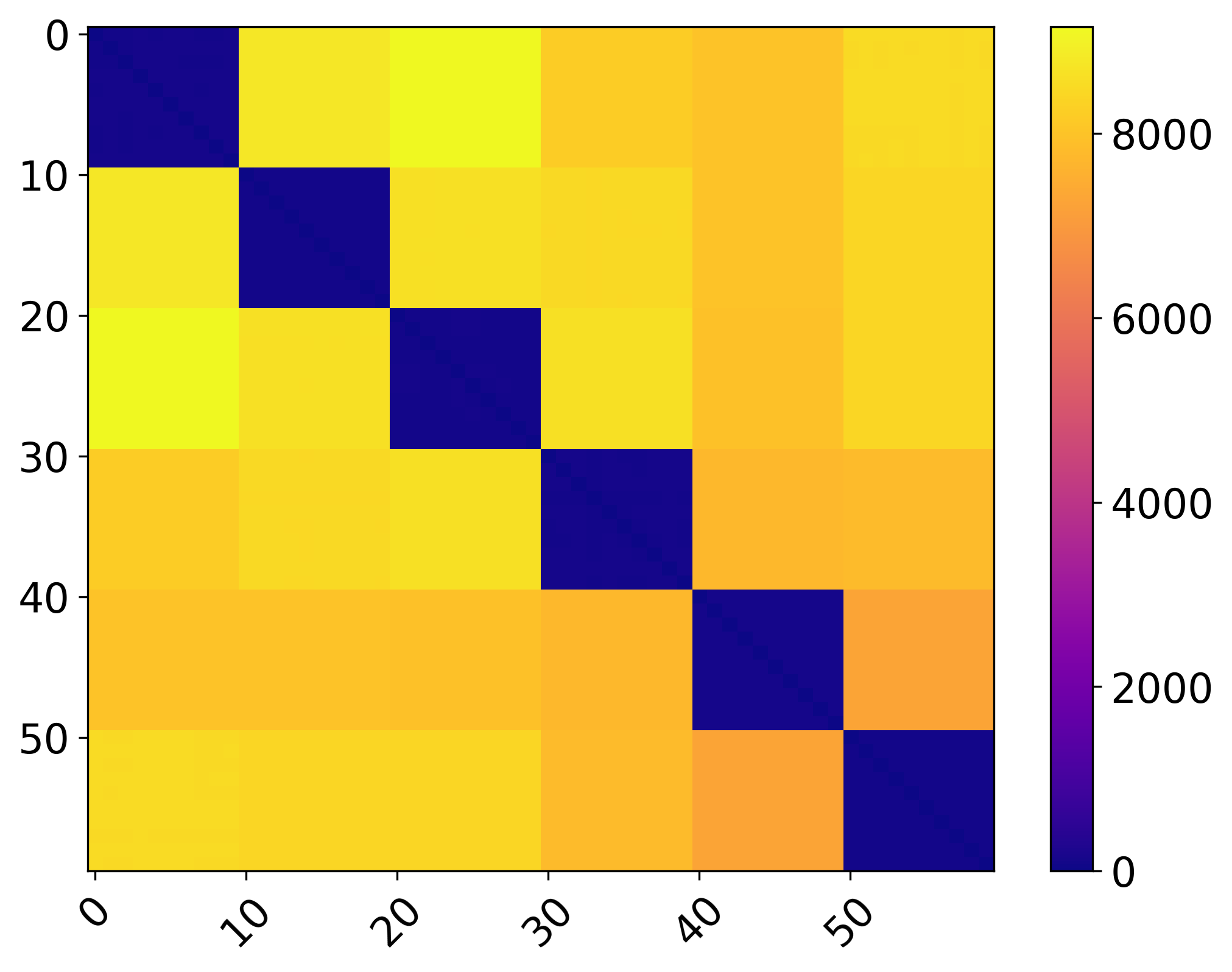}
        \caption{HyDRO}  
        \label{fig:ct_gdem3}
    \end{subfigure}%
    \caption{Commute time between pairs of nodes on Citeseer dataset (reduction rate 1.8\%). ({\bf Note}: {\it x} and {\it y}-axis denote the node ID. 
    The values closer to the original graph (\ref{fig:ct_original}) indicate better preservation of random walk properties.
    The color bars denote the value of commute time.)}
    \label{fig:commute3}
    
\end{figure*}

\begin{figure*}[ht]
    \centering
    \begin{subfigure}[t]{0.20\textwidth}
        \centering
        \includegraphics[width=\textwidth]{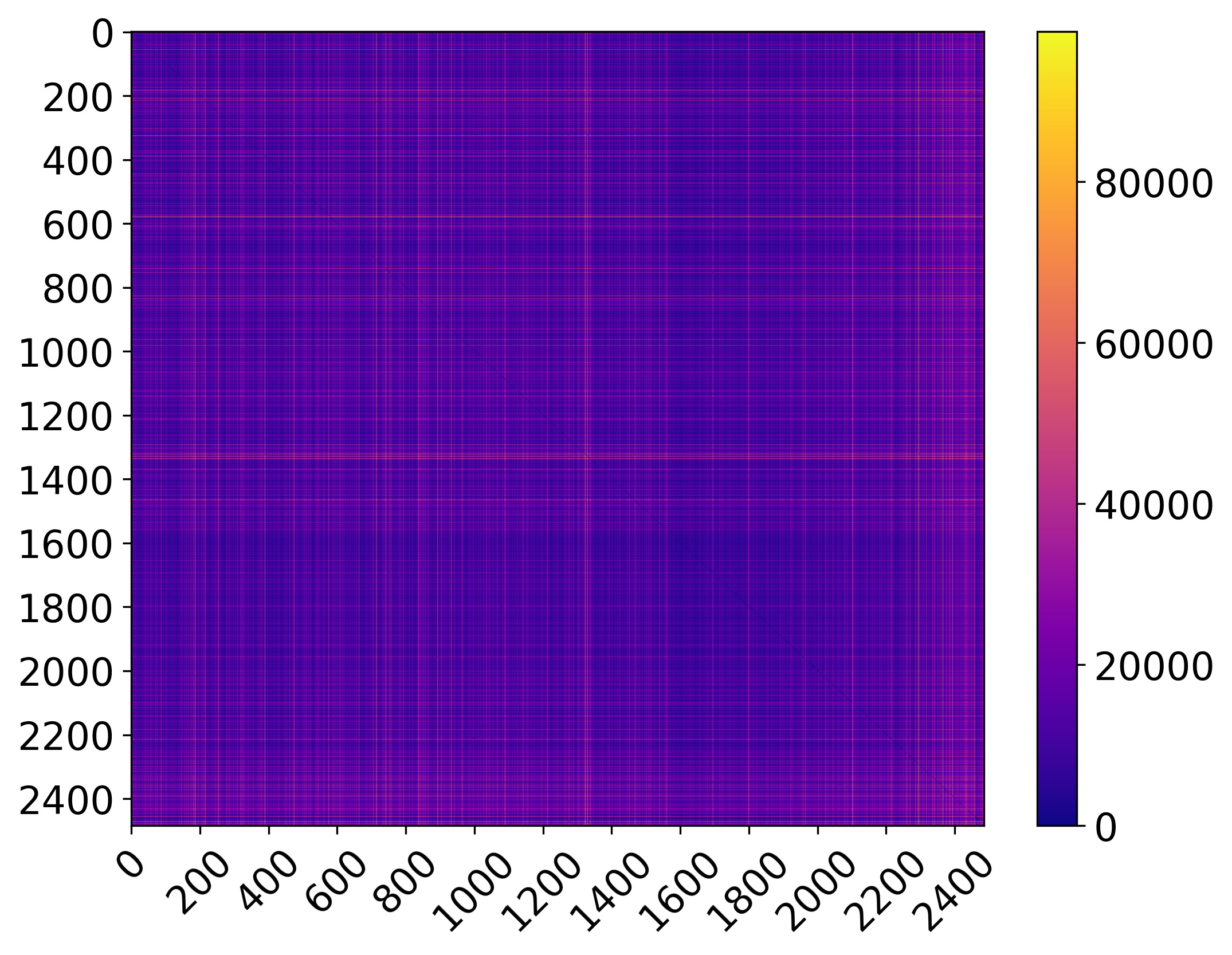}
        \caption{Original}  
        \label{fig:ct_original_cora}
    \end{subfigure}
    \begin{subfigure}[t]{0.20\textwidth}
        \centering
        \includegraphics[width=\textwidth]{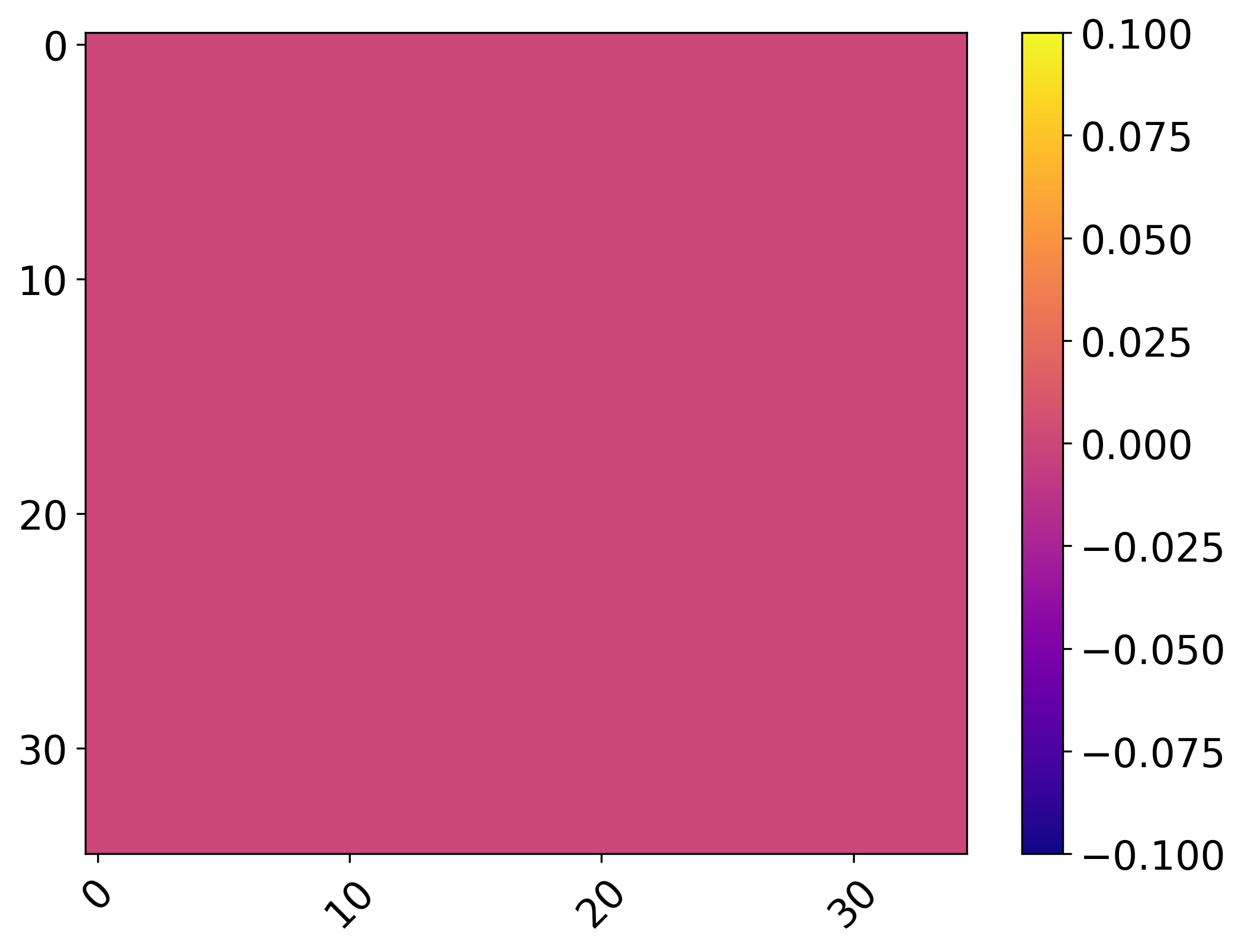}
        \caption{GEOM}  
        \label{fig:ct_original3}
    \end{subfigure}
    \begin{subfigure}[t]{0.20\textwidth}
        \centering
        \includegraphics[width=\textwidth]{figs/commute_time_Citeseer0.25_SGDD.png}
        \caption{SGDD}  
        \label{fig:ct_hydro3}
    \end{subfigure}%
    \begin{subfigure}[t]{0.20\textwidth}
        \centering
        \includegraphics[width=\textwidth]{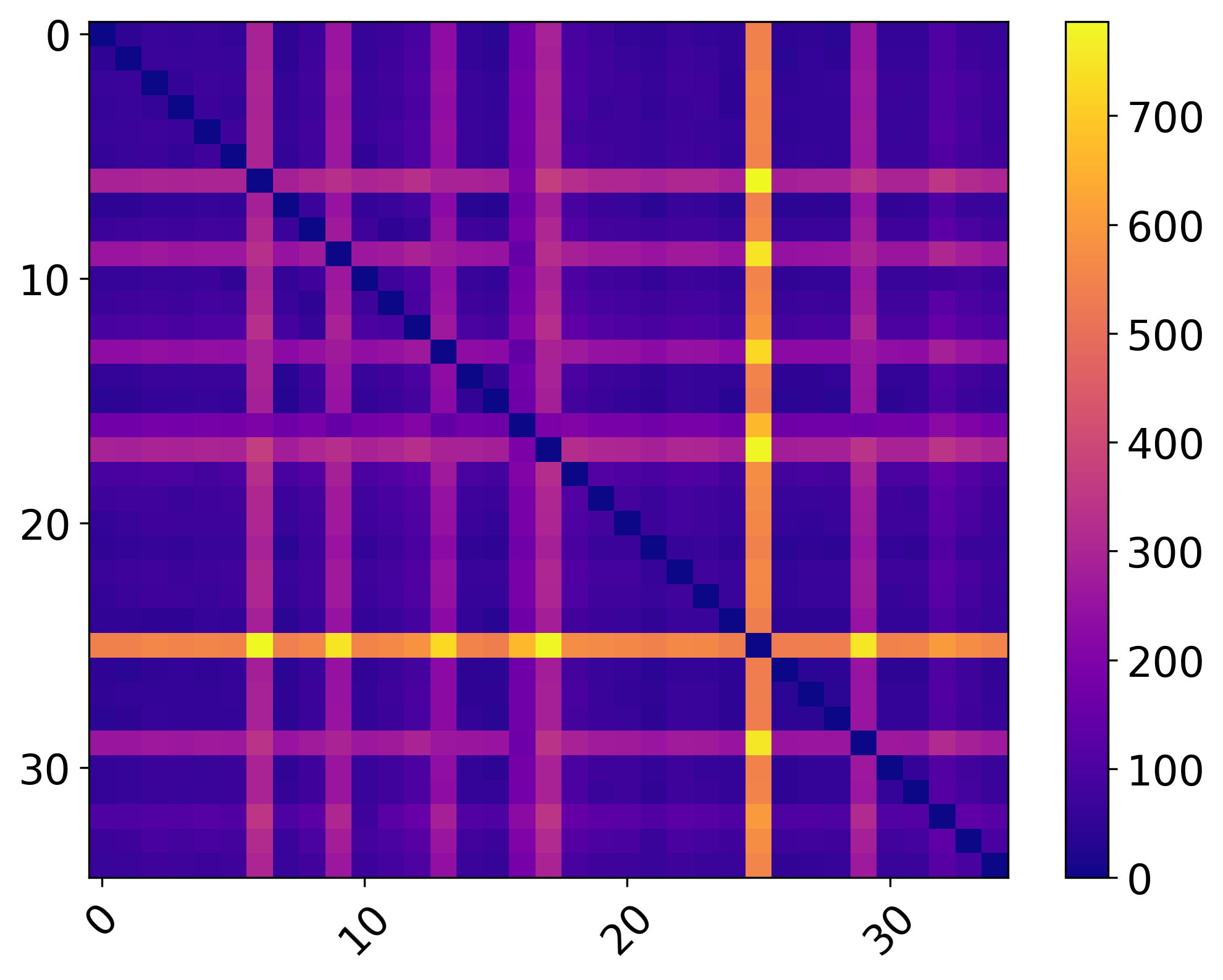}
        \caption{GDEM}  
        \label{fig:ct_sgdd3}
    \end{subfigure}%
    \begin{subfigure}[t]{0.20\textwidth}
        \centering
        \includegraphics[width=\textwidth]{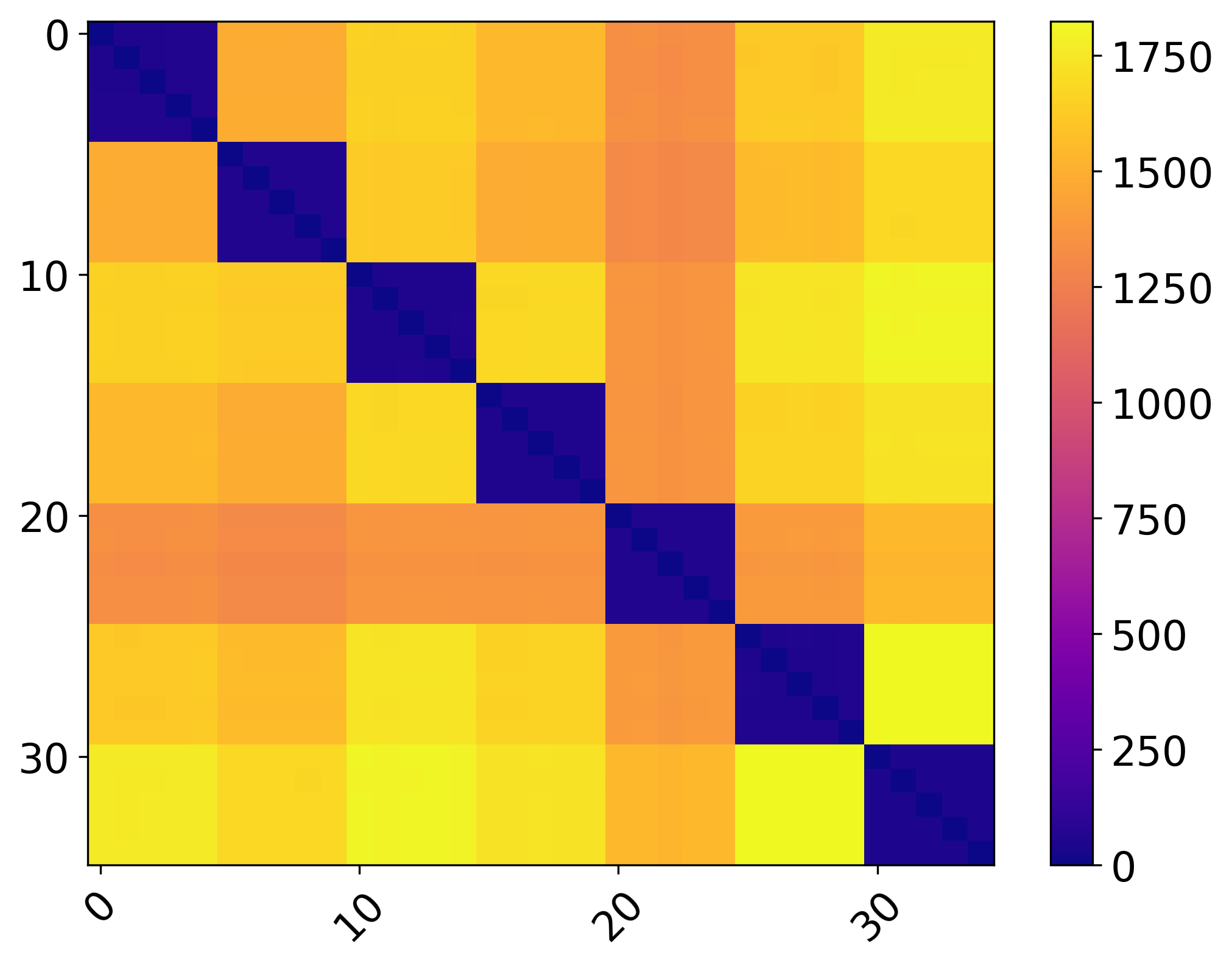}
        \caption{HyDRO}  
        \label{fig:ct_gdem3}
    \end{subfigure}%
    \caption{Commute time between pairs of nodes on Cora dataset (reduction rate 0.5\%). ({\bf Note}: {\it x} and {\it y}-axis denote the node ID. 
    The values closer to the original graph (\ref{fig:ct_original_cora}) without filtering indicate better preservation of random walk properties.
    The color bars denote the value of commute time.)}
    \label{fig:commute4}
    
\end{figure*}

\begin{figure*}[ht]
    \centering
    \begin{subfigure}[t]{0.20\textwidth}
        \centering
        \includegraphics[width=\textwidth]{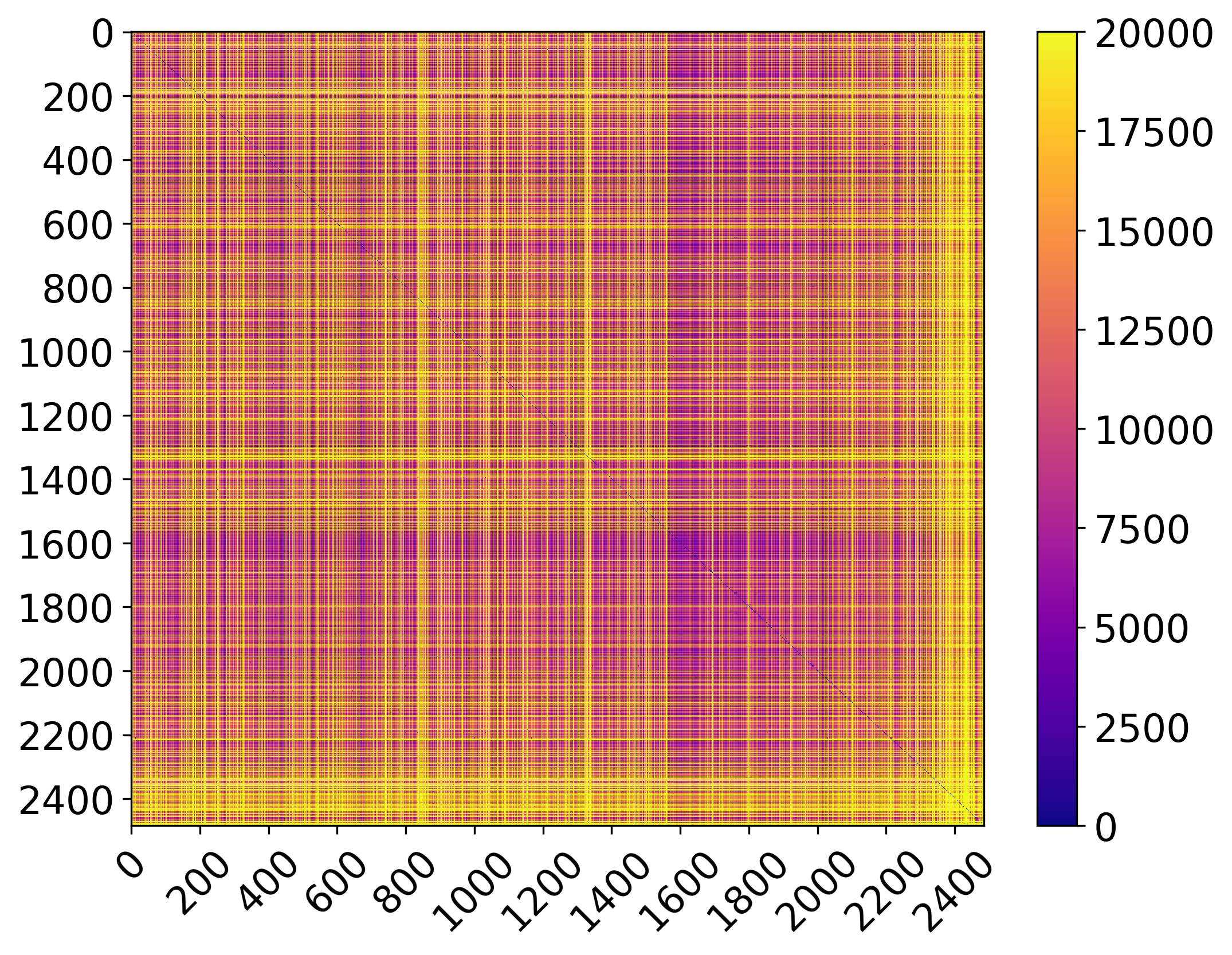}
        \caption{Original}  
        \label{fig:ct_original3_cora2}
    \end{subfigure}
    \begin{subfigure}[t]{0.20\textwidth}
        \centering
        \includegraphics[width=\textwidth]{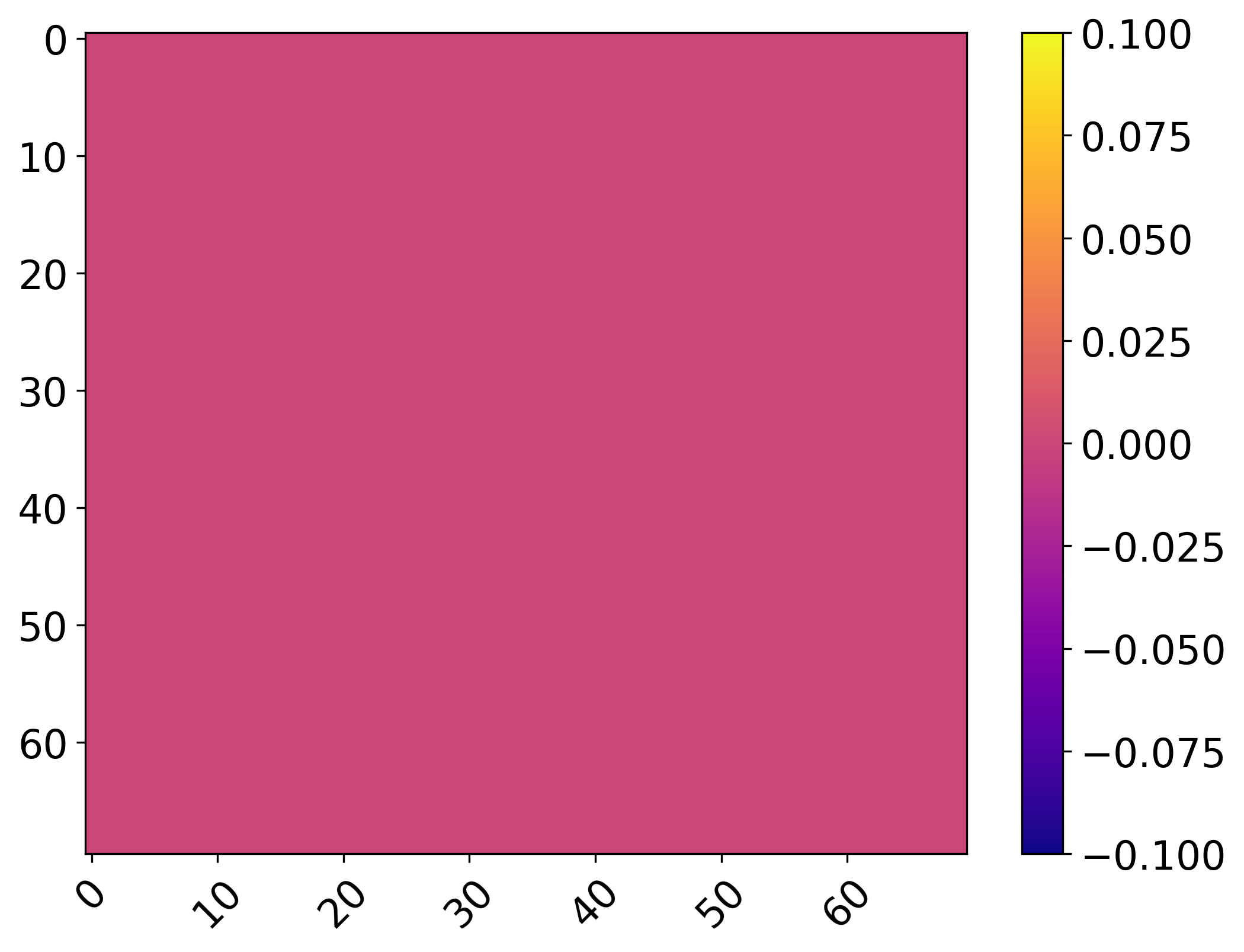}
        \caption{GEOM}  
        \label{fig:ct_original3}
    \end{subfigure}
    \begin{subfigure}[t]{0.20\textwidth}
        \centering
        \includegraphics[width=\textwidth]{figs/commute_time_Citeseer0.5_SGDD.png}
        \caption{SGDD}  
        \label{fig:ct_hydro3}
    \end{subfigure}%
    \begin{subfigure}[t]{0.20\textwidth}
        \centering
        \includegraphics[width=\textwidth]{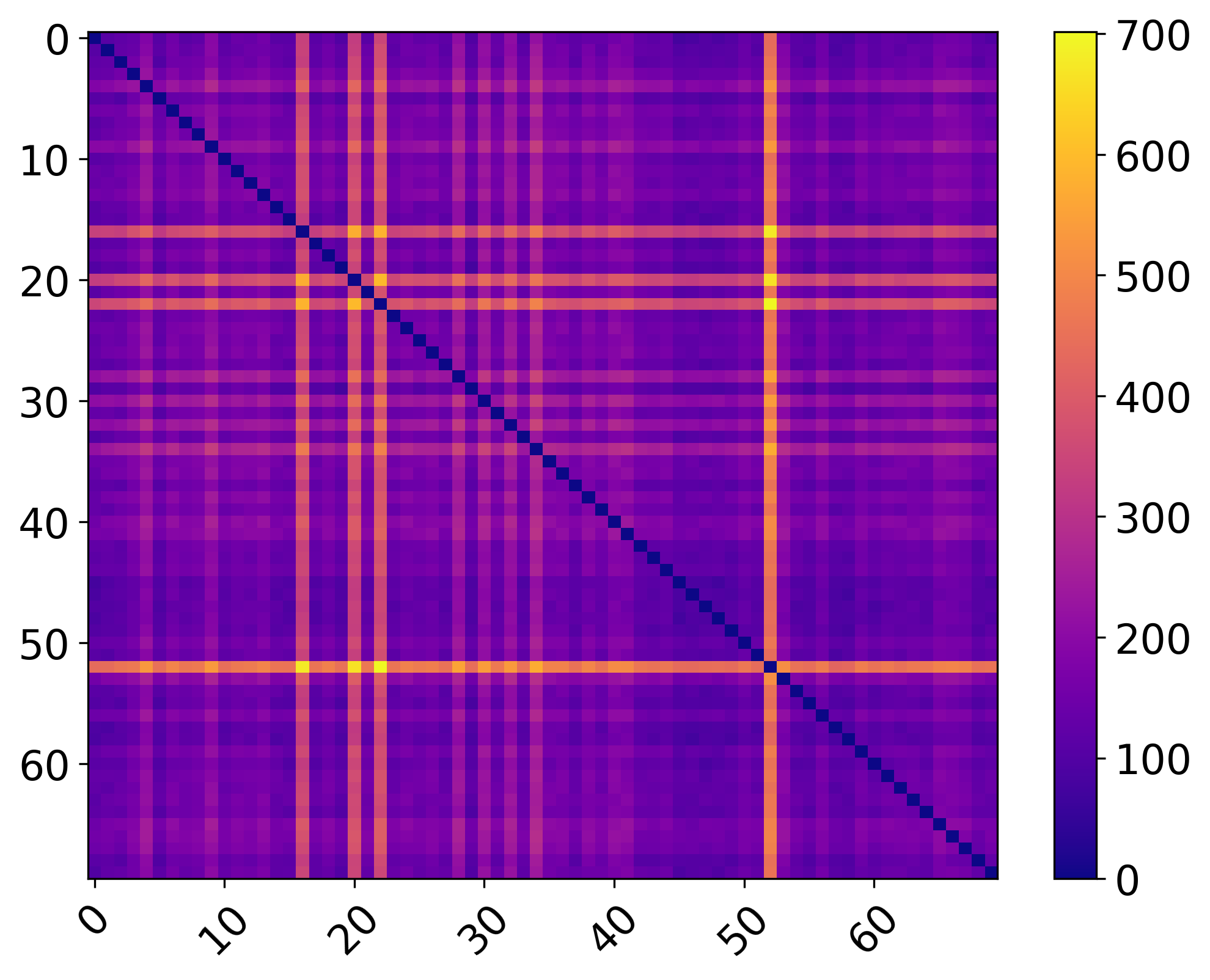}
        \caption{GDEM}  
        \label{fig:ct_sgdd3}
    \end{subfigure}%
    \begin{subfigure}[t]{0.20\textwidth}
        \centering
        \includegraphics[width=\textwidth]{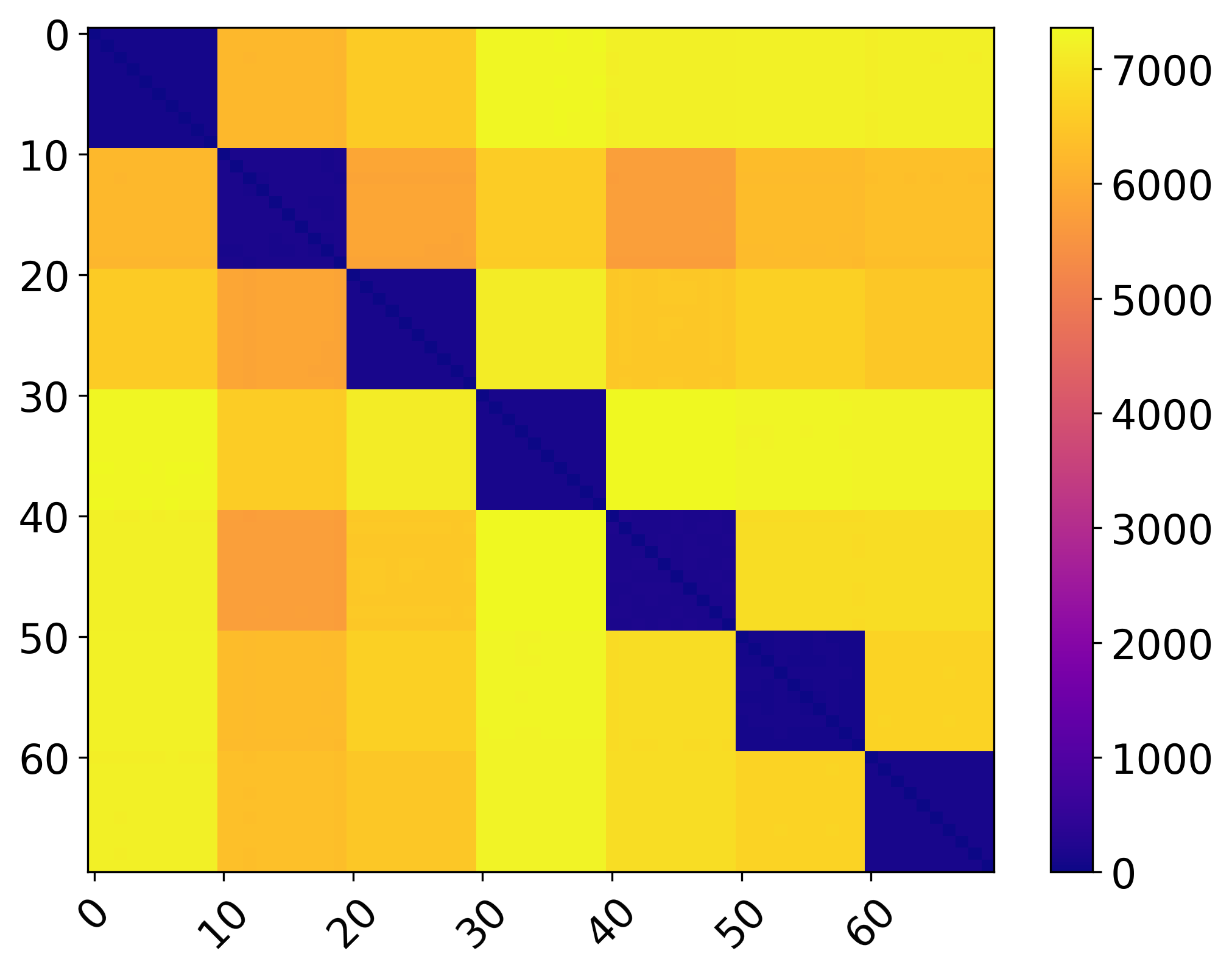}
        \caption{HyDRO}  
        \label{fig:ct_gdem3}
    \end{subfigure}%
    \caption{Commute time between pairs of nodes on Citeseer dataset (reduction rate 1.3\%). ({\bf Note}: {\it x} and {\it y}-axis denote the node ID. 
    The values closer to the original graph (\ref{fig:ct_original3_cora2}) with fitering indicate better preservation of random walk properties.
    The color bars denote the value of commute time.)}
    \label{fig:commute5}
    
\end{figure*}

\begin{figure*}[ht]
    \centering
    \begin{subfigure}[t]{0.20\textwidth}
        \centering
        \includegraphics[width=\textwidth]{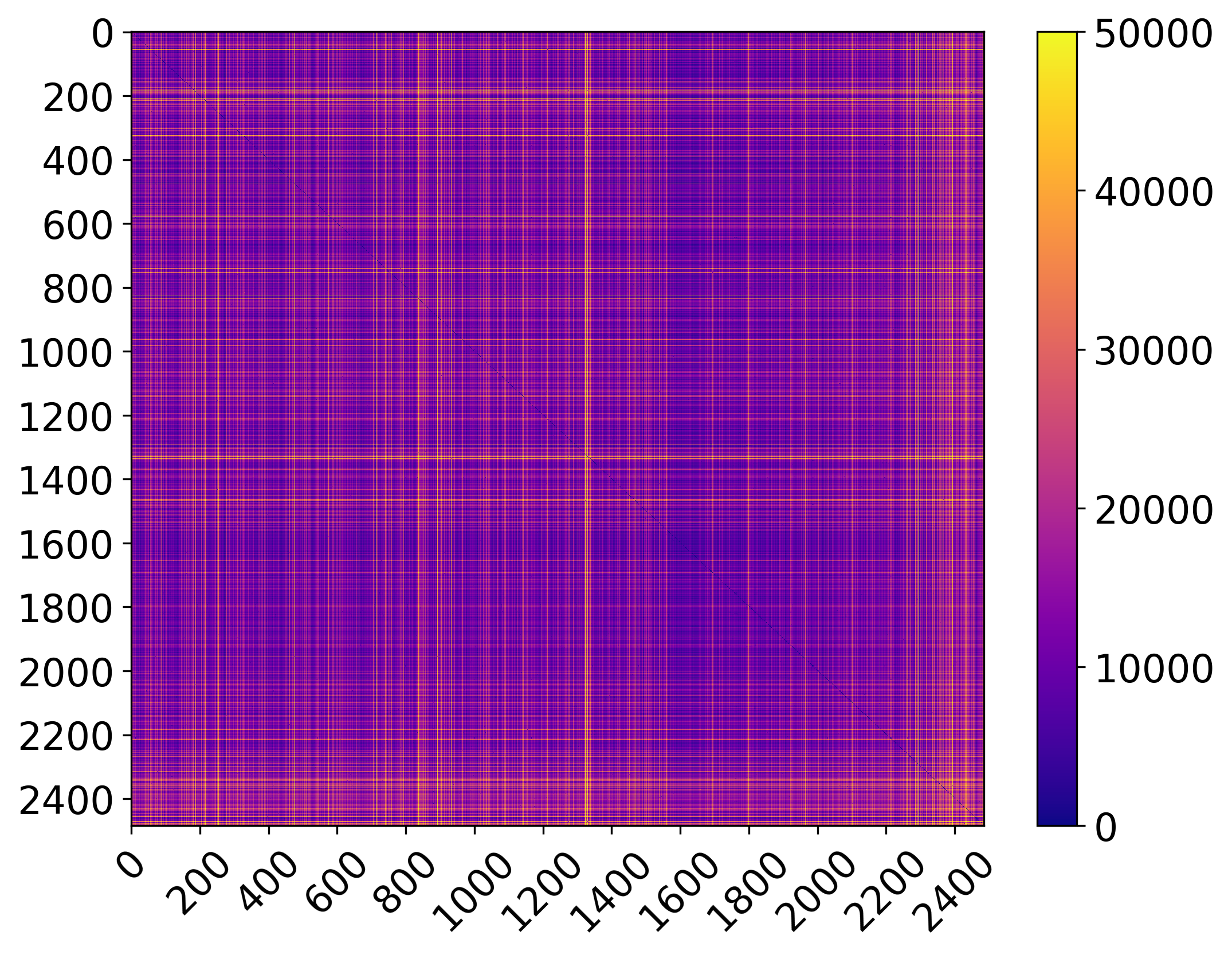}
        \caption{Original}  
        \label{fig:ct_original3_cora3}
    \end{subfigure}
    \begin{subfigure}[t]{0.20\textwidth}
        \centering
        \includegraphics[width=\textwidth]{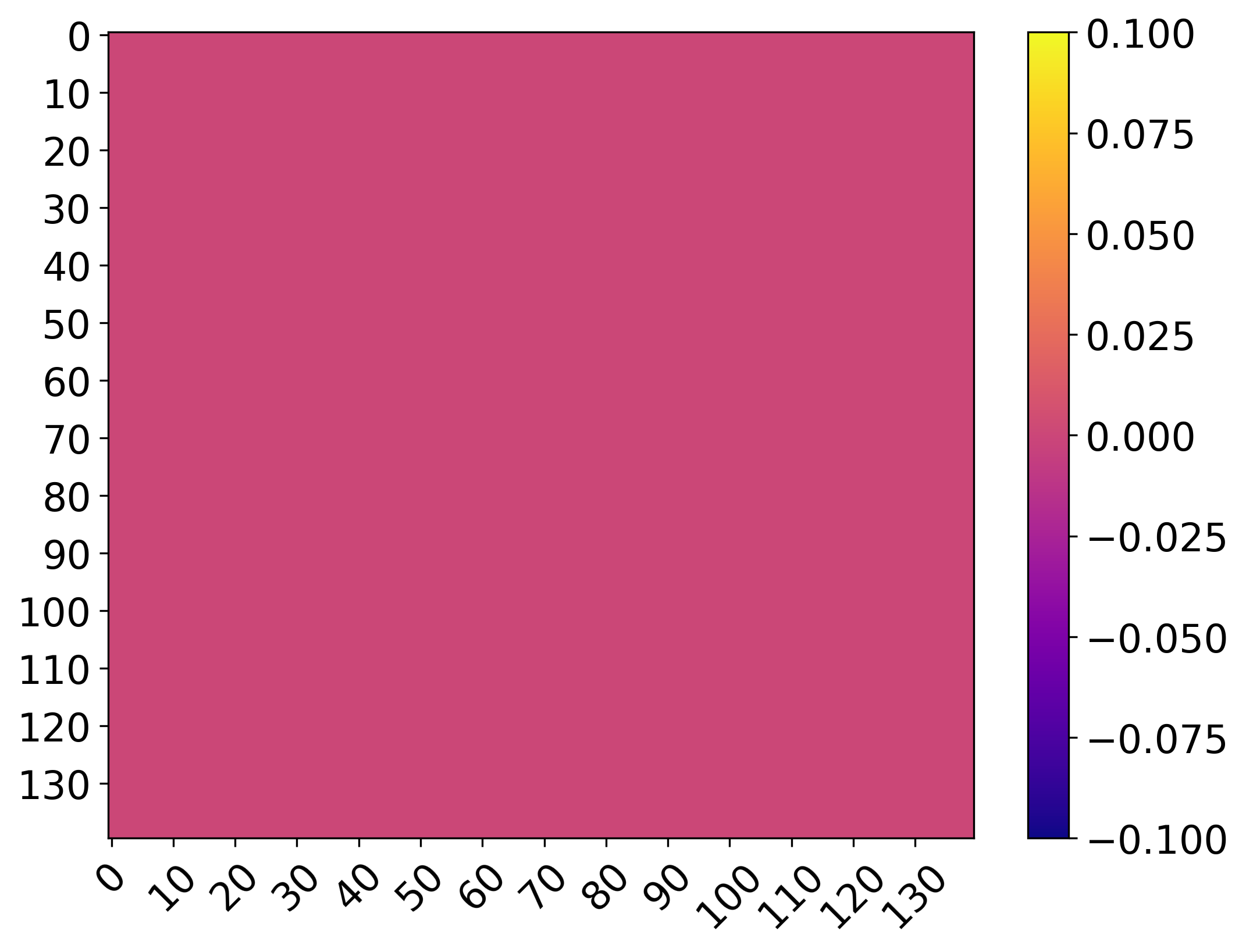}
        \caption{GEOM}  
        \label{fig:ct_original3}
    \end{subfigure}
    \begin{subfigure}[t]{0.20\textwidth}
        \centering
        \includegraphics[width=\textwidth]{figs/commute_time_Citeseer1_SGDD.png}
        \caption{SGDD}  
        \label{fig:ct_hydro3}
    \end{subfigure}%
    \begin{subfigure}[t]{0.20\textwidth}
        \centering
        \includegraphics[width=\textwidth]{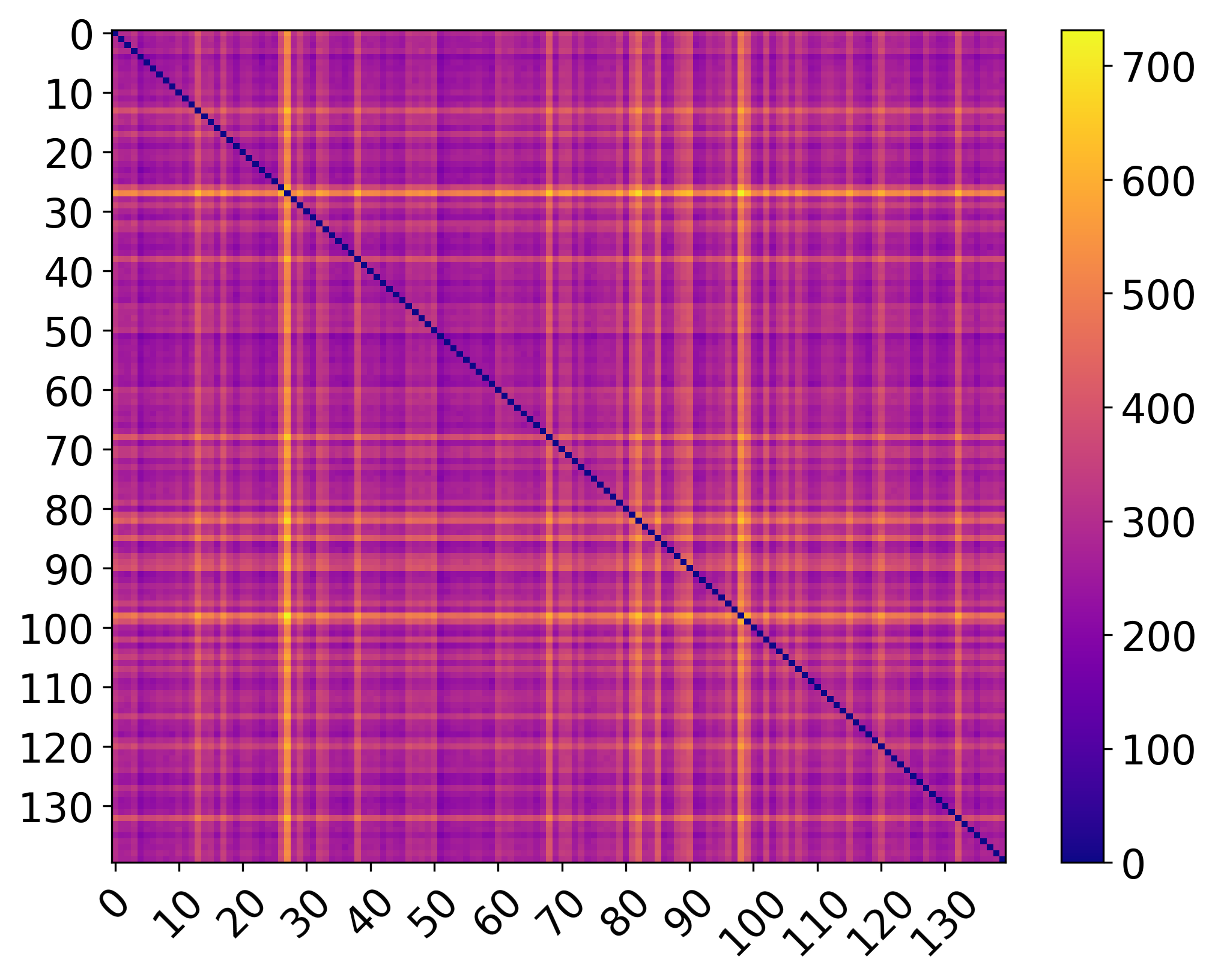}
        \caption{GDEM}  
        \label{fig:ct_sgdd3}
    \end{subfigure}%
    \begin{subfigure}[t]{0.20\textwidth}
        \centering
        \includegraphics[width=\textwidth]{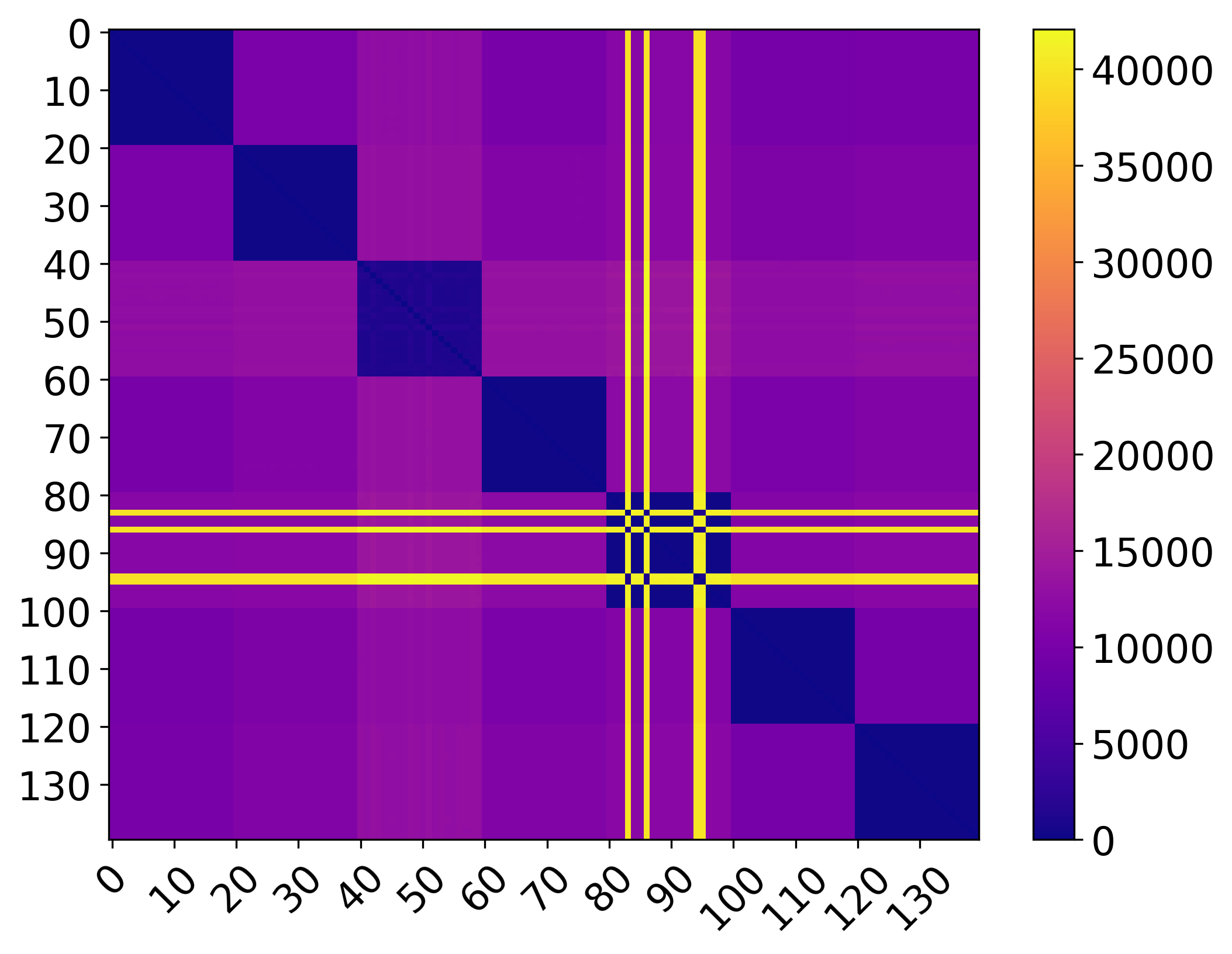}
        \caption{HyDRO}  
        \label{fig:ct_gdem3}
    \end{subfigure}%
    \caption{Commute time between pairs of nodes on Citeseer dataset (reduction rate 2.6\%). ({\bf Note}: {\it x} and {\it y}-axis denote the node ID. 
    The values closer to the original graph (\ref{fig:ct_original3_cora3}) with filtering indicate better preservation of random walk properties.
    The color bars denote the value of commute time.)}
    \label{fig:commute6}
    
\end{figure*}

To reduce randomness, we also compute the commute time matrix on the Citeseer and Cora dataset across all reduction rates. Similar to the approach in \autoref{sec:random}, we set a maximum limit of 20,000 for the commute time values in the original matrix. This ensures a better comparison of the values distribution in the matrix , minimizing the impact of a few extremely high values. And we present the commute time matrix in \autoref{fig:commute2} and \autoref{fig:commute3}. It is apparent that GEOM and SGDD fail to capture sufficient information about the commute time from the original graphs. For GDEM, while it can capture the commute time for a few pairs of nodes, there remains a significant gap in capturing the overall distribution of values in the original graphs. In contrast, HyDRO consistently captures the main distribution of the commute time matrix for most node pairs, preserving the commute time matrix in a way that is closer to the original graph across different reduction rates. For the Cora dataset shown in \autoref{fig:commute4}, \autoref{fig:commute5}, and \autoref{fig:commute6}
, similarly, HyDRO consistently performs much better than other benchmarks in preserving the commute time values from the original graphs, especially for larger condensed graphs, such as \( r = 3.6\% \) in Cora or \( r = 5.2\% \) in Citeseer.

\subsection{Flow Distance Calculation}
\begin{figure*}[t]
    \centering
    \small
    \begin{subfigure}[t]{0.33\textwidth}
        \centering
        \includegraphics[width=\textwidth]{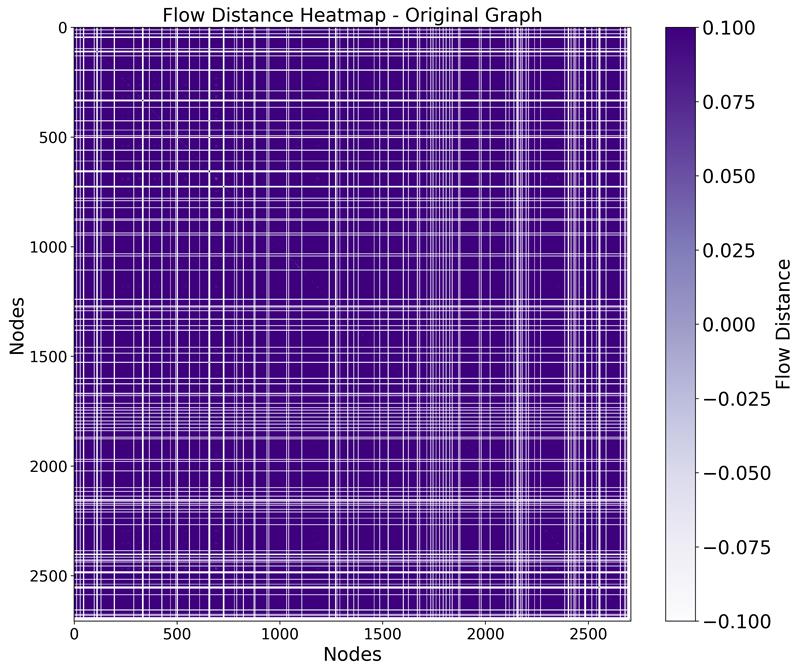}
        \caption{Original Cora}
        \label{fig:real_original} 
    \end{subfigure}%
    \begin{subfigure}[t]{0.33\textwidth}
        \centering
        \includegraphics[width=\textwidth]{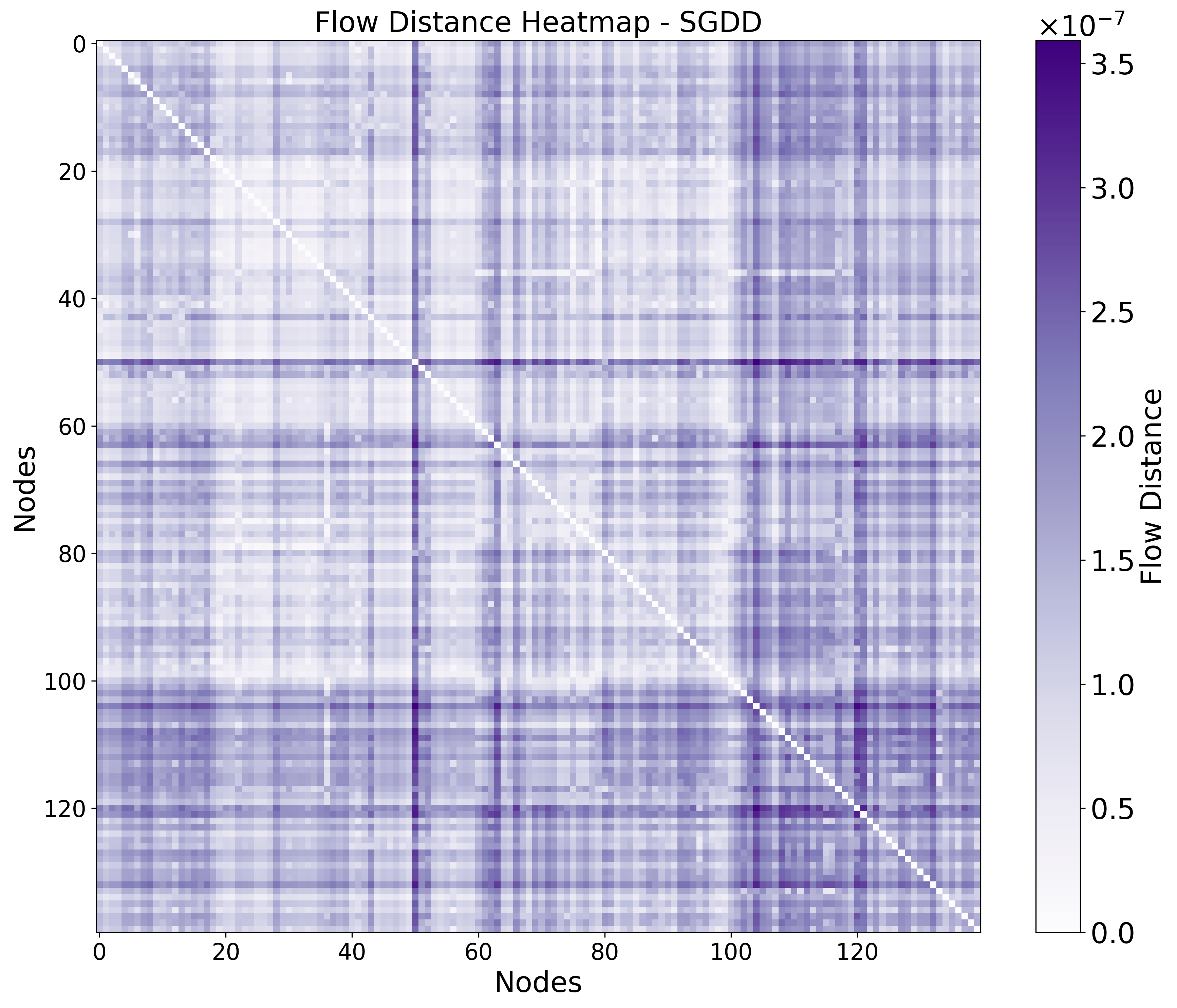}
        \caption{Condensed by SGDD}
        \label{fig:real_sgdd}
    \end{subfigure}%
    \begin{subfigure}[t]{0.33\textwidth}
        \centering
        \includegraphics[width=\textwidth]{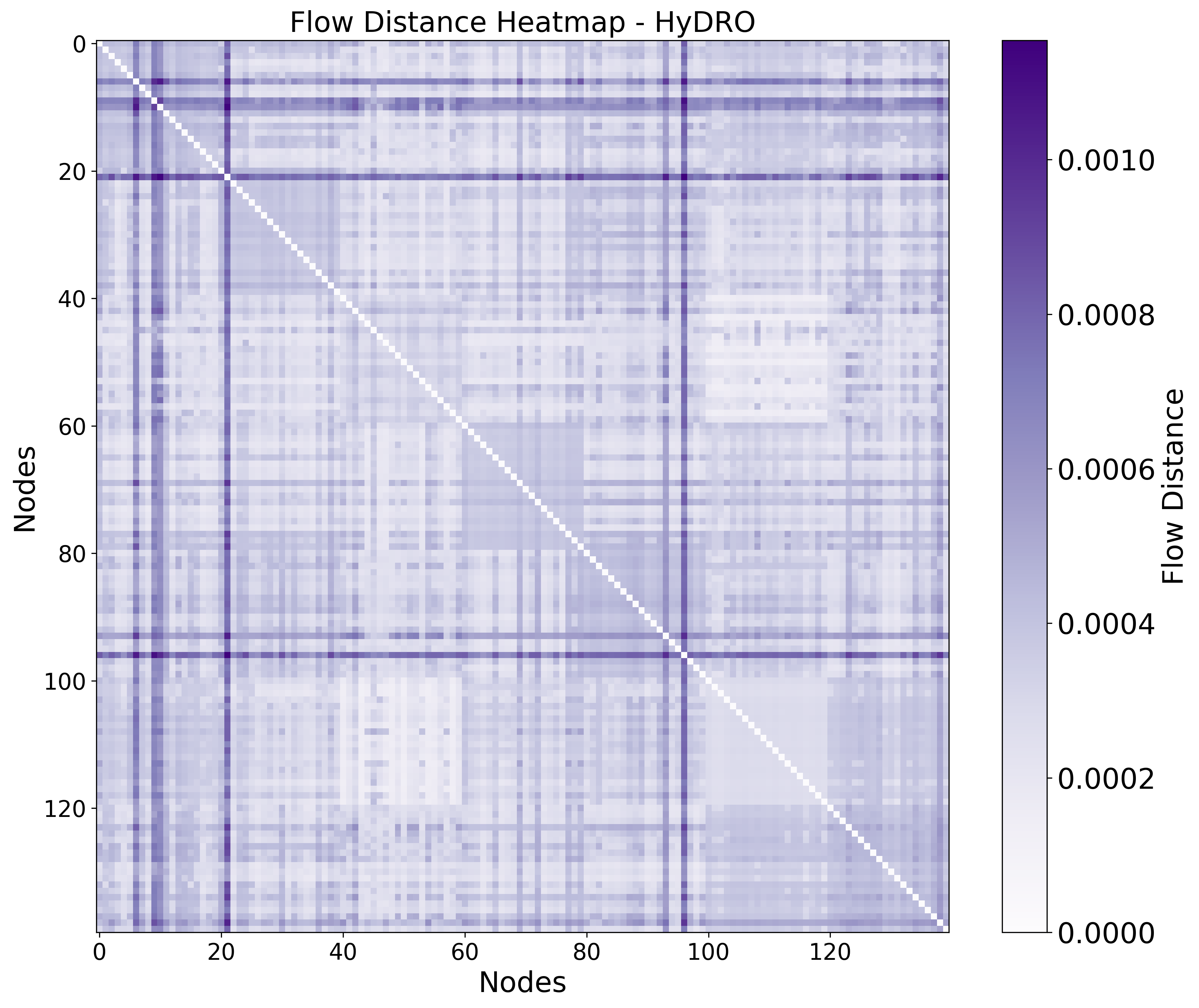}
        \caption{Condensed by HyDRO}
        \label{fig:real_hydro}
    \end{subfigure}%

    \caption{Flow distance matrices illustrating the connectivity between nodes in different graph representations: (a) Original Cora graph, (b) Synthetic graph distilled by SGDD, and (c) Synthetic graph distilled by HyDRO.  The reduction ratio $r$ is set to 5.2\%.}
    \label{fig:flow_distance}
\end{figure*}

In addition to commute time, the flow distance matrix \cite{gu2017hidden} plays a crucial role in understanding random walks, as it quantifies the likelihood of transitioning between nodes through the shortest paths. The flow distance captures the "cost" or "effort" required to travel between any two nodes in the graph, which is integral for modeling the efficiency of random walks.
Mathematically, the flow distance between two nodes \(i\) and \(j\) in a weighted graph can be computed using Dijkstra’s algorithm, which determines the shortest path between nodes. The flow distance \(d_{ij}\) between nodes \(i\) and \(j\) is the shortest path distance, defined as:
\[
d_{ij} = \min \left( \sum_{k=1}^{|P_{ij}|} w_k \right)
\]
where \(P_{ij}\) represents the set of all possible paths from node \(i\) to node \(j\), and \(w_k\) is the weight of the edge in the \(k\)-th position of the shortest path from \(i\) to \(j\). The quantity \( |P_{ij}| \) represents the number of edges in the path \(P_{ij}\). The algorithm computes this shortest path by minimizing the total sum of edge weights along the path.
To calculate the flow distance matrix for all pairs of nodes in the graph, we apply Dijkstra’s algorithm iteratively for each node \(i\). For each node \(i\), we calculate the shortest path distances \(d_{ij}\) from node \(i\) to every other node \(j\) in the graph. This results in the flow distance matrix \(D\), where each element \(D_{ij}\) represents the flow distance from node \(i\) to node \(j\).

\[
D_{ij} = d_{ij} = \min_{p \in P_{ij}} \sum_{k=1}^{|p|-1} w_{k}
\]
where \(p\) is a specific path between nodes \(i\) and \(j\), and \(w_k\) are the weights of the edges along that path.
The flow distance matrix \(D\) is then populated by repeating this procedure for all pairs of nodes. This matrix encapsulates the minimum travel cost (i.e., flow distance) between every pair of nodes in the graph.
Flow distances are crucial because smaller flow distances indicate that nodes are more tightly interconnected, which means the graph supports more efficient transitions (random walks). For example, in the Cora dataset, the synthetic graph distilled by HyDRO \autoref{fig:real_hydro} has smaller flow distances, signifying tighter interconnectivity that enhances random walk efficiency. In contrast, the original Cora graph \autoref{fig:real_original} exhibits slightly larger flow distances, suggesting that HyDRO effectively mirrors the original graph's structure.
On the other hand, the graph generated by the SGDD method \autoref{fig:real_sgdd} shows significantly smaller flow distances, indicating a considerable deviation from the original graph's connectivity pattern. This highlights how the flow distance matrix, computed using Dijkstra's algorithm, reflects the topological structure of the graph and the efficiency of random walks.

\section{Neural Architecture Search}  
\label{app:nas}

The architecture search space for APPNP includes the following hyperparameters: the number of propagations \( K \) is chosen from the set \(\{2, 4, 6, 8, 10\}\), the residual coefficient \( \alpha \) is selected from \(\{0.1, 0.2\}\), the hidden dimension can be one of \(\{16, 32, 64, 128, 256, 512\}\), and the activation function is chosen from the set \{Sigmoid, Tanh, ReLU, Linear, Softplus, LeakyReLU, ReLU6, ELU\}.
For the evaluation of NAS performance for all the condensed graphs, top-1 Test Accuracy gives the final evaluation of how well the selected architecture performs in practice after NAS. It is the most straightforward metric for assessing the success of the architecture search.
Besdies, validation Accuracy Correlation measures the reliability of the condensed graph as a proxy for the original graph during NAS. It tells you if the performance on the condensed graph reflects what would happen on the original graph. 
Additionally, rank Correlation ensures that the relative ranking of architectures remains consistent across both graphs. This helps to identify whether the best architectures for the condensed graph will still perform well on the original graph.

\input{tables/NAS}

\section{Cross-Architecture Transferability}  
\label{app:cross}

\subsubsection{Hyperparameters Searching}\label{app:hpo}
To ensure a fair comparison between different architectures, we perform hyperparameter optimization during the training process on the condensed graph. We identify the optimal hyperparameter configurations based on validation performance and report the corresponding test results. The hyperparameter search space for each graph neural network (GNN) includes the following options: the number of hidden units is chosen from {64, 256}, the learning rate is selected from {0.01, 0.001}, weight decay can either be 0 or 5e-4, and the dropout rate is set to either 0 or 0.5. For the Graph Attention Network (GAT), the number of attention heads is fixed at 8 to prevent memory issues, so we select the number of hidden units from {16, 64} and the dropout rate is evaluated within {0.0, 0.5, 0.7}. Furthermore, for the Simplified Graph Convolution (SGC) and Approximate Personalized Propagation of Neural Predictions (APPNP) models, we also investigate the impact of the number of linear layers, with possible values of {1, 2}. For APPNP, we additionally explore the effect of the parameter alpha, which is selected from {0.1, 0.2}.

\begin{table*}[h]
\centering
\caption{Experimental results on the {\bf cross-architecture transferability} task. The overall results are presented as the mean and standard deviation across all the seven architectures.}
\label{tab:cross_table_1}
\centering
\resizebox{0.70\textwidth}{!}{%
\begin{scriptsize}
\begin{tabular}{cccccccccccc}
\cline{1-11}
                          &         & \multicolumn{7}{c}{Architectures}                & \multicolumn{2}{c}{Result}                                                                                        &  \\ \cline{3-11}
\multirow{-2}{*}{Datasets} & \multirow{-2}{*}{Methods} & SGC  & GCN & Cheby & SAGE & APPNP& GAT & SGFormer & \textbf{Overall}  \\ \cline{1-11}
                    
                          & GCond     & 73.3 & 73.2 & 71.4  & 70.1  & 72.9 & 70.9 & 69.3 & \cellcolor{gray!30}71.6 $\pm$ 1.6         \\ 
                      
                          & SGDD     & 70.5 & 70.2 & 70.4  & 69.8  & 70.7 & 70.6 & 68.5 & \cellcolor{gray!30}70.1 $\pm$ 0.8          \\ 
                          & DosCond     & 64.5 & 68.3 & 51.8  & 60.3  & 50.5 & 47.1 & 38.3 & \cellcolor{gray!30}54.4 $\pm$ 10.5          \\

                          & GCSNTK   & 62.3 & 62.0 & 61.0  & 65.7  & 61.6 & 65.7 & 53.2 & \cellcolor{gray!30}61.7 $\pm$ 4.1          \\ 
                          & MSGC   & 70.3 & 70.2 & 70.5  & 69.7  & 70.7 & 71.4 & 69.5 & \cellcolor{gray!30}70.3 $\pm$ 0.6          \\ 
                          & GEOM   & 65.9 & 65.0 & 65.7  & 62.9  & 65.5 & 67.5 & 60.9 & \cellcolor{gray!30}64.8 $\pm$ 2.1           \\ 
                          & GDEM   & 72.5 & 72.5 & 73.1  & 72.8  & 73.7 & 72.7 & 70.9 & \cellcolor{gray!30}\textcolor{blue}{\textbf{72.6 $\pm$ 0.8} }          \\ 
                        
\multirow{-10}{*}{Citeseer(1.8\%)}  & HyDRO    & 72.8 & 72.8 & 72.4  & 72.1  & 72.9 & 69.6 & 70.0 & \cellcolor{gray!30}\textcolor{black}{\textbf{71.8 $\pm$ 1.4}}  \\ 

\midrule
& GCond     & 84.3 & 85.1 & 78.7  & 79.1  & 83.4 & 80.0 & 80.5 & \cellcolor{gray!30}81.6 $\pm$ 2.6       \\ 
                      
                          & SGDD     & 81.3 & 80.1 & 80.8  & 81.8  & 81.1 & 79.4 & 79.7 & \cellcolor{gray!30}80.6 $\pm$ 0.9       \\ 
                          & DosCond     & 81.7 & 82.9 & 81.4  & 81.6  & 81.3 & 80.4 & 73.9 & \cellcolor{gray!30}80.5 $\pm$ 2.9          \\

                          & GCSNTK   & 77.8 & 78.0 & 80.0  & 80.1  & 80.5 & 79.8 & 76.8 & \cellcolor{gray!30}79.4 $\pm$ 1.2           \\ 
                          & MSGC   & 84.9 & 84.8 & 84.0  & 83.8  & 84.9 & 74.2 & 70.7 & \cellcolor{gray!30}81.0 $\pm$ 5.9          \\ 
                          & GEOM   & 82.7 & 83.4 & 81.7  & 81.9  & 83.1 & 81.9 & 81.6 & \cellcolor{gray!30}\textcolor{blue}{\textbf{82.3 $\pm$ 0.7} }          \\
                          & GDEM   & 80.2 & 80.1 & 80.2  & 81.6  & 79.2 & 80.1 & 73.6 & \cellcolor{gray!30}79.3 $\pm$ 2.6           \\

\multirow{-10}{*}{DBLP(0.8\%)}  & HyDRO    & 83.0 & 83.0 & 81.1  & 81.4  & 82.8 & 80.7 & 81.6 & \cellcolor{gray!30}\textbf{82.0 $\pm$ 0.9}  \\ 

\midrule
& GCond     & 78.0 & 76.7 & 77.3  & 76.5  & 77.8 & 76.0 & 71.5 & \cellcolor{gray!30}76.2 $\pm$ 2.2         \\
                      
                          & SGDD     & 78.5 & 76.4 & 76.7  & 76.2  & 78.6 & 75.8 & 70.6 & \cellcolor{gray!30}76.1 $\pm$ 2.6         \\ 
                          & DosCond     & 75.0 & 77.7 & 74.9  & 71.6  & 76.3 & 68.6 & 69.3 & \cellcolor{gray!30}73.3 $\pm$ 3.5          \\ 
                        
                          & GCSNTK   & 77.1 & 75.8 & 74.3  & 76.3  & 76.2 & 76.8 & 71.3 & \cellcolor{gray!30}76.1 $\pm$ 1.0          \\ 
                          & MSGC   & 78.3 & 78.1 & 77.0  & 76.9  & 78.8 & 76.3 & 74.2 & \cellcolor{gray!30}77.1 $\pm$ 1.5           \\ 
                          & GEOM   & 78.6 & 78.4 & 78.2  & 77.3  & 78.4 & 78.0 & 78.0 & \cellcolor{gray!30}\textcolor{blue}{\textbf{78.1 $\pm$ 0.4}}           \\ 
                          & GDEM   & 78.2 & 76.6 & 76.8  & 77.2  & 77.6 & 74.6 & 77.0 & \cellcolor{gray!30}76.8 $\pm$ 1.1           \\ 
                        
\multirow{-10}{*}{Pubmed(0.15\%)}  & HyDRO    & 78.5 & 78.2 & 78.01  & 77.35  & 78.6 & 78.0 & 77.0 & \cellcolor{gray!30}\textbf{78.0 $\pm$ 0.6}  \\

\cline{1-11}

\end{tabular}%
\end{scriptsize}
}
\end{table*}


\section{Continual Graph Learning}  
\label{app:cgl}

\begin{figure*}[!h]
    \centering
    \begin{subfigure}[t]{0.49\textwidth}
        \centering
        \includegraphics[width=\textwidth]{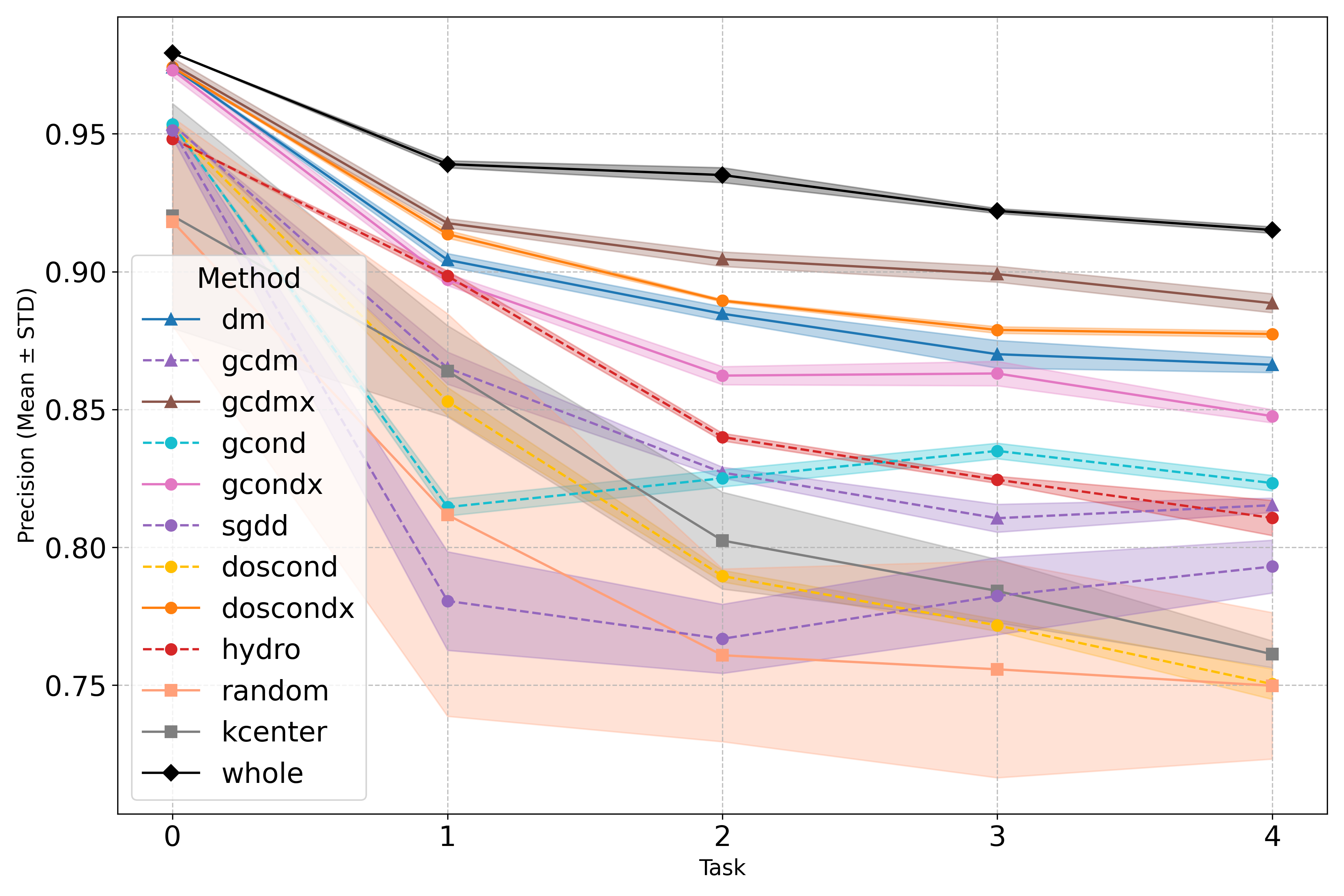}
        \caption{Wiki-CS dataset}
        \label{fig:cgl_wiki_cs}
    \end{subfigure}%
    \hfill
    \begin{subfigure}[t]{0.49\textwidth}
        \centering
        \includegraphics[width=\textwidth]{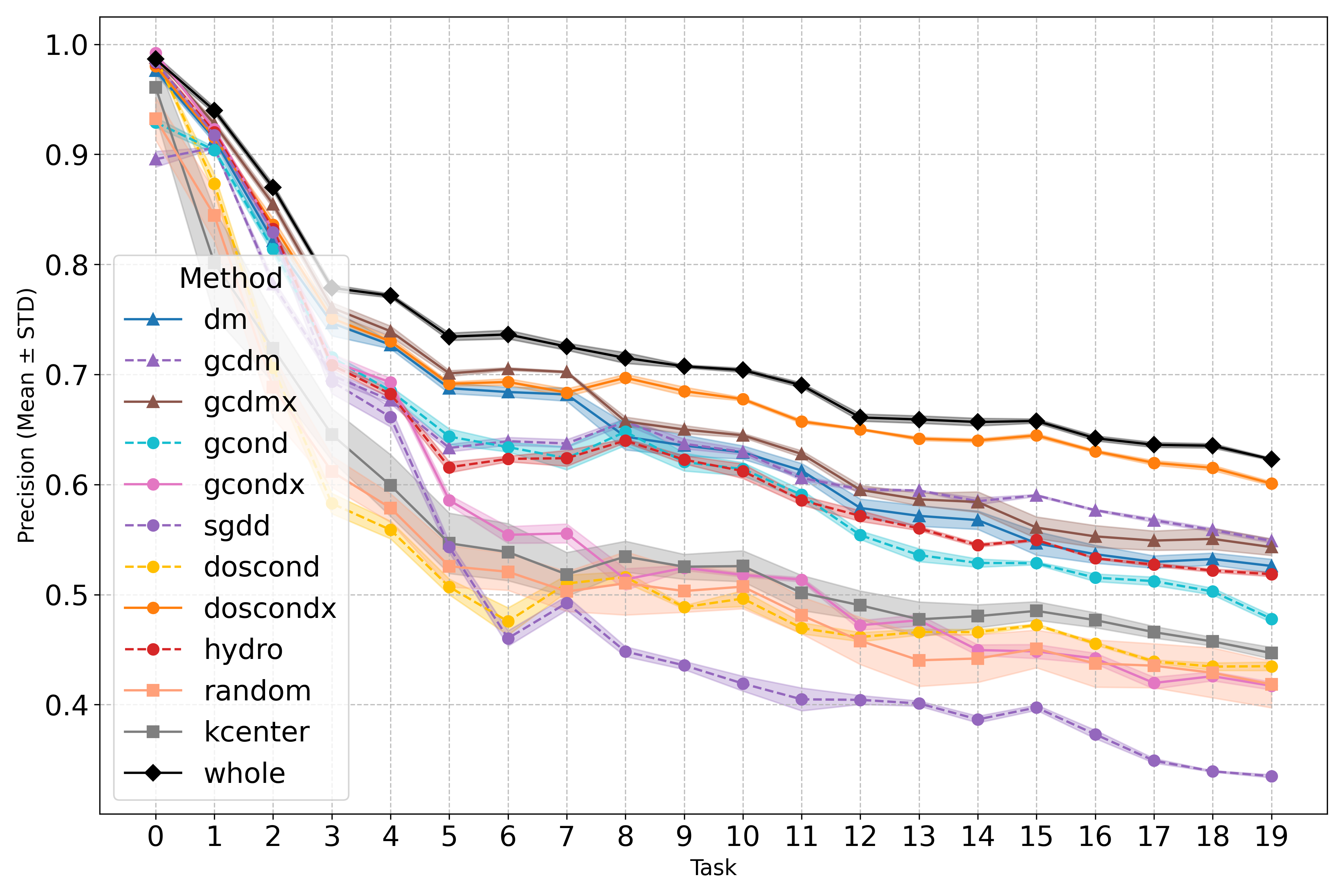}
        \caption{Arxiv dataset}
        \label{fig:cgl_arxiv}
    \end{subfigure}%
    \begin{subfigure}[t]{0.49\textwidth}
        \centering
        \includegraphics[width=\textwidth]{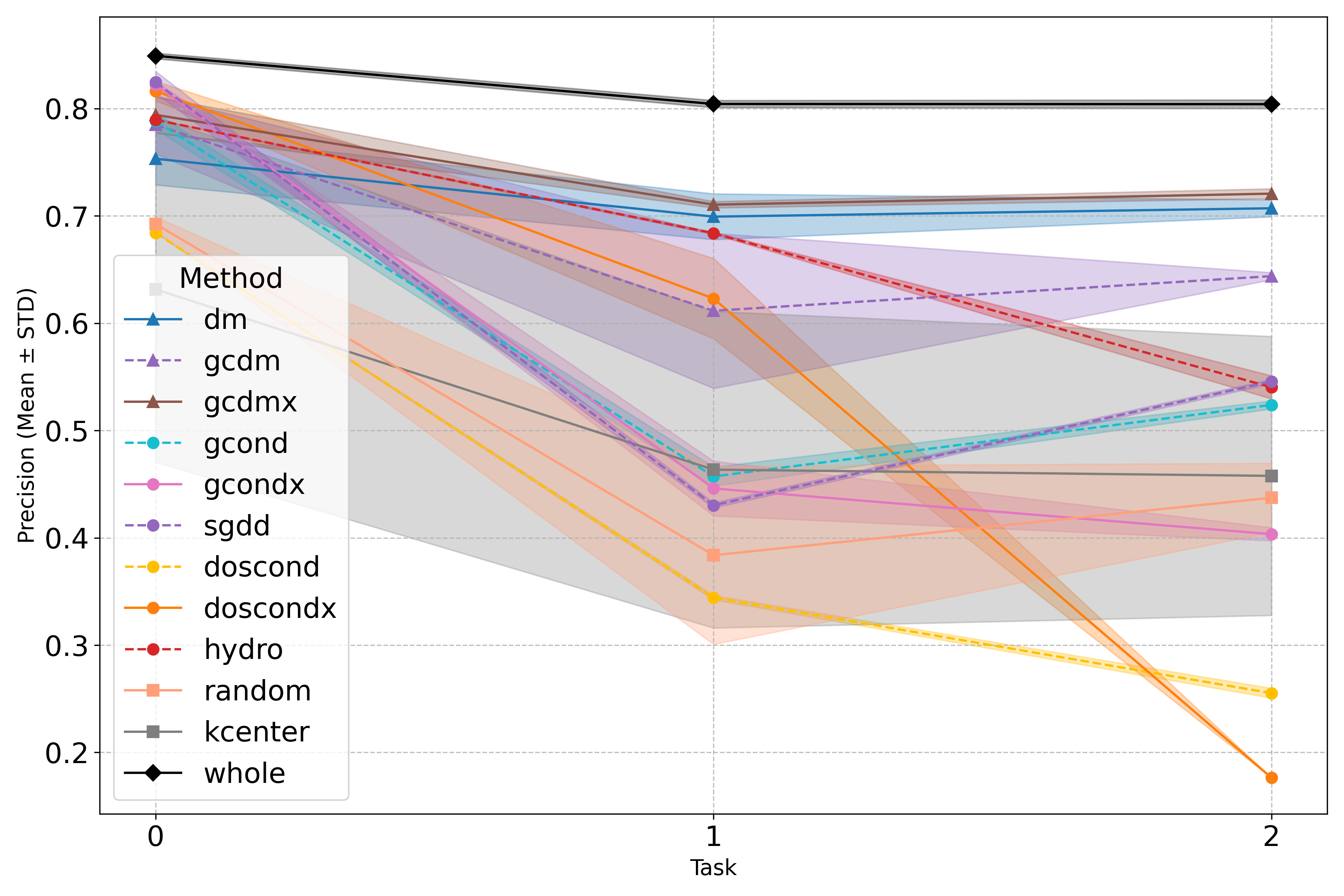}
        \caption{Citeseer dataset}
        \label{fig:cgl_amazon}
    \end{subfigure}
    
    
    \caption{Performance of GD methods on baseline datasets under the continual graph learning setting.}
    \label{fig:cgl-all2}
\end{figure*}

Continual Graph Learning (CGL) presents unique challenges compared to traditional node classification tasks, mainly due to the need to balance latency constraints with the ever-evolving nature of graph data. These challenges necessitate careful hyperparameter optimization to achieve optimal performance. Notably, simpler models, such as DM (Decoupled Method), tend to perform better in CGL settings, as they require fewer hyperparameters, thus simplifying the tuning process. For consistency across experiments, we follow the respective papers or source code to adopt either Simplified Graph Convolutions (SGC) or Graph Convolutional Networks (GCNs) for all methods. SGC is specifically used in HyDRO.
To ensure a fair comparison in continual graph learning, we fix the size of the condensed graph at 20 nodes across all datasets. Additionally, we standardize the train/validation/test split ratio to 6:2:2 for all streaming datasets.
We carefully fine-tuned key hyperparameters, with particular emphasis on the learning rates for feature learning ($l_{\text{feat}}$) and structure learning ($l_{\text{struct}}$), while fixing the number of optimization epochs at 300. For graph distillation methods such as GCond, GCDM, SGDD, and HyDRO, we explored a range of values for $l_{\text{feat}}$ and $l_{\text{struct}}$ from $\{0.1, 0.01, 0.001, 0.0001, 0.00001\}$, with inner loop epochs set to $\{1, 10\}$ and outer loop epochs ranging from $\{10, 20\}$. For DosCond, DM, and DosCondX, we fixed the outer loop number at 1.
For structure-free methods such as GCondX, DosCondX, and GCDMX, $l_{\text{struct}}$ was excluded. However, we retained similar settings for other hyperparameters as used in the structure-based methods. This meticulous and systematic approach highlights the advantages of condensed graphs, enabling efficient and scalable continual learning while addressing the unique challenges of the continual graph learning (CGL) paradigm. More experiments results are shown in \autoref{fig:cgl-all2}, indicating that HyRo can still performs well for the continual graph learning on a variety of graph datasets.

\section{Membership Inference Attack}
\label{app:mia}
\input{tables/MIA}

As shown in \autoref{tab:pp}, no method guarantees consistently high membership inference attack (MIA) accuracy across all three datasets—PubMed, Citeseer, and Flickr. For smaller datasets, such as PubMed and Citeseer, GEOM and GCond are the most effective at leaking membership privacy information, while GDEM follows closely behind. On the other hand, HyDRO can achieve higher original classification accuracy, offering a better utility balance while maintaining relatively strong privacy protection. For larger datasets like Flickr, GEOM and SGDD perform the best in terms of MIA accuracy, although they do not fare well in preserving the original task's accuracy. In contrast, HyDRO demonstrates a strong trade-off between utility and privacy, achieving favorable MIA results while maintaining competitive performance on the original classification task.

\section{Condensed Graph Denoising}
\label{app:denoising}
\input{tables/Noise}

As shown in \autoref{tab:robustness}, the performance of various methods is notably impacted by different types of noise, particularly for smaller datasets like Citeseer. SGDD and GEOM experience significant performance degradation under all types of noise. For example, for SGDD, feature noise causes a drop of approximately 30\%, while both structural and adversarial structural noise result in about a 10\% performance reduction. In contrast, MSGC, GCSNTK, and HyDRO perform relatively well across all types of noise, with only a 7-8\% drop in performance when subjected to structural noise. For larger datasets such as Flickr, methods like GCond, Doscond, and HyDRO demonstrate superior robustness to noise, with minimal performance degradation—only around 0-1\% drop in test accuracy under structural and adversarial structural noise conditions.

\section{Environment Settings}
\label{General Settings}

All experiments in this study are conducted within a high-performance computational cluster equipped with multiple GPUs, including NVIDIA A100 and NVIDIA GeForce RTX 4090 models. These GPUs provide substantial parallel processing capabilities to handle the large-scale computations required for our experiments. The underlying computational environment is configured with a Linux kernel version 5.15.0-113-generic, and the operating system is Ubuntu 20.04.6 LTS, which provides a stable and well-supported platform for machine learning tasks. This environment ensures a robust platform for conducting large-scale machine learning experiments, optimizing computational efficiency and accuracy in our results.

\begin{table}[h]
    \caption{Explanation of Reduction Rate under transductive and inductive settings}
    \centering
    \begin{tabular}{lccc}
        \toprule
        Dataset & Labeling Rate & Reduction Rate of Labeled Nodes & Reduction Rate $r$  \\
        \midrule
        \multirow{3}{*}{\textit{Citeseer}} && 25\% & 0.9\%\\
                                  &3.6\%& 50\% & 1.8\%\\
                                  && 100\% &  3.6\%\\  
        \cmidrule(l){3-4}
        \multirow{3}{*}{\textit{Cora}} && 25\% & 1.3\%\\
                              &5.2\%& 50\% & 2.6\%\\
                              && 100\% & 5.2\% \\
        \cmidrule(l){3-4}
        \multirow{3}{*}{\textit{Pubmed}} && 25\% & 0.08\%\\
                              &0.3\%& 50\% & 0.15\%\\
                              && 100\% & 0.30\% \\
        \cmidrule(l){3-4}
        \multirow{3}{*}{\textit{Arxiv}} && 0.1\% & 0.05\%\\
                               &53\%& 0.5\% & 0.25\%\\
                               && 1\% & 0.5\%\\
        \cmidrule(l){3-4}
        \multirow{3}{*}{DBLP} && 0.1\% & 0.08\%\\
                                &80\%& 0.5\% & 0.4\%\\
                                && 1\% & 0.8\%\\
        \cmidrule(l){3-4}
        \multirow{3}{*}{Wiki-CS} && 0.5\% & 0.4\%\\
                                &80\%& 1\% & 0.8\%\\
                                && 5\% & 4\%\\
                                \cmidrule(l){3-4}
        \multirow{3}{*}{Coauthor-Physics} && 0.1\% & 0.08\%\\
                                &80\%& 0.5\% & 0.4\%\\
                                && 1\% & 0.8\%\\
        \midrule
        \multirow{3}{*}{\textit{Flickr}} && 0.1\% & 0.1\%\\
                                &100\%& 0.5\% & 0.5\%\\
                                && 1\% & 1\%\\
        \cmidrule(l){3-4}
        \multirow{3}{*}{\textit{Reddit}} && 0.05\% & 0.05\%\\
                                &100\%& 0.1\% & 0.1\%\\
                                && 0.5\% & 0.5\%\\
        
        \bottomrule
    \end{tabular}
    \label{tab:reduction_rate_statistics}
\end{table}

\end{document}

%% file: tables/NC.tex
\renewcommand{\arraystretch}{1.5} 
\begin{table*}[ht!]
\caption{
    Experimental results on {\bf node classification task}. }

\label{tab:node_classification}
\resizebox{\textwidth}{!}{%
\begin{scriptsize}

\begin{tabular}{cclllllllllll}
\hline
\multirow{3}{*}{\textbf{Datasets}} &
  \multirow{3}{*}{\textbf{Ratio (\%)}} &
  \multicolumn{4}{c}{\multirow{2}{*}{\textbf{Traditional Methods}}} &
  \multicolumn{6}{c}{\textbf{Graph Distillation Methods}} &
  \multicolumn{1}{c}{\multirow{3}{*}{\textbf{\begin{tabular}[c]{@{}c@{}}Whole\\ Dataset\end{tabular}}}} \\ \cline{7-12}
 &
   &
  \multicolumn{4}{c}{} &
  \multicolumn{2}{c}{\textbf{Structure-free}} &
  \multicolumn{4}{c}{\textbf{Structure-based}} &
  \multicolumn{1}{c}{} \\ \cline{3-12}
 &
   &
  \multicolumn{1}{c}{\begin{tabular}[c]{@{}c@{}}Random\\ (A', X')\end{tabular}} &
  \multicolumn{1}{c}{\begin{tabular}[c]{@{}c@{}}Herding\\ (A', X')\end{tabular}} &
  \multicolumn{1}{c}{\begin{tabular}[c]{@{}c@{}}KCenter\\ (A', X')\end{tabular}} &
  \multicolumn{1}{c}{\begin{tabular}[c]{@{}c@{}}Coarsening\\ (A', X')\end{tabular}} &
  \multicolumn{1}{c}{\begin{tabular}[c]{@{}c@{}}SFGC\\ (X')\end{tabular}} &
  \multicolumn{1}{c}{\begin{tabular}[c]{@{}c@{}}GEOM\\ (X')\end{tabular}} &
  \begin{tabular}[c]{@{}l@{}}GCond\\ (A', X')\end{tabular} &
  \begin{tabular}[c]{@{}l@{}}SGDD\\ (A', X')\end{tabular} &
  \begin{tabular}[c]{@{}l@{}}GDEM\\ (U', X')\end{tabular} &
  \begin{tabular}[c]{@{}l@{}}HyDRO\\ (A', X')\end{tabular} &
  \multicolumn{1}{c}{} \\ \hline
\multirow{3}{*}{Cora} &
  1.30\% &
  63.6±3.7 &
  67.0±1.3 &
  64.0±2.3 &
  31.2±0.2 &
  80.7±0.4 &
  \textcolor{blue}{\textbf{82.5±0.4}} &
  79.8±1.3 &
  80.1±0.7 &
  79.4±0.3 &
  \textcolor{black}{\textbf{81.5±0.4}} &
  \multirow{3}{*}{81.2±0.2} \\ \cline{2-12}
 &
  2.60\% &
  72.8±1.1 &
  73.4±1.0 &
  73.2±1.2 &
  65.2±0.6 &
  81.7±0.5 &
  \textcolor{blue}{\textbf{83.6±0.7}} &
  80.1±0.6 &
  80.6±0.8 &
  80.0±0.9 &
  \textcolor{black}{\textbf{81.8±0.6}} &
   \\ \cline{2-12}
 &
  5.20\% &
  76.8±0.1 &
  76.8±0.1 &
  76.7±0.1 &
  70.6±0.1 &
  81.6±0.8 &
  \textcolor{blue}{\textbf{82.8±0.7}} &
  79.3±0.3 &
  80.4±1.6 &
  80.6±0.6 &
  \textcolor{black}{\textbf{{82.0±0.4}}} &
   \\ \hline
\multirow{3}{*}{Citeseer} &
  0.90\% &
  54.4±4.4 &
  57.1±1.5 &
  52.4±2.8 &
  52.2±0.4 &
  71.4±0.5 &
  \textcolor{blue}{\textbf{73.0±0.5}} &
  70.5±1.2 &
  69.5±0.4 &
  72.3±0.3 &
  \textcolor{black}{\textbf{{72.4±0.3}}} &
  \multirow{3}{*}{71.7±0.1} \\ \cline{2-12}
 &
  1.80\% &
  64.2±1.7 &
  66.7±1.0 &
  64.3±1.0 &
  59.0±0.5 &
  72.4±0.4 &
  \textcolor{blue}{\textbf{74.3±0.1}} &
  70.6±0.9 &
  70.2±0.8 &
  \textcolor{black}{\textbf{{72.6±0.6}}} &
  \textcolor{black}{\textbf{{72.6±0.4}}} &
   \\ \cline{2-12}
 &
  3.60\% &
  69.1±0.1 &
  69.0±0.1 &
  69.1±0.1 &
  65.3±0.5 &
  70.6±0.7 &
  \textcolor{blue}{\textbf{73.3±0.4}} &
  69.8±1.4 &
  70.3±1.7 &
  \textcolor{black}{\textbf{{72.6±0.5}}} &
  \textcolor{black}{\textbf{{72.6±0.7}}} &
   \\ \hline
\multirow{3}{*}{Pubmed} &
  0.08\% &
  69.4±0.2 &
  76.7±0.7 &
  64.5±2.7 &
  18.1±0.1 &
  76.4±1.2 &
  \textcolor{black}{\textbf{78.3±0.2}}&
  76.5±0.2 &
  77.1±0.5 &
  77.7±0.7 &
  \textcolor{blue}{\textbf{78.5±0.4}} &
  \multirow{3}{*}{79.3±0.2} \\ \cline{2-12}
 &
  0.15\% &
  73.3±0.7 &
  76.2±0.5 &
  69.4±0.7 &
  28.7±4.1 &
  77.5±0.4 &
  78.1±0.3 &
  77.1±0.5 &
  78.0±0.3 &
  \textcolor{black}{\textbf{78.4±1.8}} &
  \textcolor{blue}{\textbf{78.6±0.8}} &
   \\ \cline{2-12}
 &
  0.30\% &
  77.8±0.3 &
  78.0±0.5 &
  78.2±0.4 &
  42.8±4.1 &
  77.9±0.3 &
  77.2±0.5 &
  77.9±0.4 &
  77.5±0.5 &
  \textcolor{black}{\textbf{78.2±0.8}} &
  \textcolor{blue}{\textbf{78.5±0.6}} &
   \\ \hline
\multirow{3}{*}{Ogbn-arxiv} &
  0.05\% &
  47.1±3.9 &
  52.4±1.8 &
  47.2±3.0 &
  35.4±0.3 &
  \textcolor{black}{\textbf{65.5±0.7}} &
  \textcolor{blue}{\textbf{65.5±0.6}} &
  59.2±1.1 &
  60.8±1.3 &
  63.7±0.8 &
  64.3±0.5 &
  \multirow{3}{*}{71.4±0.1} \\ \cline{2-12}
 &
  0.25\% &
  57.3±1.1 &
  58.6±1.2 &
  56.8±0.8 &
  43.5±0.2 &
  66.1±0.4 &
  \textcolor{blue}{\textbf{68.8±0.2}} &
  63.2±0.3 &
  65.8±1.2 &
  63.8±0.6 &
  \textcolor{black}{\textbf{66.2±0.8}} &
   \\ \cline{2-12}
 &
  0.50\% &
  60.0±0.9 &
  60.4±0.8 &
  60.3±0.4 &
  50.4±0.1 &
  66.8±0.4 &
  \textcolor{blue}{\textbf{69.6±0.2}} &
  64.0±0.4 &
  66.3±0.7 &
  64.1±0.3 &
  \textcolor{black}{\textbf{66.9±0.3}} &
   \\ \hline
\multirow{3}{*}{Flickr} &
  0.10\% &
  41.8±2.0 &
  42.5±1.8 &
  42.0±0.7 &
  41.9±0.2 &
  46.6±0.2 &
  47.1±0.1 &
  46.5±0.4 &
  46.9±0.1 &
  \textcolor{blue}{\textbf{49.9±0.8}} &
  \textcolor{black}{\textbf{47.3±0.5}} &
  \multirow{3}{*}{47.2±0.1} \\ \cline{2-12}
 &
  0.50\% &
  44.0±0.4 &
  43.9±0.9 &
  43.2±0.1 &
  44.5±0.1 &
  47.0±0.1 &
  47.0±0.2 &
  47.1±0.1 &
  47.1±0.3 &
  \textcolor{blue}{\textbf{49.4±1.3}} &
  \textcolor{black}{\textbf{47.4±0.4}} &
   \\ \cline{2-12}
 &
  1.00\% &
  44.6±0.2 &
  44.4±0.6 &
  44.1±0.4 &
  44.6±0.1 &
  47.1±0.1 &
  47.3±0.3 &
  47.1±0.1 &
  47.1±0.1 &
  \textcolor{blue}{\textbf{49.9±0.6}} &
  \textcolor{black}{\textbf{47.3±0.6}} &
   \\ \hline

\multirow{3}{*}{Reddit} &
  0.01\% &
  46.1±4.4 &
  53.1±2.5 &
  46.6±2.3 &
  40.9±0.5 &
  89.7±0.2 &
  \textcolor{black}{\textbf{91.1±0.1}} &
  88.0±1.8 &
  90.5±2.1 &
  \textcolor{blue}{\textbf{92.9±0.3}} &
  90.6±0.7 &
  \multirow{3}{*}{93.9±0.0} \\ \cline{2-12}
 &
  0.05\% &
  58.0±2.2 &
  62.7±1.0 &
  53.0±3.3 &
  42.8±0.8 &
  90.0±0.3 &
  91.3±0.2 &
  89.6±0.7 &
  \textcolor{black}{\textbf{91.8±1.9}} &
  \textcolor{blue}{\textbf{93.1±0.2}} &
  \textcolor{black}{\textbf{{91.8±1.4}}} &
   \\ \cline{2-12}
 &
  0.50\% &
  66.3±1.9 &
  71.0±1.6 &
  58.5±2.1 &
  47.4±0.9 &
  89.9±0.4 &
  91.3±0.8 &
  90.1±0.5 &
  91.6±1.8 &
  \textcolor{blue}{\textbf{93.2±0.4}} &
  \textcolor{black}{\textbf{91.9±1.1}} &
   \\ \hline
\end{tabular}%

\end{scriptsize} 
}
\end{table*}

%% file: tables/NC_LP.tex
\begin{table*}
\centering
\caption{
    Experimental results on {\bf task generalization} from node classification (NC) to link prediction (LP).
  }
\label{tab:coss-tasks}
\resizebox{\textwidth}{!}{%
\begin{tabular}{ccclllllcllcllll}
\hline
\multirow{3}{*}{\textbf{Datasets}} &
  \multirow{3}{*}{\textbf{\begin{tabular}[c]{@{}c@{}}Downstream \\ Tasks\end{tabular}}} &
  \multirow{3}{*}{\textbf{Ratio (\%)}} &
  \multicolumn{3}{c}{\multirow{2}{*}{\textbf{Traditional Methods}}} &
  \multicolumn{9}{c}{\textbf{Graph Distillation Methods}} &
  \multicolumn{1}{c}{\multirow{3}{*}{\textbf{\begin{tabular}[c]{@{}c@{}}Whole\\ Dataset\end{tabular}}}} \\ \cline{7-15}
 &
   &
   &
  \multicolumn{3}{c}{} &
  \multicolumn{3}{c}{\textbf{Structure-free}} &
  \multicolumn{6}{c}{\textbf{Structure-based}} &
  \multicolumn{1}{c}{} \\ \cline{4-15}
 &
   &
   &
  \multicolumn{1}{c}{\begin{tabular}[c]{@{}c@{}}Random\\ (A', X')\end{tabular}} &
  \multicolumn{1}{c}{\begin{tabular}[c]{@{}c@{}}Herding\\ (A', X')\end{tabular}} &
  \multicolumn{1}{c}{\begin{tabular}[c]{@{}c@{}}KCenter\\ (A', X')\end{tabular}} &
  \multicolumn{1}{c}{\begin{tabular}[c]{@{}c@{}}SFGC\\ (X')\end{tabular}} &
  \multicolumn{1}{c}{\begin{tabular}[c]{@{}c@{}}GEOM\\ (X')\end{tabular}} &
  \begin{tabular}[c]{@{}c@{}}GCSNTK\\ (A', X')\end{tabular} &
  \begin{tabular}[c]{@{}l@{}}GCond\\ (A', X')\end{tabular} &
  \begin{tabular}[c]{@{}l@{}}SGDD\\ (A', X')\end{tabular} &
  \begin{tabular}[c]{@{}c@{}}DosCond\\ (A', X')\end{tabular} &
  \begin{tabular}[c]{@{}l@{}}MSGC\\ (A', X')\end{tabular} &
  \begin{tabular}[c]{@{}l@{}}GDEM\\ (U', X')\end{tabular} &
  \begin{tabular}[c]{@{}l@{}}HyDRO\\ (A', X')\end{tabular} &
  \multicolumn{1}{c}{} \\ \hline
\multirow{6}{*}{Cora} &
  \multirow{3}{*}{NC} &
  1.30\% &
  66.9±0.8 &
  67.1±0.6 &
  62.0±1.9 &
  73.5±1.1 &
  78.2±0.3 &
  73.4±1.8 &
  \textcolor{blue}{\textbf{80.9±0.7}} &
  78.7±0.9 &
  79.5±0.4 &
  \textbf{80.5±0.4} &
  75.1±0.9 &
  80.3±0.5 &
  \multirow{3}{*}{81.1±0.4} \\ \cline{3-15}
 &
   &
  2.60\% &
  71.2±0.4 &
  72.5±0.9 &
  66.8±0.2 &
  77.4±0.6 &
  74.1±0.6 &
  76.1±0.8 &
  \textbf{80.7±1.0} &
  80.4±0.8 &
  79.9±0.7 &
  \textcolor{blue}{\textbf{80.9±0.6}} &
  77.7±0.3 &
  79.9±0.8 &
   \\ \cline{3-15}
 &
   &
  5.20\% &
  72.6±1.7 &
  76.8±0.5 &
  73.9±0.6 &
  79.1±0.6 &
  \textcolor{blue}{\textbf{81.6±0.4}} &
  76.6±0.7 &
  79.4±0.3 &
  81.0±1.2 &
  \textbf{81.1±0.5} &
  80.8±0.4 &
  80.3±0.2 &
  80.9±0.3 &
   \\ \cline{2-16}
 &
  \multirow{3}{*}{LP} &
  \cellcolor{gray!30}1.30\% &
  \cellcolor{gray!30}70.5±1.4 &
  \cellcolor{gray!30}71.6±1.6 &
  \cellcolor{gray!30}72.1±1.4 &
  \cellcolor{gray!30}68.8±1.5 &
  \cellcolor{gray!30}71.7±1.7 &
  \cellcolor{gray!30}61.2±2.8 &
  \cellcolor{gray!30}66.1±0.4 &
  \cellcolor{gray!30}72.7±0.8 &
  \cellcolor{gray!30}59.6±1.3 &
  \cellcolor{gray!30}68.8±0.5 &
  \cellcolor{gray!30}\textbf{73.2±1.7} &
  \cellcolor{gray!30}\textcolor{blue}{\textbf{74.5±1.3}} &
  \multirow{3}{*}{78.5±1.1} \\ \cline{3-15}
 &
 &
  \cellcolor{gray!30}2.60\% &
  \cellcolor{gray!30}70.4±1.8 &
  \cellcolor{gray!30}71.9±1.4 &
  \cellcolor{gray!30}71.6±0.9 &
  \cellcolor{gray!30}69.7±1.0 &
  \cellcolor{gray!30}72.0±1.1 &
  \cellcolor{gray!30} 67.9±1.7 &
  \cellcolor{gray!30}50.9±0.1 &
  \cellcolor{gray!30} \textbf{73.6±0.7} &
  \cellcolor{gray!30}64.3±2.8 &
  \cellcolor{gray!30}70.0±1.3 &
  \cellcolor{gray!30}72.8±0.8 &
  \cellcolor{gray!30}\textcolor{blue}{\textbf{74.0±0.7}} &
   \\ \cline{3-15}
 &
   &
  \cellcolor{gray!30}5.20\% &
  \cellcolor{gray!30}71.5±1.8 &
  \cellcolor{gray!30}72.7±1.1 &
  \cellcolor{gray!30}72.6±0.9 &
  \cellcolor{gray!30}68.8±1.8 &
  \cellcolor{gray!30}71.8±1.5 &
  \cellcolor{gray!30}64.8±2.8 &
  \cellcolor{gray!30}55.9±1.6 &
  \cellcolor{gray!30}\textbf{74.5±0.4} &
  \cellcolor{gray!30}70.9±0.9 &
  \cellcolor{gray!30}69.6±2.5 &
  \cellcolor{gray!30}64.5±0.2 &
  \cellcolor{gray!30}\textcolor{blue}{\textbf{75.2±0.4}} &
   \\ \hline
\multirow{6}{*}{Citeseer} &
  \multirow{3}{*}{NC} &
  0.90\% &
  45.4±0.4 &
  58.7±0.6 &
  54.2±1.3 &
  41.9±0.4 &
  69.3±0.8 &
  65.9±0.9 &
  70.9±0.7 &
  71.8±0.2 &
  69.9±0.7 &
  70.9±0.4 &
  \textbf{72.2±0.7} &
  \textcolor{blue}{\textbf{73.2±0.5}} &
  \multirow{3}{*}{71.8±0.3} \\ \cline{3-15}
 &
   &
  1.80\% &
  63.7±0.5 &
  67.3±0.4 &
  64.2±0.3 &
  43.9±0.7 &
  64.9±1.5 &
  61.9±1.4 &
  \textcolor{blue}{\textbf{73.1±0.2}} &
  70.2±0.3 &
  68.1±0.6 &
  70.3±0.5 &
  \textbf{72.7±0.3} &
  \textbf{72.7±0.2} &
   \\ \cline{3-15}
 &
   &
  3.60\% &
  69.2±0.6 &
  69.3±0.6 &
  69.4±0.4 &
  64.5±1.0 &
  69.6±0.3 &
  66.6±0.9 &
  71.2±0.4 &
  \textbf{72.2±0.2} &
  68.8±0.5 &
  71.0±0.6 &
  \textcolor{blue}{\textbf{72.9±0.1}} &
  71.7±0.2 &
   \\ \cline{2-16} 
 &
  \multirow{3}{*}{LP} &
  \cellcolor{gray!30}0.90\% &
  \cellcolor{gray!30}71.4±1.3 &
  \cellcolor{gray!30}72.0±1.5 &
  \cellcolor{gray!30}72.7±1.4 &
  \cellcolor{gray!30}68.5±2.6 &
  \cellcolor{gray!30}73.2±1.7 &
  \cellcolor{gray!30}62.9±1.4 &
  \cellcolor{gray!30}57.4±2.6 &
  \cellcolor{gray!30}74.6±1.4 &
  \cellcolor{gray!30}72.5±2.5 &
  \cellcolor{gray!30}63.3±1.6 &
  \cellcolor{gray!30}\textcolor{blue}{\textbf{78.8±1.1}} &
  \cellcolor{gray!30}\textbf{75.4±0.5} &
  \multirow{3}{*}{81.1±1.0} \\ \cline{3-15}
 &
   &
  \cellcolor{gray!30}1.80\% &
  \cellcolor{gray!30}72.4±0.5 &
  \cellcolor{gray!30}\textcolor{black}{\textbf{72.6±1.0}} &
  \cellcolor{gray!30}71.2±2.1 &
  \cellcolor{gray!30}65.6±3.1 &
  \cellcolor{gray!30}70.5±1.5 &
  \cellcolor{gray!30}65.6±2.5 &
  \cellcolor{gray!30}70.1±0.5 &
  \cellcolor{gray!30}50.0±0.0 &
  \cellcolor{gray!30}64.89±1.2 &
  \cellcolor{gray!30}61.1±2.5 &
  \cellcolor{gray!30}57.8±0.6 &
  \cellcolor{gray!30}\textcolor{blue}{\textbf{75.6±0.6}} &
   \\ \cline{3-15}
 &
   &
  \cellcolor{gray!30}3.60\% &
  \cellcolor{gray!30}73.2±0.2 &
  \cellcolor{gray!30}72.2±1.5 &
  \cellcolor{gray!30}72.6±0.7 &
  \cellcolor{gray!30}68.2±2.8 &
  \cellcolor{gray!30}\textbf{72.3±1.2} &
  \cellcolor{gray!30}64.61±4.2 &
  \cellcolor{gray!30}64.8±1.3 &
  \cellcolor{gray!30}64.9±8.3 &
  \cellcolor{gray!30}62.0±2.5 &
  \cellcolor{gray!30}70.3±0.9 &
  \cellcolor{gray!30}56.9±1.8 &
  \cellcolor{gray!30}\textcolor{blue}{\textbf{74.8±0.9}} &
   \\ \hline
\multirow{6}{*}{DBLP} &
  \multirow{3}{*}{NC} &
  0.08\% &
  49.6±1.5 &
  60.5±0.4 &
  58.1±0.5 &
  61.8±0.5 &
  71.3±0.3 &
  62.3±0.5 &
  79.3±0.3 &
  79.4±1.6 &
  \textbf{82.8±0.2} &
  \textcolor{blue}{\textbf{84.5±0.1}} &
  77.9±0.3 &
  82.1±0.7 &
  \multirow{3}{*}{85.6±0.2} \\ \cline{3-15}
 &
   &
  0.40\% &
  67.4±0.3 &
  67.1±0.3 &
  61.7±0.4 &
  67.4±0.6 &
  75.7±0.5 &
  66.2±0.8 &
  82.7±0.2 &
  81.8±0.4 &
  \textbf{83.7±0.2} &
  \textcolor{blue}{\textbf{84.0±0.3}} &
  77.6±0.1 &
  81.6±0.6 &
   \\ \cline{3-15}
 &
   &
  0.80\% &
  71.9±0.3 &
  71.0±0.3 &
  64.3±0.2 &
  71.7±0.5 &
  80.3±0.3 &
  66.9±0.3 &
  \textbf{83.5±0.3} &
  75.3±1.7 &
  82.8±0.2 &
  \textcolor{blue}{\textbf{84.1±0.1}} &
  80.1±0.2 &
  82.7±0.8 &
   \\ \cline{2-16} 
 &
  \multirow{3}{*}{LP} &
  \cellcolor{gray!30}0.08\% &
  \cellcolor{gray!30}51.7±0.7 &
  \cellcolor{gray!30}51.1±1.7 &
  \cellcolor{gray!30}50.8±0.2 &
  \cellcolor{gray!30}52.1±0.6 &
  \cellcolor{gray!30}73.2±1.7 &
  \cellcolor{gray!30}62.9±1.4 &
  \cellcolor{gray!30}67.9±0.5 &
  \cellcolor{gray!30}\textbf{74.5±1.4} &
  \cellcolor{gray!30}72.5±2.5 &
  \cellcolor{gray!30}73.4±0.4 &
  \cellcolor{gray!30}67.1±3.4 &
  \cellcolor{gray!30}\textcolor{blue}{\textbf{74.5±0.5}} &
  \multirow{3}{*}{78.1±0.1} \\ \cline{3-15}
 &
   &
  \cellcolor{gray!30}0.40\% &
  \cellcolor{gray!30}50.9±0.3 &
  \cellcolor{gray!30}51.3±0.5 &
  \cellcolor{gray!30}51.0±0.3 &
  \cellcolor{gray!30}51.6±2.3 &
  \cellcolor{gray!30}70.5±1.5 &
  \cellcolor{gray!30}65.6±2.5 &
  \cellcolor{gray!30}72.9±0.4 &
  \cellcolor{gray!30}50.0±0.0 &
  \cellcolor{gray!30}64.9±1.2 &
  \cellcolor{gray!30}\textbf{74.7±0.5} &
  \cellcolor{gray!30}66.2±1.5 &
  \cellcolor{gray!30}\textcolor{blue}{\textbf{78.0±0.4}} &
   \\ \cline{3-15}
 &
   &
  \cellcolor{gray!30}0.80\% &
  \cellcolor{gray!30}51.4±0.6 &
  \cellcolor{gray!30}51.5±0.5 &
  \cellcolor{gray!30}51.9±0.4 &
  \cellcolor{gray!30}53.5±1.8 &
  \cellcolor{gray!30}72.3±1.2 &
  \cellcolor{gray!30}64.6±4.2 &
  \cellcolor{gray!30}\textbf{74.6±0.2} &
  \cellcolor{gray!30}64.9±8.3 &
  \cellcolor{gray!30}62.0±2.5 &
  \cellcolor{gray!30}73.2±1.2 &
  \cellcolor{gray!30}67.1±1.5 &
  \cellcolor{gray!30}\textcolor{blue}{\textbf{77.5±0.3}} &
   \\ \hline
\end{tabular}%
}
\end{table*}

%% file: tables/NAS.tex
\begin{table*}[h!]
\centering
\caption{Evaluation results for the {\bf Neural Architecture Search} are presented, including mean accuracy and standard deviation. The best results are highlighted in \textcolor{blue}{\textbf{bold}}, while the second-best values are marked in \textbf{\textit{bold italic}}}
\resizebox{0.9\linewidth}{!}{%
\setlength{\tabcolsep}{1.5pt} 
\renewcommand{\arraystretch}{0.9} 
\begin{scriptsize} 

\begin{tabular}{@{}llccccccccccccccc@{}}
\toprule
\textbf{Dataset} & \textbf{Metrics} & \textbf{Averaging} & \textbf{Random} & \textbf{Herding} & \textbf{K-Center} & \textbf{GCond} & \textbf{SGDD} & \textbf{DosCond} & \textbf{MSGC} & \textbf{GCSNTK}  & \textbf{SFGC} & \textbf{GEOM} & \textbf{GDEM} &  \textbf{HyDRO} \\ 
\midrule

\multirow{3}{*}{\centering \makecell{Citeseer\\(0.9\%)}} 
& Top 1 (\%)                     & 70.00           & 47.92             & 60.22           & 58.35 & 71.76         & \textbf{73.03}                                & {\ul 65.84}   & 71.38            & 68.78                          & 42.51 & 70.75 & 71.67& \textcolor{blue}{\textbf{73.40}} \\ 

& Acc. Corr.                           & 18.21            & 3.29 & 17.09          & -36.31           &  62.39 & {\ul 22.94}    & 5.61             & 82.45             & -18.99             & 26.85   & \textbf{58.80}& 9.69& \textcolor{blue}{\textbf{76.80}} \\ 
& Rank Corr.                           & 49.71              & 47.18                      & {\ul 54.69}    & 14.25           & 44.69 & 55.33 & 26.27             & \textcolor{blue}{\textbf{79.69 }}            & -13.27             & 71.03    & 60.67 &-2.47& \textbf{72.87} \\ 
\midrule

\multirow{3}{*}{\centering \makecell{Pubmed\\(0.08\%)}}
& Top 1 (\%)                               & 76.56             & 70.53           & 75.48 & 73.31         & 78.53                               & {\ul 78.34}   & 76.14            & 77.79            & 77.29            & 77.92    & \textcolor{blue}{\textbf{79.34}}  & 77.91& \textbf{78.92} \\ 
& Acc. Corr.                              & 14.84            & 32.09  & 44.95          & 48.28           & \textbf{ 63.50} & \textcolor{blue}{\textbf{ 64.96}}    & 14.99             & 50.93             & 48.51             & 55.43    & 41.68 &-11.35& 59.60 \\ 
& Rank Corr.                        & 20.65              & 51.17                       & {\ul 60.06}    & 60.58           & \textcolor{blue}{\textbf{ 82.88}} & 76.92 & 7.22            & 52.05             & 33.63             & 66.12    & 54.53 & -9.55& \textbf{77.75} \\ 
\midrule

\multirow{3}{*}{\centering \makecell{Flickr \\(0.5\%)}}
& Top 1 (\%)                    & 44.77             & 43.86         & 41.28 & 43.09               & \textbf{46.84}                      & {\ul 44.79}   & 46.72            & 46.76            & 46.01            & 44.24  & 45.05 & 46.35 & \textcolor{blue}{\textbf{46.90}} \\ 
& Acc. Corr.                        & 72.40              & \textcolor{blue}{\textbf{87.94}}            & \textbf{87.56}           & 89.38          & 75.43 & {\ul 77.96}    & 84.07             & 75.69             & 70.31             & 86.68   & 85.59 &74.88& 80.80 \\ 
& Rank Corr.                             & 73.08              & 72.09                    &\textbf{ 76.70}    & 74.19           & 70.61 & 66.24 & 76.43             & 73.14            & 69.33             & 73.79   & 71.18 & 71.69&\textcolor{blue}{\textbf{80.11}} \\ 

\midrule

\multirow{3}{*}{\centering \makecell{DBLP\\(0.08\%)}}
& Top 1 (\%)                                & \textcolor{blue}{\textbf{ 82.87}}           & 54.28 & 62.96         & 60.76         & 81.72                       & {\ul 80.79}   & \textbf{82.41}            & 84.25            & 79.65            & 63.68    & 71.58 &78.49 & 82.10 \\ 
& Acc. Corr.                                  & 65.13           & 21.32  & 39.40          & 38.42          & 55.26 & {\ul 52.39}    & 51.67             & \textcolor{black}{\textbf{73.63}}             & 43.11             & 52.07    & 50.04 & \textcolor{blue}{\textbf{81.08}}& 63.54 \\ 
& Rank Corr.                     & 62.11              & 20.79                      & {\ul 37.93}    & 36.38           & 57.81 & 52.12 & 56.13             & 77.60             & 47.62             & 65.97   & 63.25 & \textcolor{blue}{\textbf{82.19}} &\textcolor{black}{\textbf{68.62}} \\ 

\bottomrule
\end{tabular}%
\end{scriptsize}
}
\label{tab:NAS}
\end{table*}

%% file: tables/MIA.tex
\begin{table*}[h]

\centering
\caption{
    The experimental results for the {\bf Membership Inference Attacks} are presented, including mean accuracy and standard deviation. For both MIA accuracy and original accuracy in the normal settings, the best results are highlighted in \textcolor{blue}{\textbf{bold}}, while the second-best values are marked in \textbf{bold}. 
    And $\downarrow$ indicates ``the lower the better'', and $\uparrow$ means the opposite.
}
\label{tab:pp}
\setlength{\tabcolsep}{5pt} 
\renewcommand{\arraystretch}{1} 
\begin{scriptsize} 
\begin{tabular}{@{}ccccccc@{}}
\toprule
\multirow{2}{*}{Methods} & \multicolumn{2}{c}{Flickr, r = 0.5\%} & \multicolumn{2}{c}{PubMed, r = 0.15\%} & \multicolumn{2}{c}{Citeseer, r = 1.8\%} \\ \cmidrule(l){2-3} \cmidrule(l){4-5} \cmidrule(l){6-7} 
                         & Original Acc $\uparrow$         & MIA Acc $\downarrow$      & Original Acc  $\uparrow$        & MIA Acc $\downarrow$          & Original Acc $\uparrow$         & MIA Acc $\downarrow$      \\ \midrule

Random          & 43.9 ± 0.5       & 53.4 ± 0.2            & 71.8 ± 0.7         & 71.6 ± 0.6              & 63.8 ± 0.5            & 69.1 ± 0.5             \\
Averaging          & 37.8 ± 0.4       & 56.0 ± 0.1            & 75.6 ± 0.4         & 70.0 ± 0.4              & 69.8 ± 0.3            & 75.5 ± 0.3             \\
Herding         & 42.0 ± 0.3       & 53.3 ± 0.1            & 75.6 ± 0.5         & 72.7 ± 0.4              & 67.3 ± 0.4            & 69.7 ± 0.3             \\
KCenter         & 43.2 ± 0.7       & 53.00 ± 0.2            & 74.0 ± 0.3         & 67.1 ± 0.36             & 64.2 ± 0.3            & 69.3 ± 0.4             \\
\midrule

GCSNTK            & 46.7 ± 0.2       & 54.2 ± 0.4& 75.8 ± 0.7         & 75.3 ± 2.0& 61.9 ± 1.3            & 72.8 ± 1.6\\ 
MSGC            & 46.6 ± 0.2       & 53.9 ± 0.1& 78.1 ± 0.2         & 74.8 ± 0.5& 70.2 ± 0.5            & 77.7 ± 0.5\\ 


GEOM            & 46.4 ± 0.2        & \textcolor{black}{\textbf{53.7 ± 0.1}} & \textcolor{blue}{\textbf{78.3 ± 0.6}}        & 78.8 ± 0.6 & 64.9 ± 1.5            & 84.7 ± 0.8\\




GCond          & \textcolor{black}{\textbf{47.1 ± 0.2}}       & 53.8 ± 0.1& 76.8 ± 0.5         & 72.5 ± 1.3& \textcolor{blue}{\textbf{73.1 ± 0.2} }          & 74.9 ± 0.2\\
SGDD          & 44.9 ± 0.2       & \textcolor{blue}{\textbf{53.3 ± 0.2}}& 76.3 ± 0.4         & 71.2 ± 0.5& 70.2 ± 0.3            & 74.3 ± 0.4\\
DosCond         & 46.4 ± 0.1       & 54.4 ± 0.1& 77.7 ± 0.2         & 72.0 ± 0.2& 68.1 ± 0.8            & 74.3 ± 0.6\\

GDEM         & 43.4 ± 1.3         & 54.2 ± 0.3& 76.3 ± 0.8         & \textcolor{blue}{\textbf{59.2 ± 0.3}}& 72.5 ± 0.3            & \textcolor{blue}{\textbf{70.1 ± 1.5}}\\
{\bf HyDRO}            & \textcolor{blue}{\textbf{47.1 ± 0.1}}      & 53.9 ± 0.1& \textcolor{black}{\textbf{78.2 ± 0.2}}        & \textcolor{black}{\textbf{71.0 ± 0.5}}& \textcolor{black}{\textbf{72.8 ± 0.2} }           & \textcolor{black}{\textbf{72.3 ± 0.6}}\\ 

\midrule

Whole           & 46.9 ± 0.2       & 53.5 ± 0.1 & 78.6 ± 0.6         & 78.8 ± 1.1& 71.8 ± 0.3            & 81.7 ± 1.1\\ 

\bottomrule
\end{tabular}%
\end{scriptsize}
\end{table*}

%% file: tables/Noise.tex
\begin{table*}[h]
\centering
\caption{
The experimental results for denoising ability are presented, including the mean accuracy and standard deviation for the test accuracy. 
The performance drop indicates the relative loss of accuracy compared to the original results before corruption. 
The best results are highlighted in \textcolor{blue}{\textbf{bold}}, while the second-best and third-best values are marked in \textbf{bold} and \textbf{\textit{bold}} respectively. 
$\downarrow$ and $\uparrow$ indicate that lower and higher values are better, respectively.
}
\setlength{\tabcolsep}{5pt} 
\renewcommand{\arraystretch}{1} 
\begin{scriptsize} 
\begin{tabular}{@{}cccccccc@{}}
\toprule
                             &                               & \multicolumn{2}{c}{{\bf Feature Noise}} & \multicolumn{2}{c}{{\bf Structural Noise}} & \multicolumn{2}{c}{{\bf Adversarial Structural Noise}}      \\ \cmidrule(l){3-4} \cmidrule(l){5-6} \cmidrule(l){7-8} 
\multirow{-2}{*}{{\bf Dataset}}    & \multirow{-2}{*}{{\bf Method}}      & Test Acc. $\uparrow$       & {\bf Perf. Drop} $\downarrow$      & {\bf Test Acc} $\uparrow$         & {\bf Perf. Drop} $\downarrow$     & {\bf Test Acc} $\uparrow$       & {\bf Perf. Drop} $\downarrow$ \\

\midrule
          & Whole                         & 46.71±2.39                & 41.10\%                & 73.20±0.22                   & 7.69\%              & 74.40±1.9                 & 6.18\%          \\

\cmidrule(l){2-8} 
    & {GCond}  & 55.48±0.60                & 30.04\%                & 72.52±0.72                   & 8.55\%              & 71.52±1.99                 & 9.81\%          \\

    & {DosCond} & 56.71±0.45                & 28.49\%                & \textcolor{black}{\textbf{73.32±0.30}}                   & 7.54\%              & \textcolor{black}{\textbf{72.32±1.62 }}                & 8.80\%          \\

    & { SGDD}  & 55.54±1.04                & 29.96\%                & 72.07±0.57                   & 9.12\%              & 70.65±2.04                 & 10.91\%          \\


    & {GEOM}  & 55.48±0.62                & 30.04\%                & 72.62±0.77                   & 8.42\%              &    71.32±0.81          & 10.06\%          \\

    & {MSGC}  & \textcolor{black}{\textbf{57.58±0.83}}               & 27.39\%                & \textcolor{blue}{\textbf{73.70±0.73}}                   & 7.06\%              & \textcolor{blue}{\textbf{73.39±0.83}}          & 7.45\%          \\
    
    & { GCSNTK}  & \textcolor{blue}{\textbf{57.59±0.40}}                & 27.38\%                & 71.83±0.53                   & 9.42\%              & \textcolor{black}{\textbf{\textit{72.16±0.22}}}                 & 9.00\%          \\

 \multirow{-10}{*}{\textit{Pubmed (0.30\%)}}                            & { HyDRO}   & \textbf{\textit{56.25±0.88}}                & 29.07\%                & \textbf{\textit{73.04±0.68}}                   & 7.89\%              & 71.80±0.67                 & 9.46\%          \\

\midrule
          & Whole                         & 38.73±0.61                & 17.94\%                & 47.16±0.16                   & 0.08\%              & 46.58±0.13                 & 1.31\%          \\

\cmidrule(l){2-8} 
    & {GCond}  & 41.69±0.49                & 11.67\%                & \textcolor{blue}{\textbf{47.19±0.11 }}                  & 0.02\%              & \textcolor{black}{\textbf{46.63±0.09 }}                & 1.21\%          \\

    & {DosCond}  & 42.95±0.32                & 9.00\%                & \textbf{\textit{46.76±0.18}}                   & 0.93\%              & \textcolor{blue}{\textbf{46.66±0.12}}                 & 1.14\%          \\

    & { SGDD}  & 41.49±0.69                & 12.10\%                & 43.68±0.37                   & 7.46\%              & 44.32±0.36                 & 6.10\%          \\

    & { SFGC}  & \textbf{\textit{43.53±0.47}}                & 7.78\%                & 44.22±0.46                  & 46.31\%              & 41.19±0.43          & 12.73\%          \\

    & {GEOM}  & 42.75±0.43                & 9.43\%                & 45.54±0.38                    & 3.52\%              &  46.02±0.26          & 2.50\%          \\

    & {MSGC}  & \textcolor{blue}{\textbf{43.86±0.39 }}               & 7.08\%                & 46.38±0.10                    & 1.74\%              &  45.88±0.17          & 2.80\%          \\
    & { GCSNTK}  & 43.13±0.82                & 8.64\%                & 45.89±1.02                   & 2.78\%              & 46.11±0.76                 & 2.31\%          \\

 \multirow{-10}{*}{\textit{Flickr (0.5\%)}}                            & { HyDRO}   & \textcolor{black}{\textbf{43.82±0.83}}                & 7.16\%                & \textcolor{black}{\textbf{47.04±0.10}}                    & 0.34\%              & \textbf{\textit{46.55±0.14}}                & 1.38\%          \\

 \bottomrule
\end{tabular}%
\end{scriptsize}
\label{tab:robustness}
\end{table*}
